\documentclass{ieeeaccess}
\usepackage{cite}
\usepackage{amsmath,amssymb,amsfonts}
\usepackage{algorithmic}
\usepackage{graphicx}
\usepackage{textcomp}
\usepackage{tabularx,booktabs}
\newcolumntype{C}{>{\centering\arraybackslash}X} 
\usepackage{lipsum}

\usepackage{amsmath}
\usepackage[flushleft]{threeparttable}
\usepackage{makecell,booktabs}
\usepackage{pdflscape}
\usepackage{afterpage}
\usepackage{array}
\usepackage{graphicx}
\usepackage{caption}

\def\BibTeX{{\rm B\kern-.05em{\sc i\kern-.025em b}\kern-.08em
    T\kern-.1667em\lower.7ex\hbox{E}\kern-.125emX}}
\begin{document}
\history{Date of publication xxxx 00, 0000, date of current version xxxx 00, 0000.}
\doi{10.1109/ACCESS.2024.DOI}

\title{Object Tracking in a $360^o$ View: A Novel Perspective on Bridging the Gap to Biomedical Advancements}
\author{\uppercase{Mojtaba S. Fazli}\authorrefmark{1,2} and
\uppercase{Shannon Quinn\authorrefmark{3, 4}}}
\address[1]{Stanford Center for Artificial Intelligence in Medicine and Imaging, Stanford University, Palo Alto , CA, USA (e-mail: mfazli@stanford.edu)}
\address[2]{Harvard Ophthalmology Artificial Intelligence lab, Harvard University, Cambridge, MA, USA (e-mail: mfazli@meei.harvard.edu)}
\address[3]{School of Computing, The University of Georgia, Athens, GA, 30602 USA}
\address[4]{Department of Cellular Biology, The University of Georgia, Athens, GA, 30602 USA (e-mail: spq@uga.edu)}

\tfootnote{
The authors would like to thank Prof. Silvia N.J. Moreno and Prof. Gary E. Ward for their collaboration on Toxoplasma projects. We also acknowledge Prof.  Chakra Chennubhotla and Prof. Frederick D Quinn for their collaboration on OrNet project. We also acknowledge and the Georgia Informatics Institutes for providing the computational resources that made this project possible. Also We kindly acknowledge Prof. Tobias Elze, Dr. Mengyu Wang, and Dr. Nazlee Zebardast from Harvard Ophthalmology AI lab. This work was supported in part by the GII Fellowship. We also gratefully acknowledge the support of NVIDIA Corporation for their donation of a Titan X Pascal GPU, which was utilized in this research. Additionally, we extend our thanks to Google for their generous research grant, which enabled us to use their Compute Platform. This work was partially supported by the NSF Advances in Biological Informatics (ABI) under award number 1458766, and by the US Public Health Service grants AI139201 and AI137767.
}

\markboth
{MS. Fazli \headeretal: Object Tracking in a $360^o$ View: A Novel Perspective on Bridging the Gap to Biomedical Advancements}
{MS. Fazli \headeretal: Object Tracking in a $360^o$ View: A Novel Perspective on Bridging the Gap to Biomedical Advancements}


\begin{abstract}

Object tracking serves as a cornerstone of modern technological innovation, with applications spanning diverse fields such as defense systems, autonomous vehicles, and the cutting edge of biomedical research. Fundamentally, it involves the precise identification, monitoring, and spatiotemporal analysis of objects across sequential frames, enabling a deeper understanding of dynamic behaviors. In cell biology, this capability is indispensable for unraveling the intricacies of cellular mechanisms—offering insights into cell migration, interactions, and responses to external stimuli like drugs or pathogens. These insights not only illuminate fundamental biological processes but also pave the way for breakthroughs in understanding disease progression and therapeutic interventions.

Over the years, object tracking methodologies have evolved significantly, progressing from traditional feature-based approaches—leveraging color, shape, and edges—to advanced machine learning and deep learning frameworks. While classical methods demonstrate reliability under controlled conditions, they falter in complex environments characterized by occlusions, variable lighting, and high object density. In contrast, modern deep learning models excel in such challenging scenarios, offering unparalleled accuracy, adaptability, and robustness.

This paper presents a comprehensive review of object tracking techniques, systematically categorizing them into traditional, statistical, feature-based, and machine learning paradigms. We place a particular emphasis on their applications in biomedical research, where precise tracking of cells and subcellular structures is critical for advancing our understanding of health and disease. Key performance metrics, including accuracy, computational efficiency, and adaptability, are examined to facilitate meaningful comparisons across methodologies.

Despite remarkable advancements, the field still faces a pivotal challenge: \textit{Why does the development of a fully integrated, robust, and scalable end-to-end tracking system capable of handling diverse biomedical scenarios remain elusive?} Addressing this question, we identify the limitations of current technologies and explore emerging trends poised to overcome these barriers. Our goal is to provide a roadmap for the development of next-generation object tracking systems—tools that will not only transform biomedical research but also catalyze innovations across broader scientific and technological domains.

\end{abstract}

\titlepgskip=-15pt

\maketitle

\section{Introduction}
\label{sec:introduction}

The study of motion patterns and trajectory analysis is fundamental across a broad spectrum of disciplines, ranging from security and autonomous systems to medical diagnostics and cellular biology \cite{b175},\cite{b42}. At the heart of these systems lies object tracking, which facilitates the extraction of critical metrics such as distance, velocity, acceleration, and deformation angles. These metrics find applications in a wide range of domains, including infrastructure monitoring, traffic analysis, and, most significantly, the observation of morphological and behavioral changes in biological entities.

\subsection{Advanced Tracking in Biomedical Imaging: Unveiling Cellular Dynamics}

In biomedical research, the ability to accurately track objects at the cellular level is not just a technical achievement—it represents a critical step toward understanding the dynamic behavior of living systems. Cells are constantly in motion, responding to their environment, undergoing transformation, and interacting with other biological entities. By tracking these movements with precision, researchers are able to delve into the hidden mechanisms of health and disease, uncovering the subtle yet profound changes that can signify everything from the effectiveness of a drug to the progression of a disease. The sheer diversity of biological objects—from single cells to multicellular organisms—demands tracking solutions that are both flexible and robust, capable of adapting to the varying scales and behaviors of these entities. Through advanced tracking technologies, researchers are now able to quantify these dynamic processes with unparalleled accuracy, offering new insights into the underlying principles that govern biological function.

The advent of modern microscopy techniques has revolutionized biomedical imaging, transforming video microscopy into a key instrument for unveiling the complex choreography of cellular and subcellular dynamics. With the ability to capture high-resolution videos in real time, researchers can now visualize the intricate behaviors of cells and tissues as they respond to various stimuli. Whether observing cells in their natural environment or under experimental conditions, video microscopy provides a window into the immediate and long-term effects of external perturbations. These may include the introduction of pathogens, the application of therapeutic drugs, or exposure to environmental toxins—each of which can trigger unique cellular responses that must be carefully tracked to understand their full impact.

One of the most powerful applications of tracking in biomedical imaging lies in the study of the immune system. Immune cells such as neutrophils, for example, play a vital role in the body’s defense against infection. By tracking the movement of these cells within the bloodstream, researchers can monitor how they navigate toward infection sites, respond to pathogens, and engage in the complex process of immune activation. This real-time tracking of immune cell behavior is essential for deciphering the body's natural defense mechanisms and for developing treatments that can modulate immune responses more effectively.

Similarly, cancer research has greatly benefited from advances in cellular tracking. Tumor cells are known for their ability to move and metastasize, spreading from the original tumor site to other parts of the body. By tracking the migration of cancer cells, researchers can gain insights into the mechanisms of tumor progression and metastasis, revealing potential targets for intervention. This is particularly important for developing strategies that can prevent or slow down the spread of cancer, offering hope for more effective therapies in the fight against this complex disease.

Another critical area of application is pathogen tracking. Pathogens such as \textit{Toxoplasma gondii} and malaria parasites undergo complex life cycles within host organisms, interacting with host cells in ways that can be both subtle and devastating. The ability to track these pathogens as they invade, replicate, and spread within the host is crucial for understanding their biology and for developing interventions that can disrupt their lifecycle. Advanced tracking algorithms have enabled researchers to capture these events with remarkable precision, leading to new discoveries about how pathogens evade immune responses and establish infections.

Cell tracking has also played a transformative role in the field of protein design, a cornerstone of biotechnology and therapeutic innovation. Advanced tracking methodologies allow researchers to visualize and quantify protein interactions, cellular uptake, and intracellular trafficking in real time. For instance, monitoring how designed proteins bind to receptors or other cellular targets provides critical insights into their efficacy and specificity. These capabilities are invaluable for developing next-generation therapeutic proteins, optimizing drug delivery systems, and studying dynamic biological responses under diverse conditions. By leveraging such technologies, researchers are pushing the boundaries of molecular design and accelerating the discovery of novel treatments \cite{b286}, \cite{b287}, \cite{b288}, \cite{b289}.

As biomedical research continues to evolve, the need for sophisticated tracking systems becomes ever more pressing. Biological systems are inherently variable, with cells and pathogens exhibiting diverse behaviors that can be difficult to capture using traditional methods. The development of new algorithms that can reliably handle the complexity and variability of biological data is therefore a key focus for researchers in the field. These systems must be able to track objects over time and across different spatial scales, accommodating the unique challenges posed by each biological application.

In summary, advanced tracking in biomedical imaging is not merely a technical tool—it is a gateway to understanding the dynamic processes that define life. Whether tracking immune responses, cancer cell migration, pathogen behavior, or protein interactions, these technologies are helping researchers to unravel the mysteries of biology in ways that were previously unimaginable. As new algorithms and imaging techniques continue to emerge, the potential for uncovering new insights into cellular dynamics grows exponentially, bringing us closer to a future where we can predict, prevent, and treat diseases with unprecedented precision. The ongoing advancements in tracking are truly unveiling the hidden dynamics of life at the cellular level, enabling biomedical breakthroughs that have the power to transform healthcare.

\subsection{Pushing the Boundaries: Technological Progress in Object Tracking Algorithms}

In recent years, the convergence of computer vision and machine learning has transformed object tracking, paving the way for more sophisticated and adaptive systems. These advancements are underpinned by breakthroughs in deep learning architectures, which have enabled models to capture intricate motion patterns and dependencies from vast datasets. Consequently, object tracking systems now demonstrate greater accuracy, robustness, and generalizability across a wide variety of applications \cite{b122},\cite{b114},\cite{b242}. One domain that has particularly benefited from these innovations is biomedical image analysis, where object tracking has proven vital in tasks such as cell segmentation, motion tracking, and anomaly detection within video microscopy.

The infusion of deep learning into tracking systems has led to the development of more autonomous methods, especially through the utilization of convolutional neural networks (CNNs) and recurrent neural networks (RNNs). CNNs, known for their exceptional ability to capture spatial features, have revolutionized object detection and tracking in dense, cluttered environments. They can automatically extract relevant features from raw video data, reducing the need for extensive human intervention. RNNs, on the other hand, excel at modeling temporal dependencies across video frames, allowing tracking systems to better predict and maintain object trajectories over time, especially in highly dynamic scenes. This synergy between CNNs and RNNs has been the backbone of many high-performance tracking models that excel in handling challenges like occlusion, variable motion patterns, and changes in object appearance over time.

More recently, the integration of transformer-based architectures has pushed the boundaries of object tracking even further. These architectures, initially designed for natural language processing, have shown significant promise in tracking applications by modeling long-range dependencies more effectively than traditional RNNs. Research presented at CVPR 2023 highlights the superior ability of transformers to handle complex multi-object tracking (MOT) scenarios, where multiple objects need to be tracked across varied domains and scales \cite{b243}. These models, often referred to as Vision Transformers (ViTs), have outperformed traditional CNN-RNN architectures in cases where spatial and temporal relationships are particularly complex.

Another area where object tracking has seen substantial progress is in self-supervised and unsupervised learning techniques. Traditionally, tracking systems have relied heavily on manually labeled data to train models, which is often time-consuming and prone to human error. However, NeurIPS 2023 introduced new self-supervised learning methods capable of learning to track objects without large volumes of annotated data. These methods are particularly useful in domains like biomedical research, where labeled data is scarce or expensive to obtain \cite{b258}. By leveraging self-supervision, these models can autonomously discover meaningful patterns in video data, leading to better generalization and reducing reliance on labor-intensive labeling processes.

Moreover, object tracking has seen improvements in multi-object environments, a task that has historically posed significant challenges due to factors like occlusion, object deformation, and complex interactions. In addition to the progress made by transformer-based models, graph neural networks (GNNs) have emerged as powerful tools for modeling relationships between multiple objects within a scene. By treating objects as nodes and their interactions as edges in a graph, GNNs can dynamically track objects even in highly crowded environments, as demonstrated by recent works\cite{b260}. These approaches are proving instrumental in applications such as pedestrian tracking in crowded urban spaces or studying cell dynamics in biological systems.

Finally, end-to-end tracking frameworks are becoming more prevalent, offering fully integrated pipelines that combine detection, association, and tracking in a seamless process. These systems bypass the traditional multi-stage processes, resulting in improved efficiency and performance. MOTR, a transformer-based end-to-end tracking framework introduced at \cite{b256}, is an example of this trend, excelling in scenarios that require tracking across long sequences with minimal computational overhead \cite{b256}.

These technological advancements represent a paradigm shift in object tracking, allowing systems to handle increasingly complex environments and applications. In the following sections of this paper, we will review the most important and impactful advancements in greater detail, offering a deeper dive into the state-of-the-art models and frameworks shaping the future of object tracking.






\subsection{Current Limitations in Object Tracking: A Broad Perspective}

Despite significant advancements in object tracking, state-of-the-art methods continue to face numerous challenges, particularly when applied across diverse domains. Traditional tracking methods often rely on assumptions about the data, such as object shape, size, or movement patterns, which limits their generalizability to more complex environments, such as biomedical research or autonomous driving. For example, many existing algorithms are designed to track rigid objects, making them unsuitable for non-rigid entities that change shape over time, such as cells in biomedical videos \cite{b162}, \cite{b163}, \cite{b231}.

A key limitation of current methods is their reliance on specific types of videos or domains. Algorithms that work well in controlled environments often fail in more dynamic, multi-scale settings. Recently multi-scale object tracking approaches are introduced to addressing this challenge to some extent, but occlusions and object interactions remain unsolved problems in real-world applications \cite{b256}. Further, state-of-the-art techniques are predominantly single-target trackers, which hinders their performance in multi-object scenarios, such as studying cellular interactions or pedestrian tracking in crowded scenes \cite{b231}. \textit{CVPR 2023} introduced advanced multi-object tracking (MOT) algorithms designed to handle occlusions and object deformations, but these models still struggle with cluttered environments where objects overlap and interact in unpredictable ways \cite{b257}.

Another significant issue is the heavy dependency on manually annotated data. Many tracking algorithms, particularly those based on supervised learning, require thousands of labeled samples to achieve acceptable performance. In domains like biomedical research, acquiring such data is time-consuming and prone to human error. Recent developments have explored self-supervised learning approaches to mitigate this issue, reducing the need for large labeled datasets. However, achieving competitive performance with unsupervised or self-supervised methods in real-world applications is still an ongoing challenge \cite{b258}.

Furthermore, although some tracking methods show robustness in certain applications, such as in cell biology, they often fail to meet the required accuracy for generalization across multiple domains. Current models also lack end-to-end architectures capable of tracking multiple objects across different domains and timescales. An object-centric multiple-object tracking approach is recently presented that addresses the need for reduced annotation while maintaining temporal consistency across long sequences, but it remains limited when applied to more complex multi-domain settings \cite{b259}.

The absence of a robust end-to-end tracking system that generalizes across different object types and domains underscores the need for further innovation. For instance, Zeng F. et al. proposed transformer-based solutions like \textit{MOTR}, which use self-attention mechanisms for end-to-end tracking, but these approaches are still in early stages of development and struggle with scalability across diverse applications \cite{b256}. Furthermore, despite promising progress in multi-object tracking with transformers, as explored in \cite{b257} and \cite{b258}, more work is needed to improve tracking performance in complex, real-world scenarios \cite{b258}, \cite{b259}.

In Summary, The field of object tracking continues to advance, but there remain significant challenges in adapting existing models for use in real-world, dynamic, and multi-domain environments. Future research must focus on developing more adaptive, scalable, and generalizable tracking frameworks, particularly by leveraging advances in self-supervised learning, multi-object tracking, and end-to-end architectures.

\subsection{Key Objectives of This Study}

The primary objective of this review is to address the persistent limitations of current object tracking systems, with a special emphasis on developing a new framework that is specifically tailored to the unique challenges of biomedical applications. As the demand for more precise, scalable, and adaptive tracking solutions grows within the biomedical domain, it is essential that object tracking systems evolve accordingly. By building on the latest advancements in computer vision and machine learning, this framework aims to enhance the accuracy, robustness, and generalizability of object tracking algorithms across a variety of real-world and experimental scenarios.

To achieve this vision, the review is structured in several key parts. First, we provide a comprehensive overview of image processing techniques essential for video analysis, ranging from classical approaches to the most recent innovations in deep learning. This exploration of the literature offers a foundation for understanding the essential building blocks of object tracking. Next, we present a detailed taxonomy of existing object tracking systems, highlighting their core strengths and pinpointing critical limitations. This classification serves as a basis for evaluating the current state of the art in object tracking and identifying where future advancements are most needed.

Building upon this foundation, the review then shifts focus to segmentation and tracking methods specifically designed for biomedical applications. Accurate segmentation and tracking are paramount in biological systems, where researchers must precisely quantify changes in cellular behavior in response to perturbations such as pathogen invasion, toxin exposure, or drug treatment. By critically evaluating existing approaches, we aim to identify the main challenges limiting current systems and propose innovative solutions that can improve the robustness, scalability, and accuracy of object tracking in this demanding context. 

A key aspect of this review is the identification of the key features that any next-generation object tracking system must possess to meet the evolving needs of biomedical research:

\begin{itemize}
    \item \textbf{Extensiveness:} The ability to handle a wide range of video types, from high-resolution microscopy videos to lower-quality surveillance footage, ensuring versatility and broad applicability.
    \item \textbf{Robustness:} Resilience against common challenges such as occlusions, shifts, and misalignments, ensuring reliable performance under adverse conditions.
    \item \textbf{Trainability:} The capacity to learn from past experiences and adapt to new data, allowing the system to improve its accuracy and efficiency over time.
    \item \textbf{Multi-Domain Compatibility:} Flexibility to be deployed across different domains, including video microscopy of \textit{Toxoplasma gondii}, malaria parasites, and other in vivo or in vitro biomedical contexts \cite{b114}.
    \item \textbf{End-to-End Functionality:} The ability to autonomously identify and track objects across varied video types without requiring manual parameter tuning or supervision.
    \item \textbf{Scalability:} The capability to efficiently manage and process both small-scale and large-scale videos, including those exceeding typical memory and computational constraints.
    \item \textbf{Code Availability:} Open-source implementation of the tracking system, fostering transparency, reproducibility, and collaboration within the research community.
\end{itemize}

By meeting these criteria, the proposed frame aims to Reimagine the landscape of biomedical object tracking, enabling it to rise to the complexity and demands of modern scientific inquiry. In the sections that follow, we will provide an in-depth review of the most critical advancements and innovations in this field, offering a detailed roadmap for the development of next-generation object tracking systems that meet the rigorous needs of biomedical research.

\section{From Frames to Flow: The Art and Science of Object Tracking}

\textit{Video processing} involves the analysis and manipulation of a sequence of time-varying images, with each image, or \textit{frame}, representing a snapshot in time. These video signals can be analog or digital, but in modern computer vision and video analysis, digital signals are favored due to their superior precision, flexibility, and seamless integration with automated systems \cite{b1}. In essence, a \textit{digital video} is a series of still images, where each frame is composed of \textit{pixels}. These pixels, the fundamental building blocks of an image, represent varying intensity levels of light, and their spatial arrangement forms the visual structure of each frame.

The process of capturing these images, known as imaging, is pivotal in video analysis across multiple domains, from security systems \cite{b175} to cellular biology \cite{b42}. Unlike static images, videos introduce a temporal component, where pixel intensity values change over time, necessitating the use of advanced video processing techniques. These techniques enable the modification, quantification, and mathematical interpretation of thousands of sequential images, a task that requires sophisticated digital image processing tools \cite{b37, b38}.

In the biomedical field, video processing serves as a critical tool for tracking the movement and interactions of biological entities such as cells and pathogens. The dynamic nature of biological systems, including the behavior of subcellular structures like mitochondria and microtubules, makes video analysis indispensable for understanding cellular functions and disease mechanisms \cite{b139, b140}. Through the precise analysis of motion and interactions, video processing can uncover key insights into biological processes that static imaging methods simply cannot capture.

Recent advancements in machine learning, particularly artificial intelligence (AI) and deep learning, have dramatically improved video processing capabilities. These AI-driven techniques have been seamlessly integrated into video analysis pipelines, enhancing the accuracy of critical tasks such as motion detection, object tracking, and scene understanding \cite{b116, b117}. Machine learning models now autonomously learn patterns and features from data, allowing them to handle increasingly complex video processing challenges with minimal human intervention.

One of the most significant advantages of AI-based video processing is its ability to adapt to real-time, dynamic environments. This is particularly vital in biomedical applications, where data often contain high levels of noise, occlusions, and fluctuations in imaging conditions. In such scenarios, AI algorithms can make real-time adjustments to improve tracking accuracy, ensuring that critical biological events—such as cellular interactions or movement—are captured despite challenging conditions. For instance, the ability to dynamically adjust to noise or interference allows for more robust cell movement tracking, safeguarding the integrity of the analysis and minimizing the risk of missing key biological phenomena.

\subsection{ Empowering Automation: AI\textquotesingle{s} Impact on Large-Scale Video Processing}

The integration of AI-powered video processing techniques has revolutionized the analysis of large-scale video datasets, offering unprecedented automation and accuracy across various domains. In medical diagnostics, these advancements have been particularly transformative, enabling the automated analysis of complex biological activities that were previously too intricate for manual interpretation. AI models, trained on extensive video sequences of cellular activities, can detect subtle changes and anomalies that may indicate the early onset of diseases, such as cancer or neurodegenerative disorders. This early detection capability facilitates timely diagnosis, which in turn enables more effective treatment interventions, ultimately improving patient outcomes \cite{b116, b117}.

Moreover, AI-driven automation significantly reduces the workload for human analysts, making the analysis of large datasets not only feasible but also more efficient. In fields such as pathology, where vast quantities of video data are generated daily, manual analysis can be overwhelming and prone to human error. AI models alleviate this burden by processing and analyzing the data rapidly, allowing for real-time insights and high-throughput diagnostic workflows. This level of automation is crucial for scalability in modern medical research and clinical settings, where both speed and precision are essential.

Another compelling example of AI's impact in medical diagnostics and biomedical research can be found in the field of ophthalmology, particularly in glaucoma research. The integration of unsupervised and self-supervised learning methods has proven highly effective in the phenotyping of optical coherence tomography (OCT) scans, as demonstrated by Kazeminasab et al. In their study, deep learning models, coupled with dimensionality reduction techniques like UMAP, enabled the successful transfer of phenotypes across diverse datasets, significantly improving the generalization of AI models across varying clinical environments \cite{b251}. This advancement not only facilitates more accurate diagnosis but also strengthens the robustness of AI applications in ophthalmology.

In a related development, Eslami et al. introduced PyVisualFields, a Python-based package designed specifically for visual field analysis. This tool addresses the limitations of existing R-based solutions by offering an AI-driven, accessible platform that simplifies ophthalmological research \cite{b252}. PyVisualFields enhances the efficiency of visual field data analysis, enabling more scalable and user-friendly research workflows.

Further contributions in this field include advancements in artifact-tolerant clustering for ophthalmic images, where robust AI models have been employed to detect glaucoma more effectively, even in the presence of image artifacts. These models, detailed in studies on glaucoma detection, highlight the significance of AI’s role in ensuring more reliable and accurate diagnoses under challenging conditions \cite{b253, b254}.

Collectively, these developments underscore AI's potential not only to automate the analysis of large-scale video datasets but also to elevate diagnostic capabilities across various medical fields, such as pathology and ophthalmology. Such innovations mark an essential step toward advancing our understanding of complex diseases and improving patient outcomes.

In addition to diagnostics, AI has been pivotal in advancing video processing techniques for research in cellular biology and other biomedical fields. Automated cell segmentation and tracking are prime examples of how AI can be used to study the behavior of biological entities over time. By integrating deep learning algorithms with video data, AI enables the identification, segmentation, and tracking of cells as they move and interact with each other in dynamic environments. This automated approach allows researchers to gain deeper insights into processes such as cellular migration, mitosis, and response to drug treatments, without the need for labor-intensive manual tracking.

The impact of AI on video processing extends beyond just automation. These intelligent systems continuously learn and improve through exposure to more data, becoming more adept at handling complex variations in video quality, noise, and occlusions. In real-time applications, AI algorithms can adapt on-the-fly, adjusting to changing environmental conditions, ensuring that critical biological events are captured with minimal interference or signal degradation. This adaptability is especially valuable in biomedical applications where the precision of tracking and segmentation is vital for understanding disease mechanisms, developmental processes, or treatment effects.

\begin{figure}
   \centerline{\includegraphics[width=.5\textwidth]{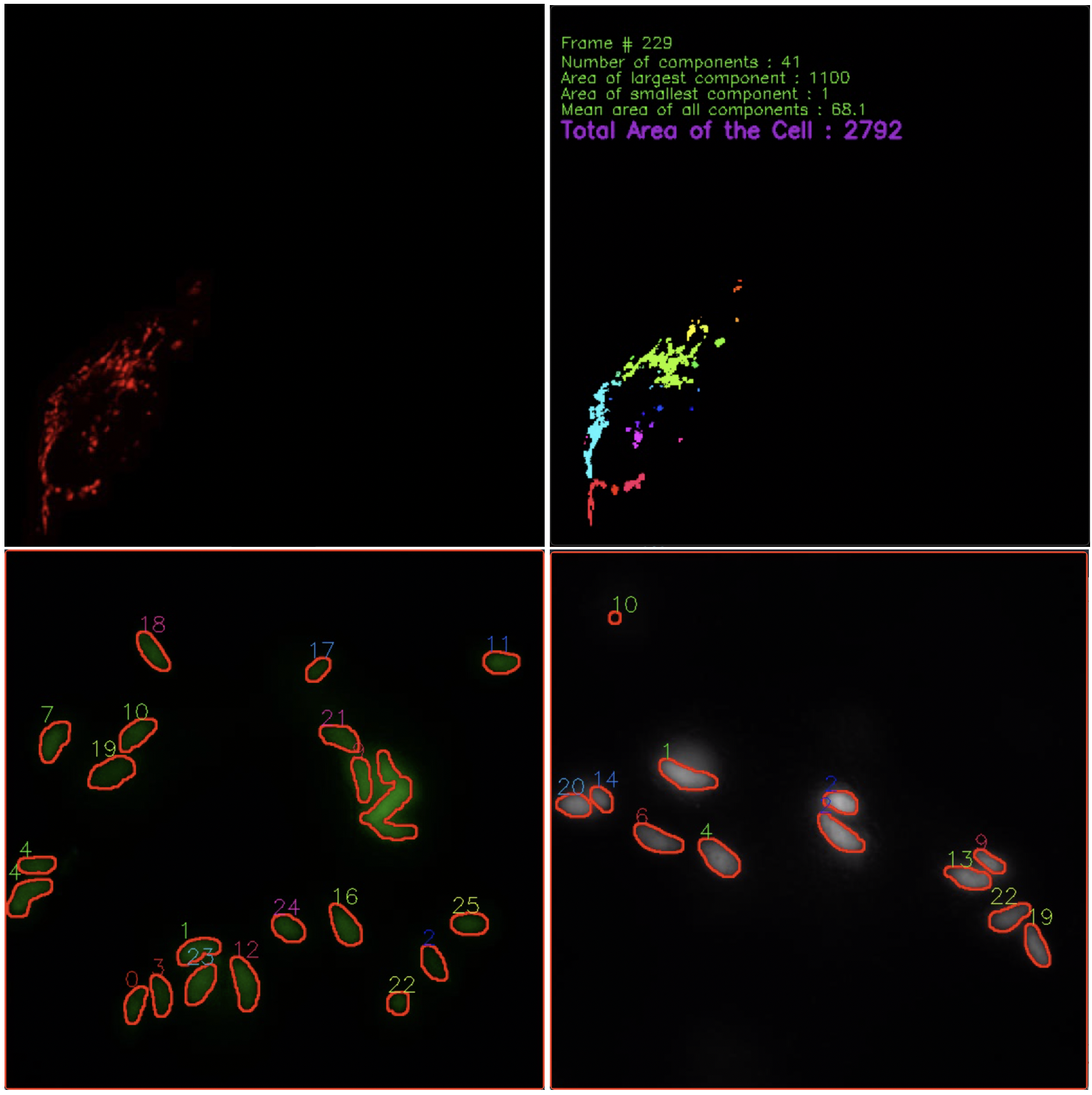}}
\caption{\textit{Sample cell segmentation and tracking in mitochondria of lung cell and} \textit{T. gondii}. \textit{The top row indicates the mitochondria in an original frame and the segmented one across time in the top right. The cells are segmented using the watershed algorithm and connected component labeling to see how many connected components we have over time and how they are changing along with the video. The bottom row illustrates sample tracking of} \textit{T. gondii} \textit{using segmentation and object association.}}
\label{fig:1-1}
\end{figure}

Figure \ref{fig:1-1} illustrates some advancements in AI-driven video processing within biomedical applications. It showcases examples of cell segmentation and tracking using advanced algorithms, demonstrating how these techniques are being effectively employed in real-world biological studies. These methods, powered by deep learning and other AI models, have not only improved the accuracy and scalability of video analysis but also opened new avenues for research, providing a robust foundation for future innovations in medical and biological sciences.

\subsection{Seamless Detection, Classification, and Tracking: Unraveling Temporal and Spatial Dynamics in Object Tracking}

Object tracking refers to the process of continuously estimating the trajectory of an object as it moves through a scene, typically within the image plane of a video \cite{b6}. This task involves detecting the object in consecutive frames and associating the detections to form a coherent temporal trajectory. Object tracking is pivotal in deriving metrics such as an object’s texture, area, and motion model, all of which describe its behavior over time. Depending on the complexity of the video and the application, object tracking can be applied to either single or multiple objects.

Tracking becomes particularly challenging due to factors such as object appearance variations, occlusions, and complex background environments. Objects may change shape, fade away, or move in unpredictable trajectories, complicating the task for tracking algorithms. Furthermore, object classification into rigid (e.g., cars, balls) and non-rigid (e.g., humans, animals) categories is crucial for selecting the appropriate algorithm. Non-rigid object tracking requires sophisticated models capable of handling deformable shapes and dynamic environments.

\subsubsection{Significance of Temporal Analysis in Object Tracking}

Temporal analysis is essential in object tracking, allowing models to capture the continuity of an object’s motion over time. Conventional methods, such as optical flow, estimate an object’s trajectory based on pixel movement between consecutive frames but are often limited in handling long-term dependencies, occlusions, and abrupt motion changes. Probabilistic methods like Kalman filters and particle filters offer predictive capabilities by estimating an object’s future positions based on past observations, yet these methods still struggle with more complex dynamics and high-dimensional data \cite{b244}.

Recurrent Neural Networks (RNNs) introduced memory into temporal analysis by maintaining hidden states that store previous time-step information. However, RNNs suffer from vanishing and exploding gradients, which limit their ability to capture long-range dependencies \cite{b245}. This challenge was addressed by Long Short-Term Memory (LSTM) networks, which use gates to regulate the flow of information and retain important data over extended time periods, making them more effective for long-sequence tracking \cite{b246}, \cite{b247}.

In recent years, transformer models have replaced LSTMs in temporal analysis tasks due to their self-attention mechanisms. Unlike RNNs and LSTMs, transformers process entire sequences in parallel, making them highly efficient and scalable for large datasets. Transformers such as DETR (DEtection TRansformers) and Temporal Fusion Transformer (TFT) have demonstrated their ability to model both spatial and temporal dependencies effectively, even in complex environments \cite{b178, b248, b249}. Hybrid models that integrate transformers with convolutional networks have further advanced the field, achieving a balance between spatial and temporal data processing \cite{b261}.

\subsubsection{LSTMs}

Long Short-Term Memory (LSTM) networks, a type of recurrent neural network (RNN), were developed to solve the vanishing gradient problem that impairs traditional RNNs when learning long-term dependencies. LSTMs include memory cells and three key gates—input, forget, and output gates—that control the flow of information. This structure allows the network to retain important data over long time spans and discard irrelevant information. LSTMs have been widely used in tasks requiring both short- and long-term dependency modeling, such as time series forecasting, speech recognition, and natural language processing (NLP). Despite their utility, LSTMs face scalability and parallelization challenges, prompting their gradual replacement by transformer architectures \cite{b249, b250}.

\subsubsection{Transformers}

In recent years, transformers have increasingly replaced LSTMs for temporal analysis tasks, including time series forecasting and sequence prediction. While LSTMs are powerful for capturing long-range dependencies, they struggle with parallelization and suffer from vanishing gradient issues in long sequences \cite{b250}. Transformers, introduced by Vaswani et al. \cite{b178}, utilize self-attention mechanisms that process entire sequences at once, enhancing scalability and efficiency for large datasets. Their ability to model long-term dependencies without sequential constraints makes transformers faster and more accurate than LSTMs.

Transformers have evolved through variations such as the Temporal Fusion Transformer (TFT) \cite{b262} and Informer\cite{b263}, both of which are designed specifically for time series data. These models integrate both static and dynamic features, capturing complex temporal dependencies. Hybrid approaches, which combine transformers with convolutional networks, have also gained traction, offering enhanced handling of both spatial and temporal data \cite{b265}. Transformers’ recent advancements focus on reducing memory requirements and improving interpretability, making them ideal for various temporal analytics applications \cite{b262,b264}.

\subsubsection{Transformers in Temporal Analysis}

Transformers, first introduced by Vaswani \textit{et al.} \cite{b178}, have revolutionized fields like natural language processing (NLP) and time series analysis by replacing recurrent layers with self-attention mechanisms. Unlike RNNs and LSTMs, transformers capture relationships between distant elements in a sequence without processing them sequentially. This capability enables transformers to handle long-term dependencies more effectively while improving parallelization.

In time series forecasting, transformers eliminate the need for recurrent layers, enabling efficient parallel processing. This capability enhances the model's ability to capture long-term dependencies across sequences. Extensions such as the Temporal Fusion Transformer (TFT) \cite{b262} and Informer \cite{b263} were developed to address specific challenges in time series data, such as multiple seasonalities and non-linearities. These models significantly improve scalability and performance while retaining interpretability.

\subsubsection{Transformers in Object and Cell Tracking}

Transformers have recently become prominent in object and cell tracking, offering a powerful alternative to traditional models like convolutional neural networks (CNNs) and recurrent neural networks (RNNs). In object tracking, transformers’ self-attention mechanism efficiently models long-range dependencies and spatial-temporal relationships, crucial for tracking objects through multiple video frames. Unlike LSTM-based models that struggle with long sequences and parallelism, transformers process multiple frames simultaneously, making them highly scalable. Vision Transformers (ViTs) and DETR (DEtection TRansformers) are prime examples adapted for object tracking by encoding temporal information alongside spatial features \cite{b248, b266, b267, b268, b271}. DETR has demonstrated success in end-to-end tracking tasks without requiring handcrafted features, simplifying the tracking process.

In cell tracking, transformers address complex spatiotemporal dynamics, such as cell movement in biological tissues or time-lapse microscopy videos. These models effectively capture subtle changes in cell morphology and movement, improving tracking accuracy. Additionally, transformers like the Temporal Fusion Transformer (TFT) have been highly effective in integrating multimodal data (images and time-series data), providing interpretable and accurate tracking results \cite{b252, b269}. These advancements are critical in applications such as cancer research, drug discovery, and developmental biology, where precise cell tracking is essential.

In the following sections, we review further examples and applications of these tracking methods in various domains, illustrating their adaptability and effectiveness in addressing complex tracking challenges.

\subsection{Object Detection}
Object detection plays a crucial role in object tracking by identifying and localizing objects within each frame of a video. It often involves segmenting moving objects from a static or dynamic background, producing pixel-wise masks that delineate the boundaries of each object \cite{b43}. Segmentation techniques such as \textit{watershed segmentation} and \textit{k-means clustering} are commonly employed in biomedical imaging to separate objects based on distinct characteristics like color, intensity, or texture. Additionally, modern deep learning methods like You Only Look Once (YOLO), Region-Based Convolutional Neural Networks (R-CNN), and Mask R-CNN have gained popularity for real-time object detection and segmentation tasks. These approaches are particularly useful when objects may appear or disappear within a video, necessitating continuous detection and segmentation for accurate tracking.

In biomedical applications, object detection faces several challenges, such as variations in object appearance caused by changes in viewpoint, occlusions, background clutter, and inconsistent lighting conditions \cite{b46}. The complexity of biological samples, which often contain visually similar objects, further complicates detection. Despite these challenges, advancements in machine learning and computer vision have led to the development of robust object detection methods that can handle these complexities.

Deep learning-based models, such as those utilizing Convolutional Neural Networks (CNNs), have significantly improved the accuracy and robustness of detection in complex environments \cite{b160, b161}. These models, such as Region-Based Convolutional Neural Networks (R-CNN), Faster R-CNN, Mask R-CNN, and YOLO, have achieved great success in real-time object detection tasks. These methods use deep hierarchical features to detect and localize objects with high speed and accuracy, making them ideal for applications requiring real-time processing.

Cheng et al. \cite{b46} proposed a taxonomy of object detection methods, categorizing them into four types: \textit{template matching-based}, \textit{knowledge-based}, \textit{Object-Based Image Analysis (OBIA)}, and \textit{machine learning-based} methods. While originally developed for remote sensing images, this taxonomy is generalizable to other domains, including biomedical imaging. Machine learning-based methods, particularly deep learning models, have become dominant in object detection tasks across multiple fields, consistently demonstrating superior performance in detecting and localizing objects.

\subsubsection{Watershed Segmentation}
Watershed segmentation is a classical region-based image segmentation technique that draws inspiration from geographical topography. The method visualizes an image as a topographic surface where the pixel intensity corresponds to elevation. The idea is to flood this surface from the lowest elevation points, treating bright regions as ridges and dark regions as valleys. The process involves filling the valleys (or basins) with water, and as the water rises, separate basins merge along the watershed lines, which are used to define object boundaries.

In biomedical imaging, watershed segmentation is particularly useful for separating overlapping or closely packed objects, such as touching cells in microscopy images \cite{b152}. For example, in histopathological images, where individual cells often cluster tightly together, watershed segmentation can isolate each cell by detecting the boundaries between them.

However, the algorithm is highly sensitive to noise and gradient variations, which can lead to over-segmentation. To mitigate these issues, pre-processing steps like filtering, edge detection, or the use of markers are commonly applied. Marker-controlled watershed segmentation is a common improvement, where markers are used to define regions of interest more accurately and limit the flooding process. This approach ensures that the algorithm is more robust against noise, improving its applicability in noisy biomedical datasets.

\subsubsection{Clustering-Based Segmentation (K-Means)}
Clustering-based segmentation techniques, such as \textit{k-means clustering}, are popular methods for partitioning an image into segments based on the similarity of pixel characteristics like intensity, color, or texture. The k-means algorithm works by dividing the pixels of an image into \(k\) clusters, where each pixel is assigned to the cluster whose mean value is closest to its intensity. The process iteratively refines the cluster assignments until the variance within each cluster is minimized.

In biomedical imaging, k-means clustering is used to separate distinct regions, such as tissues, cells, or other structures in medical scans. One key application is in magnetic resonance imaging (MRI) and computed tomography (CT) scans, where the algorithm can be used to differentiate between tissues with different intensity values, such as distinguishing tumors from surrounding healthy tissues. This method is highly effective in cases where the intensity contrast between different regions is significant \cite{b154}.

Although k-means clustering is efficient and easy to implement, it has limitations. It struggles with complex images where object boundaries are not well-defined or where there is significant overlap between the intensity values of different regions. Moreover, the number of clusters \(k\) must be pre-defined, which can be challenging in applications where the exact number of distinct regions is unknown. To address these limitations, adaptive versions of k-means, such as fuzzy c-means and hierarchical clustering, have been developed, which provide more flexibility in determining the number of clusters and handling overlapping regions.

\subsubsection{Region-Based Convolutional Neural Networks (R-CNN)}
Region-Based Convolutional Neural Networks (R-CNN) is a seminal object detection method that revolutionized the way object detection was performed by combining region proposals with CNN-based feature extraction. The R-CNN approach works by first generating several region proposals from an input image using selective search. Each region is then passed through a CNN to extract features, and these features are used to classify the region as containing an object or not, as well as to predict bounding boxes.

R-CNN was a significant improvement over traditional sliding window methods, which were computationally expensive. However, it still suffered from slow processing speed due to the need to run a CNN for each proposed region, making it less suitable for real-time applications \cite{b187}. In biomedical imaging, R-CNN has been used to detect and classify abnormalities in medical images, such as identifying tumors in MRI or CT scans, though its relatively slow speed limits its usage in real-time diagnostic applications.

\subsubsection{Faster R-CNN}
Faster R-CNN is an extension of the R-CNN family that addresses the slow speed of its predecessors by introducing the Region Proposal Network (RPN), which generates region proposals much more efficiently. In Faster R-CNN, the RPN shares convolutional features with the object detection network, allowing for faster and more accurate detection by reducing redundant computations \cite{b160, b188}. 

Faster R-CNN has become a popular choice for object detection in various fields, including biomedical imaging. Its ability to handle complex images with multiple objects makes it particularly well-suited for detecting multiple lesions or abnormalities in medical scans, such as detecting nodules in lung cancer screenings or identifying multiple cell types in pathology slides. Although faster than R-CNN, Faster R-CNN still struggles with real-time applications due to its relatively high computational cost compared to algorithms like YOLO.

\subsubsection{Mask R-CNN}
Mask R-CNN is an extension of Faster R-CNN designed to perform both object detection and instance segmentation, enabling the algorithm to detect objects and also generate a high-quality segmentation mask for each object. The method adds a third branch to the Faster R-CNN architecture, which predicts pixel-wise masks for each detected object in parallel with bounding box and class label predictions \cite{b172, b189}.

In biomedical imaging, Mask R-CNN is particularly useful for tasks that require precise segmentation of individual objects, such as cell or tumor segmentation in microscopy or radiology images. Its ability to simultaneously detect and segment objects makes it highly valuable in situations where object shape and boundary accuracy are critical, such as in the identification of cancerous lesions in medical images. However, like Faster R-CNN, Mask R-CNN requires significant computational resources, limiting its usage in real-time applications.

\subsubsection{You Only Look Once (YOLO)}
You Only Look Once (YOLO) is a cutting-edge, real-time object detection algorithm that has gained widespread popularity in various domains due to its efficiency and accuracy. Unlike traditional object detection systems, which perform detection by applying a sliding window or region proposal mechanism over an image, YOLO operates by dividing the image into a grid. Each grid cell is responsible for predicting bounding boxes, object classes, and confidence scores simultaneously, enabling the algorithm to detect objects in a single pass through the neural network.

YOLO is particularly suitable for real-time applications, such as video analysis, autonomous vehicles, and surveillance, where rapid object detection is essential \cite{b183}. Its ability to process images at high frame rates while maintaining accuracy makes it a popular choice in both research and industry. YOLO’s architecture, which is based on a single convolutional neural network, allows it to outperform other object detection algorithms like R-CNN and Fast R-CNN in terms of speed without compromising detection performance.

In biomedical imaging, YOLO is increasingly being used for tasks like detecting cancerous cells in histopathological images or identifying anomalies in radiology scans. One of YOLO's major advantages is its ability to detect multiple objects within a single image, making it particularly useful in detecting multiple biological entities in complex environments. However, while YOLO is fast and efficient, it may struggle with small objects or objects that are very close to each other, which can be problematic in certain biomedical applications where precision is paramount. To address these limitations, newer versions of YOLO, such as YOLOv4 and YOLOv5, have been developed, incorporating improvements in object detection accuracy and handling smaller objects.

\subsection{Object Classification}
Once objects are detected, the next step in the pipeline is object classification. It is the process of assigning detected objects to predefined categories based on their visual or structural characteristics. In the context of biomedical imaging, this could involve differentiating between cell types, identifying tissue regions, or categorizing pathological entities. It is a crucial step that adds context to the raw output of object detection systems, enabling a deeper understanding of the scene or data being analyzed.

\subsubsection{What is Object Classification?}
In object classification, detected objects are analyzed based on their features—such as shape, texture, size, or intensity—to determine their category. This process is central to many applications across domains, including medical diagnostics, where accurate classification can directly influence clinical decisions. By leveraging machine learning models, especially Convolutional Neural Networks (CNNs), classification systems can automatically learn complex patterns in the data that are difficult for traditional feature-based methods to capture.

In medical imaging, object classification might involve identifying healthy versus diseased cells, distinguishing between benign and malignant tumors, or classifying tissue types in histopathological samples. This capability is essential for automating tasks that would otherwise require manual interpretation by experts, speeding up the diagnostic process and improving consistency.

\subsubsection{Importance of Object Classification}
Object classification extends beyond merely detecting objects by providing the necessary context to interpret what each object represents. For instance, in medical applications, detecting a structure in an MRI scan is insufficient without knowing whether it is a healthy tissue or a tumor. Classification assigns meaningful labels to these detected regions, allowing for further analysis, such as tracking disease progression or planning treatment interventions.

In research and clinical settings, classification helps automate tedious tasks, such as counting and categorizing different types of cells in pathology slides or segmenting and labeling different organs in radiology images. This not only improves the efficiency of medical professionals but also enhances the accuracy of diagnosis by reducing human error.

\subsubsection{Significance in Object Tracking}
In object tracking, classification plays a critical role by linking detected objects across time, maintaining their identity as they move through different frames. This is particularly important in dynamic biomedical scenarios, where multiple objects may interact, merge, or split. Classification ensures that each object is consistently labeled, even when its appearance changes due to motion or interaction with other objects.

For example, in cell tracking, classification can help differentiate between different cell types or states (e.g., cancerous versus non-cancerous) as they move and divide in microscopy videos. This capability is essential in fields such as developmental biology or cancer research, where tracking specific cell populations over time provides insights into cell behavior, disease mechanisms, or drug responses.

Moreover, in complex multi-object tracking scenarios, classification prevents identity swaps between objects of different categories, ensuring that the system tracks not only the position but also the specific identity of each object throughout the video sequence.

\subsubsection{How Object Classification Works in Biomedical Applications}
Object classification in biomedical imaging relies heavily on machine learning, particularly deep learning approaches such as CNNs. These models automatically learn features from raw image data that are crucial for distinguishing between various categories, such as healthy versus diseased cells, or different types of tissues in radiological scans.

CNNs are especially effective because they can learn hierarchical representations—starting from basic features like edges and textures to more complex patterns that capture the unique characteristics of different biological entities. This allows CNNs to outperform traditional methods that rely on handcrafted features, especially when dealing with complex and noisy biomedical data.

In applications like histopathology, CNNs have been used to classify cancerous cells from normal cells in tissue samples, aiding pathologists in diagnosing cancer and assessing its progression \cite{b164, b163}. In radiology, classification models are used to distinguish between different tissue types, identify tumors, and detect anomalies in MRI or CT scans. The ability to classify and localize these abnormalities accurately is critical for early disease detection and treatment planning.

\textbf{Transfer Learning:} A significant advancement in biomedical image classification is the use of transfer learning, which involves fine-tuning pre-trained models (such as those trained on large datasets like ImageNet) for specific medical tasks. This technique is particularly useful when labeled biomedical data is scarce, as it allows models to adapt to new tasks with minimal training data. Transfer learning has been successfully applied in diagnosing rare diseases and classifying highly specialized medical images where obtaining large, labeled datasets is challenging.

\textbf{Explainability in Biomedical AI:} Beyond accuracy, explainability is key in biomedical applications. Clinicians need to understand why a model has classified an object in a certain way to ensure trust in AI-driven diagnostics. Techniques like Class Activation Mapping (CAM) and Grad-CAM enable visual explanations of which regions in the image contributed to the classification decision, helping experts interpret the model's predictions. This is especially important in healthcare, where model transparency is crucial for clinical adoption and patient safety.

In summary, object classification is an essential step in object detection and tracking, providing the contextual information required for more meaningful analysis. With the advancements in machine learning and deep learning, classification models have become highly effective in biomedical applications, enabling more accurate and automated diagnosis, treatment planning, and research insights.

\begin{figure}
\centerline{\includegraphics[width=.5\textwidth]{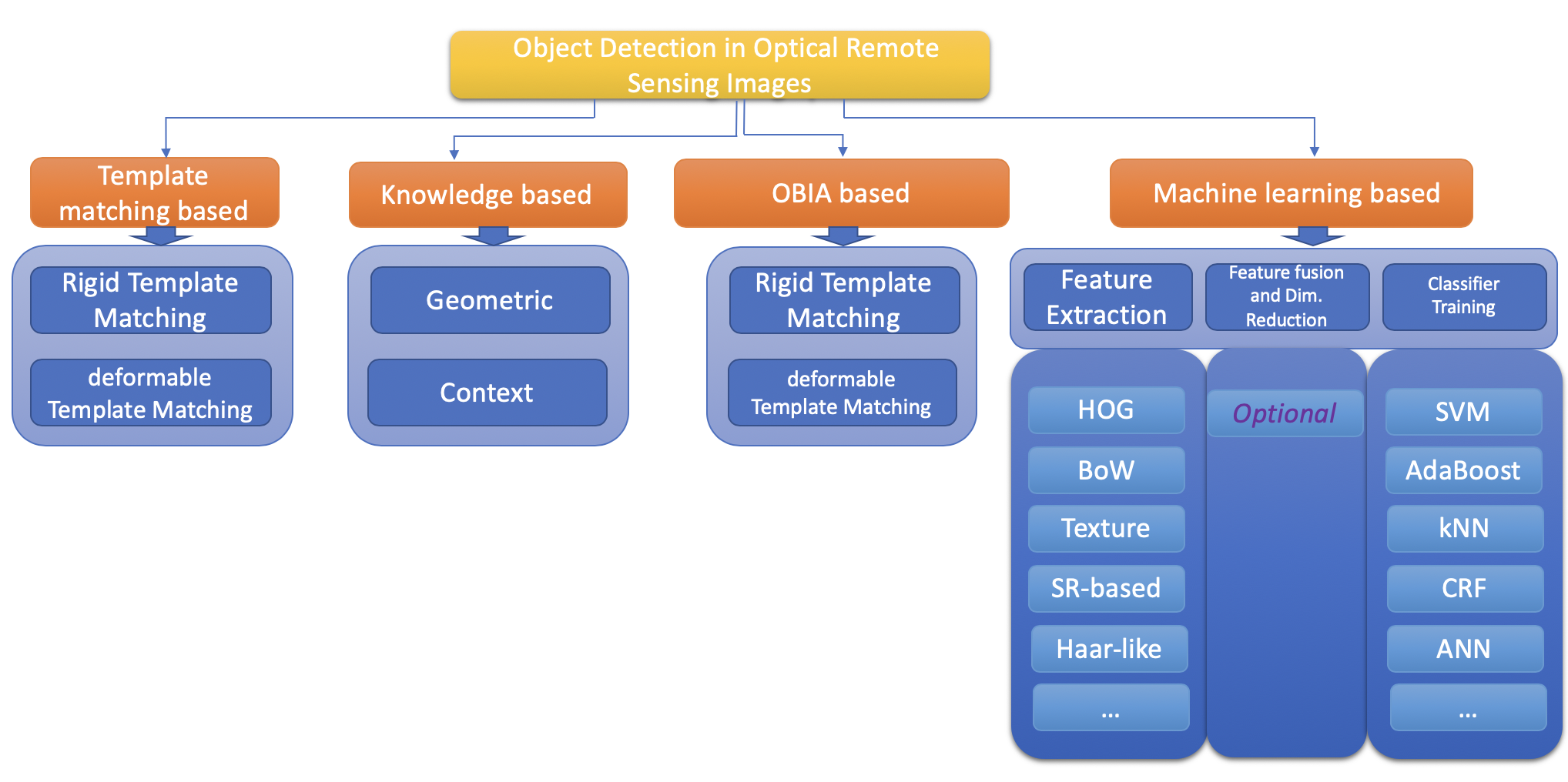}}
\caption{\textit{Cheng’s Taxonomy of methods for object detection in optical RSIs. The components are discussed in his paper \cite{b46}.}}
\label{fig:1-2}
\end{figure}

\
\section{Taxonomy of Object Tracking Methods}

Object tracking is a fundamental problem in computer vision, encompassing the processes of detecting, classifying, and continuously monitoring objects as they move through a sequence of video frames. The importance of accurate and reliable object tracking spans a wide range of applications, including surveillance, autonomous driving, human-computer interaction, and medical imaging. Given the diversity of tracking scenarios and the rapid evolution of technology, numerous methodologies have been developed, each with its own strengths and limitations. 

To systematically understand and evaluate these methodologies, we categorize existing object tracking methods into four primary groups:

\begin{itemize}
    \item \textit{Conventional and Classic Methods}
    \item \textit{Feature-based Tracking Models}
    \item \textit{Probabilistic and Statistical Methods}
    \item \textit{Machine Learning and Deep Learning-based Methods}
\end{itemize}

Each of these categories represents a distinct approach to tackling the challenges of object tracking, reflecting different historical periods, technological advancements, and underlying theoretical foundations. 
\subsection{Taxonomy of Object Tracking Methods}

Object tracking methods can be classified into four major categories, each with distinct methodologies and use cases. The following table highlights the detailed strengths and shortcomings of these categories along with examples of their application.


\setlength{\tabcolsep}{2pt} 
\renewcommand{\arraystretch}{1.4} 

\begin{table*}[htbp]
\caption{Strengths and Shortcomings of Object Tracking Methods}
\label{tab:tracking_taxonomy_extended}
\begin{tabularx}{\textwidth}{@{}l*{10}{C}c@{}}
\toprule

\hline
\textbf{Category}  & \textbf{Strengths}                                                                                 & \textbf{Shortcomings}                                                                                          & \textbf{Examples}                                                          \\ \hline

\textbf{Conventional Methods} & 
- Simple and computationally efficient.  \newline
- Useful for real-time applications with limited computational resources. \newline
- Perform well in controlled environments with static backgrounds and minimal occlusions.
& 
- Struggle with complex scenarios involving occlusions, background clutter, scale variation, and non-rigid object deformation.  \newline
- Lack robustness in dynamic or cluttered scenes.
& 
Edge detection (Canny, Sobel), Template Matching, Background Subtraction \cite{b12, b40}.   \\ \hline

\textbf{Feature-based Tracking Models} & 
- Robust against object deformation, scale variations, and rotation. \newline
- Effective in dynamic environments with motion and partial occlusions.  \newline
- Tracks reliable features invariant to transformations.
& 
- High computational cost for high-resolution data.  \newline
- Performance degrades when objects lack distinctive features or in environments with lighting or appearance changes.
& 
SIFT, SURF, KLT, Optical Flow, Dense and Sparse Trajectories \cite{b22, b23, b210, b174}.  \\ \hline

\textbf{Probabilistic and Statistical Methods} & 
- Handle noise and uncertainty effectively, ensuring continuous tracking even in challenging scenarios.  \newline
- Predict object locations during occlusion or temporary absence.
& 
- Limited to handling non-linear or unpredictable motions without advanced filtering.  \newline
- Performance depends on accurate motion models, and errors in models can lead to drift or tracking failure.
& 
Kalman Filter, Particle Filter, Hidden Markov Models, Gaussian Mixture Models (GMM) \cite{b123, b124, b125, b217}. \\ \hline

\textbf{Machine Learning and} \\ \newline 
\textbf{Deep Learning Methods} & 
- Highly accurate in challenging environments with cluttered backgrounds, deformable objects, and dynamic lighting.  \newline
- Track multiple objects simultaneously, handling occlusion, object interactions, and scale variations effectively.  \newline
- Benefit from large datasets and powerful CNN/RNN models to learn robust tracking strategies.
& 
- Require large labeled datasets and extensive computational resources for training.  \newline
- May struggle with transparency, making it hard to interpret errors.  \newline
- Risk of overfitting to specific datasets.
& 
Faster R-CNN, YOLO, DeepSORT, ROLO, Mask R-CNN, MDNet, Siamese Networks \cite{b160, b70, b157, b222}. \\ \hline
 \bottomrule

\end{tabularx}
\end{table*}

\textbf{Conventional and Classic Methods} typically rely on traditional image processing techniques, such as template matching, background subtraction, and contour detection. These methods often involve minimal learning and are based on well-established principles from early computer vision research. While these approaches can be effective in controlled environments, they often struggle with complex scenarios involving occlusions, varying lighting conditions, or significant object deformation. Despite these limitations, conventional methods are still widely used in applications where computational efficiency is critical, or where modern machine learning techniques may not be feasible.

\textbf{Feature-based Tracking Models} focus on identifying and tracking salient features of objects, such as corners, edges, or texture patterns. These methods are designed to handle the variability in object appearance by relying on stable and distinctive features that can be consistently detected across frames. Techniques such as optical flow and the use of keypoint descriptors (e.g., SIFT, SURF) fall into this category. Feature-based tracking is particularly useful in dynamic environments where objects undergo significant motion, as it allows for robust tracking even in the presence of noise and partial occlusions.

\textbf{Probabilistic and Statistical Methods} introduce a probabilistic framework to the tracking problem, often leveraging statistical models to predict the future position of objects based on their past trajectories. Methods such as the Kalman filter, particle filter, and hidden Markov models exemplify this category. These approaches are well-suited for scenarios where uncertainty and noise play a significant role, as they provide a mechanism for filtering out irrelevant information and making predictions based on probabilistic inference. The flexibility and robustness of these methods make them applicable in various domains, from robotics to financial modeling.

\textbf{Machine Learning and Deep Learning-based Methods} represent the state-of-the-art in object tracking, employing powerful learning algorithms to model complex patterns and relationships in data. These methods leverage large datasets and sophisticated neural network architectures, such as convolutional neural networks (CNNs) and recurrent neural networks (RNNs), to learn feature representations and tracking strategies directly from data. Deep learning-based trackers have achieved remarkable success in recent years, outperforming traditional methods in challenging scenarios involving cluttered backgrounds, deformable objects, and real-time processing requirements. The adaptability and scalability of these models have opened new frontiers in applications such as autonomous driving, where real-time decision-making is critical.

In the following sections, we embark on a comprehensive exploration of each of these categories, providing an in-depth analysis of their foundational methodologies, practical applications, and performance metrics. Each category represents a unique approach to the multifaceted challenge of object tracking, and by examining them individually, we aim to offer a nuanced understanding of their respective strengths and limitations.

We begin by discussing the underlying methodologies that form the backbone of these tracking approaches. This includes an exploration of the theoretical principles that guide the design and implementation of each method, as well as the specific algorithms and techniques that are commonly employed. By grounding our discussion in the fundamental concepts, we aim to provide a solid foundation for understanding the intricacies of each approach.

Following this, we delve into the practical applications of these methods across various domains. Object tracking has broad relevance, from autonomous vehicles navigating complex environments to medical imaging systems tracking cellular structures. By highlighting the diverse applications, we underscore the versatility and importance of these tracking techniques in solving real-world problems.

Performance is a critical aspect of object tracking, and our analysis includes a thorough evaluation of each method's performance across different scenarios. We consider factors such as accuracy in tracking moving objects, the computational complexity of the algorithms, their robustness against occlusion and noise, and their adaptability to dynamic changes in the environment. These criteria, outlined in the introduction, serve as a benchmark for comparing the efficacy of each approach in practical settings.

One of the key focal points of our review is the integration of deep learning architectures into modern tracking systems. Deep learning has revolutionized the field of computer vision, offering unprecedented capabilities in feature extraction, object recognition, and real-time tracking. We pay particular attention to how these advanced architectures have been incorporated into traditional tracking methods, enhancing their accuracy, scalability, and robustness. This section also highlights recent innovations that leverage neural networks to address some of the long-standing challenges in object tracking, such as dealing with occlusions, handling multiple objects simultaneously, and adapting to varying object appearances.

While previous taxonomies have provided valuable frameworks for classifying object tracking techniques \cite{b6, b46, b47, b12}, our review seeks to go beyond these by offering a fresh perspective that reflects the latest developments in the field. Specifically, we focus on the emerging applications of object tracking in biological and biomedical fields, where the demand for precise and efficient tracking solutions has driven significant advancements. These applications include tracking the movement of cells in time-lapse microscopy, monitoring the progression of diseases, and aiding in the development of personalized medical treatments. By incorporating these recent trends, our review aims to bridge the gap between traditional object tracking methods and the cutting-edge technologies that are shaping the future of the field.

Through this detailed examination, we hope to provide researchers, practitioners, and enthusiasts with a comprehensive understanding of the current state of object tracking, its challenges, and the exciting opportunities that lie ahead. As we move forward, each section will build upon the previous one, offering a coherent narrative that guides the reader through the complex and evolving landscape of object tracking technologies.

\subsection{Conventional Methods in Object Tracking}

Conventional object tracking methods are rooted in classical image processing techniques such as contour extraction, edge detection, and template matching. These methods typically follow deterministic algorithms, relying on predefined rules rather than learning from data. Although simpler than modern machine learning-based techniques, conventional methods have proven to be robust and efficient, especially in environments where computational resources are constrained.

These methods extract explicit image features—like edges and contours—which are then used to identify and track objects across frames. Edge detection techniques, such as the Canny and Sobel operators, highlight the prominent edges within an image, making it possible to track these edges based on how they evolve over time. Additionally, template matching algorithms, such as Sum of Absolute Differences (SAD) or Normalized Cross-Correlation (NCC), compare image patches between consecutive frames, enabling the tracker to follow objects based on their visual appearance.

Despite their simplicity, conventional methods perform effectively in controlled environments where backgrounds remain static, and objects exhibit consistent features. However, they face challenges in complex scenarios that involve occlusion, scale variation, or background clutter. For instance, when objects are partially or fully occluded, these methods may lose track of them. Similarly, scale variation poses a challenge for fixed-scale models, making it difficult to adapt to objects that change in size.

To mitigate these challenges, optimizations like adaptive edge detection and dynamic template matching have been introduced. Furthermore, integrating probabilistic models such as Kalman filters has enhanced the robustness of conventional methods, enabling them to handle noise and uncertainty effectively.

In summary, while conventional methods lack the adaptability of machine learning approaches, they remain valuable for their simplicity and efficiency, particularly in resource-constrained environments. Ongoing improvements ensure their continued relevance in specific object tracking applications.

\subsubsection{Color-based Tracking}

Color-based tracking is one of the most intuitive and commonly used methods in object tracking, leveraging the color information in images to track objects as they move across frames. Known as color histogram-based tracking, this approach uses the distribution of colors within a target region as a distinctive signature for tracking.

The work of McKenna \textit{et al.} pioneered the use of color histograms in face tracking, demonstrating their effectiveness in distinguishing and following faces in video sequences \cite{b1}. Since then, color-based tracking has expanded to a variety of objects beyond facial tracking.

The tracking process begins with background subtraction to isolate the target object. A color histogram representing the distribution of color intensities is computed for the pixels in the target region. This histogram serves as the object’s identifier. As the object moves across frames, the histogram is continuously updated and compared to the previous frame’s histogram to maintain tracking.

A more sophisticated variant employs Gaussian Mixture Models (GMM) to model the background, which enhances object detection based on hue and saturation from the HSV color space \cite{b2}. The HSV color space is preferred over RGB because it separates chromatic content (hue) from intensity (value), making the tracking process more robust to changes in lighting conditions.

\begin{equation}
P(x) = \sum_{i=1}^{n} \frac{w_i}{\sqrt{(2\pi)^d |\Sigma_i|}} \exp\left(-\frac{1}{2}(x - \mu_i)^T \Sigma_i^{-1} (x - \mu_i)\right)
\end{equation}

In this equation, \( \mu_i \), \( \Sigma_i \), and \( w_i \) denote the mean, covariance, and weight of the \( i \)-th Gaussian component, respectively. The model captures the color distribution of the background, making it easier to distinguish the target object by its unique color histogram.

While color-based tracking is computationally efficient and intuitive, it struggles in scenarios where objects share similar colors or textures. In such cases, it becomes challenging for the tracker to distinguish between objects, especially when they overlap or come into close proximity. Recent advancements, such as adaptive histogram updating and the integration of spatial information, have improved the performance of color-based tracking under these conditions. Additionally, deep learning techniques, particularly convolutional neural networks (CNNs), have been employed to extract more discriminative color features, further enhancing robustness to lighting changes and occlusion.

\begin{figure}
   \centerline{\includegraphics[width=0.5\textwidth]{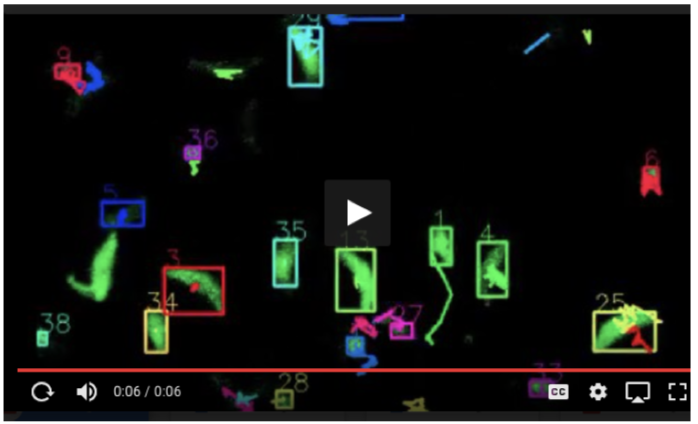}}
\caption{\textit{Contour-based tracking applied on} \textit{T. gondii} \textit{cell-tracking by setting a bounding box around the detected contours of the cells.}}
\label{fig:3}
\end{figure}

In summary, color-based tracking remains a fundamental technique for object tracking, appreciated for its simplicity and efficiency. Ongoing research continues to refine its capabilities, ensuring its relevance in more dynamic and complex environments.

\subsubsection{Kernel-based Tracking}

Kernel-based tracking methods, often implemented through the mean-shift algorithm, focus on representing and localizing objects. These methods utilize an isotropic kernel function to create a spatial representation of the target, transforming the localization problem into an iterative search process. The goal is to maximize the similarity between the current target model and candidate regions in each new frame.

First introduced in the late 1990s, kernel-based models emerged as efficient solutions for tracking non-rigid objects. The mean-shift algorithm, a key element in this framework, uses similarity functions like the Bhattacharyya coefficient to measure the overlap between the target model and the candidate region \cite{b240}. The algorithm iteratively shifts the kernel toward regions of maximum density in the feature space, effectively localizing the target.

To improve tracking performance in dynamic environments, kernel-based methods are often integrated with motion filters, such as the Kalman filter and particle filter. These filters predict the target’s future position based on its past trajectory, helping maintain continuity in cases of occlusion or appearance changes. Additionally, data association techniques are used to match predicted locations with actual observations, further enhancing accuracy.

Recent advancements in kernel-based tracking include the development of adaptive kernels that adjust their shape, size, and orientation in response to changes in the object’s appearance. This adaptability enables the tracker to handle target deformations, scale variations, and partial occlusions. Hybrid models combining kernel-based tracking with advanced background modeling techniques have also been developed, leading to improved performance in cluttered environments.

\begin{figure*}[htbp]
\centerline{\includegraphics[width=\textwidth]{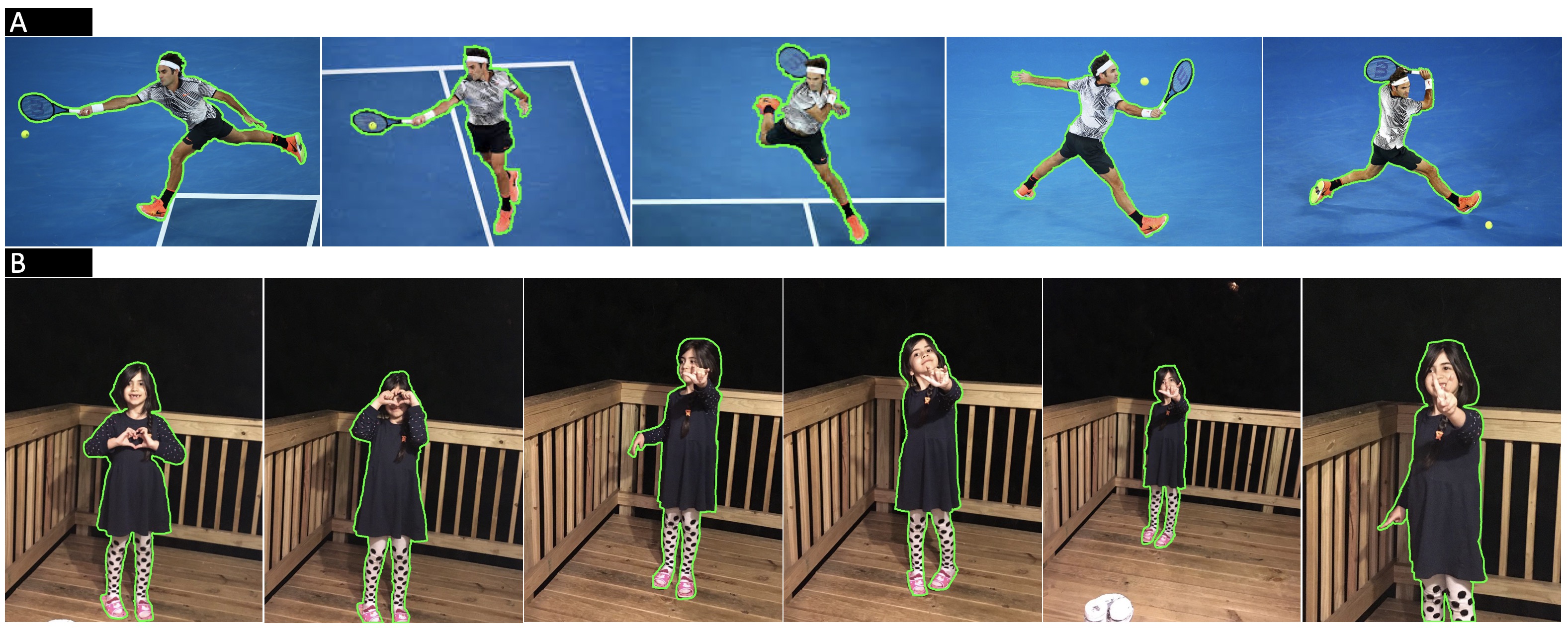}}
\caption{\textit{Contour-based tracking results in different video sequences. (A) Tennis player and (B) a girl playing on the balcony.}}
\label{fig:4-1}
\end{figure*}

In summary, kernel-based tracking offers a balance between computational efficiency and tracking accuracy, making it suitable for real-time applications. Its continued evolution ensures its relevance in increasingly complex tracking scenarios.

\subsubsection{Contour-based Tracking}

Contour-based tracking focuses on enhancing object contours to improve tracking accuracy. This method begins with preprocessing techniques, such as histogram equalization, dilation, and erosion, which enhance object boundaries and reduce background noise. These contours are then extracted to define bounding boxes that enclose the objects of interest, and tracking is performed by following the centroids of these bounding boxes across frames \cite{b5}.

Though straightforward, contour-based tracking struggles with occlusion and dynamic environments where objects change shape. To address these limitations, Bayesian frameworks have been introduced, allowing for a more probabilistic understanding of contour features. These frameworks incorporate both contour and shape energy features, improving robustness to occlusion.

The integration of deep learning into contour-based tracking has further advanced the field. Convolutional neural networks (CNNs) are now used to predict contour maps with greater accuracy, even in low-contrast or cluttered scenes. This has led to significant improvements in tracking performance, especially in complex real-world environments.

In summary, contour-based tracking remains a valuable method for object tracking, particularly in cases where precision and object boundary detection are critical. Recent advancements in machine learning have significantly enhanced its capabilities, ensuring it remains a competitive choice in object tracking.

\subsection{Feature-based Tracking Models}
Feature-based tracking models have become a cornerstone in object tracking due to their robustness and adaptability. Unlike contour-based methods, which rely on boundary information, feature-based methods emphasize the extraction and tracking of distinct, salient features such as corners, edges, and textures. These features are generally invariant to transformations and occlusions, making them ideal for long-term tracking across various domains \cite{b19}.

Prominent examples of feature extraction techniques include the Scale-Invariant Feature Transform (SIFT), Speeded-Up Robust Features (SURF), and optical flow-based methods. These techniques are essential in dynamic scenes where objects may rotate, change scale, or experience occlusions, as they provide scale, rotation, and illumination invariance.

\subsubsection{Scale-Invariant Feature Transform (SIFT)}
The Scale-Invariant Feature Transform (SIFT) is a widely used algorithm in feature-based tracking due to its ability to detect and describe local features that remain robust to scale, rotation, and partial occlusions. SIFT works by identifying keypoints in an image—points of interest that exhibit strong contrast—and describing their neighborhood using a scale-invariant descriptor \cite{b22}. This descriptor encodes information about the gradients around each keypoint, which makes SIFT particularly effective in scenes where objects may change in size or orientation.

SIFT’s resilience to variations in scale and rotation makes it suitable for applications such as object recognition, image stitching, and biomedical imaging. In time-lapse microscopy, for instance, SIFT can track cellular structures as they grow, divide, or deform over time. Its ability to maintain accuracy in challenging conditions like varying illumination or occlusion is particularly valuable in fields like biological research and medical diagnostics\cite{b23}.

\subsubsection{Speeded-Up Robust Features (SURF)}
The Speeded-Up Robust Features (SURF) algorithm is an evolution of SIFT designed to enhance computational efficiency without sacrificing robustness. SURF employs an integral image representation and a fast Hessian-based detector, which makes it significantly faster than SIFT while retaining robustness to scale, rotation, and noise \cite{b23}. The SURF descriptor is built upon wavelet responses, which allows for quick computation and strong performance in real-time applications.

SURF is commonly used in scenarios where real-time performance is critical, such as robotics, surveillance, and augmented reality. In biomedical contexts, SURF has been applied to track cells in time-lapse videos, enabling the analysis of fast-moving or rapidly changing biological entities. Its speed and ability to handle scale variations make it a popular choice for processing large-scale video data in biological research\cite{b24}.

\subsubsection{Optical Flow}
Optical flow refers to the pattern of apparent motion of objects in a video sequence, caused by the relative movement between the camera and the objects. This method computes the displacement of pixels or regions between consecutive frames to estimate the direction and speed of motion. Optical flow is widely used in feature-based tracking as it provides temporal coherence, making it possible to track objects over time by analyzing their motion \cite{b210}.

Optical flow can be computed using various methods, but two primary categories are \textit{sparse optical flow} and \textit{dense optical flow}. Sparse optical flow focuses on tracking specific keypoints or regions in the image, while dense optical flow computes motion vectors for all pixels, offering a more comprehensive motion analysis.

\subsubsection{Sparse Trajectories}
Sparse trajectories refer to the tracking of a selected subset of points—typically feature-rich regions such as corners or keypoints—over time in a video sequence. Sparse trajectory methods, like KLT, are computationally efficient because they focus only on salient features rather than processing every pixel in the image \cite{b210}. This selective tracking allows for real-time analysis in applications where computational resources are limited or where it is essential to track only the most relevant features.

Sparse trajectories are widely used in action recognition, motion analysis, and surveillance. In biomedical imaging, they are particularly useful for tracking the motion of specific cellular structures, allowing researchers to monitor individual cell behaviors without the computational overhead of dense tracking\cite{b270}.

\subsubsection{Dense Trajectories}
In contrast to sparse trajectories, dense trajectory methods compute the motion of densely sampled points across video frames, capturing detailed motion patterns across the entire image. Dense optical flow is often employed to generate dense trajectories, providing a comprehensive analysis of the motion of all objects in the scene, including background and smaller, subtler movements that sparse methods might miss \cite{b174}.

Dense trajectories are particularly effective in capturing complex motions in dynamic environments, making them suitable for applications such as human action recognition and animal behavior analysis. In biomedical research, dense trajectories have been applied to track the movement of pathogens, such as \textit{T. gondii}, across microscopy videos, providing detailed insights into their behavior over time. Dense trajectory methods can handle large datasets and are capable of tracking hundreds of features simultaneously, making them ideal for studies involving large-scale motion analysis\cite{b174}.

\subsubsection{Kanade-Lucas-Tomasi (KLT) Method}

The Kanade-Lucas-Tomasi (KLT) method is a widely adopted feature-based tracking algorithm that focuses on tracking distinctive points in video sequences. Initially developed as an extension of the optical flow technique proposed by Lucas and Kanade in 1981 \cite{b19}, KLT further improves the robustness of feature tracking by introducing a selection criterion that ensures the tracking of only the most prominent and reliable features.

The core idea behind KLT is the local optimization of an image patch surrounding a feature point, which tracks the displacement of that feature between consecutive frames by minimizing the sum of squared differences (SSD) in intensity values. The KLT tracker uses a pyramidal implementation of the Lucas-Kanade optical flow method, which allows for efficient tracking of feature points at multiple scales, ensuring better performance when objects or features move at varying speeds or change in size.

KLT is particularly effective for \textit{sparse optical flow}
applications, where only a subset of distinctive points—such as corners or edges—are tracked across frames. Instead of tracking all pixels or features in an image, the KLT method focuses on keypoints that exhibit high contrast, thus making it more computationally efficient and well-suited for real-time applications. The method is also robust in handling moderate levels of deformation, lighting changes, and noise, which makes it highly effective in dynamic environments.

\textbf{KLT Feature Selection: The Tomasi Improvement}\\
The improvement introduced by Tomasi and Kanade in 1991 \cite{b210} focuses on how to select the best features to track, ensuring the accuracy and stability of the tracking process. In their enhancement, a feature is considered "good" if it can be tracked reliably. This is determined by examining the eigenvalues of the gradient matrix in a small region around each candidate point. The eigenvalues correspond to the directional intensity changes at that point, and only points with large eigenvalues in both directions (i.e., corners) are considered good for tracking.

The Tomasi feature selection ensures that only well-defined features, which are resistant to drifting over time, are tracked, making KLT more robust in long-term tracking applications. This improvement significantly reduces errors caused by selecting ambiguous features that might disappear or deform quickly.

\textbf{Advantages of KLT in Real-time and Low-Resource Scenarios}\\
The computational efficiency of KLT, particularly in its sparse optical flow variant, makes it highly suitable for \textit{real-time applications}. This efficiency stems from the selective tracking of high-interest points rather than attempting to process the entire image. As a result, KLT has become a favored choice in fields where quick responses are essential, such as robotics, augmented reality, and real-time surveillance systems.

In addition, KLT is valuable in resource-constrained environments, such as embedded systems, where computational power and memory are limited. Its ability to focus on a small subset of key points allows for effective tracking with minimal computational overhead, which is particularly useful in mobile devices, drones, and small robotic systems that require efficient, real-time tracking capabilities without the need for high-end hardware\cite{b19, b44, b83}.

\begin{figure}
\centerline{\includegraphics[width=0.4\textwidth]{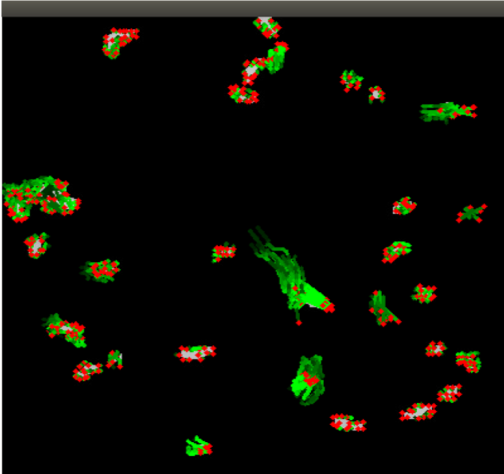}}
\caption{\textit{KLT feature tracking applied on biological cells, highlighting the movement of distinct points in a time-lapse video. The green lines represent the trajectories of the tracked features.}}
\label{fig:KLT-bio}
\end{figure}

\begin{figure*}[htbp]
\centerline{\includegraphics[width=.7\textwidth]{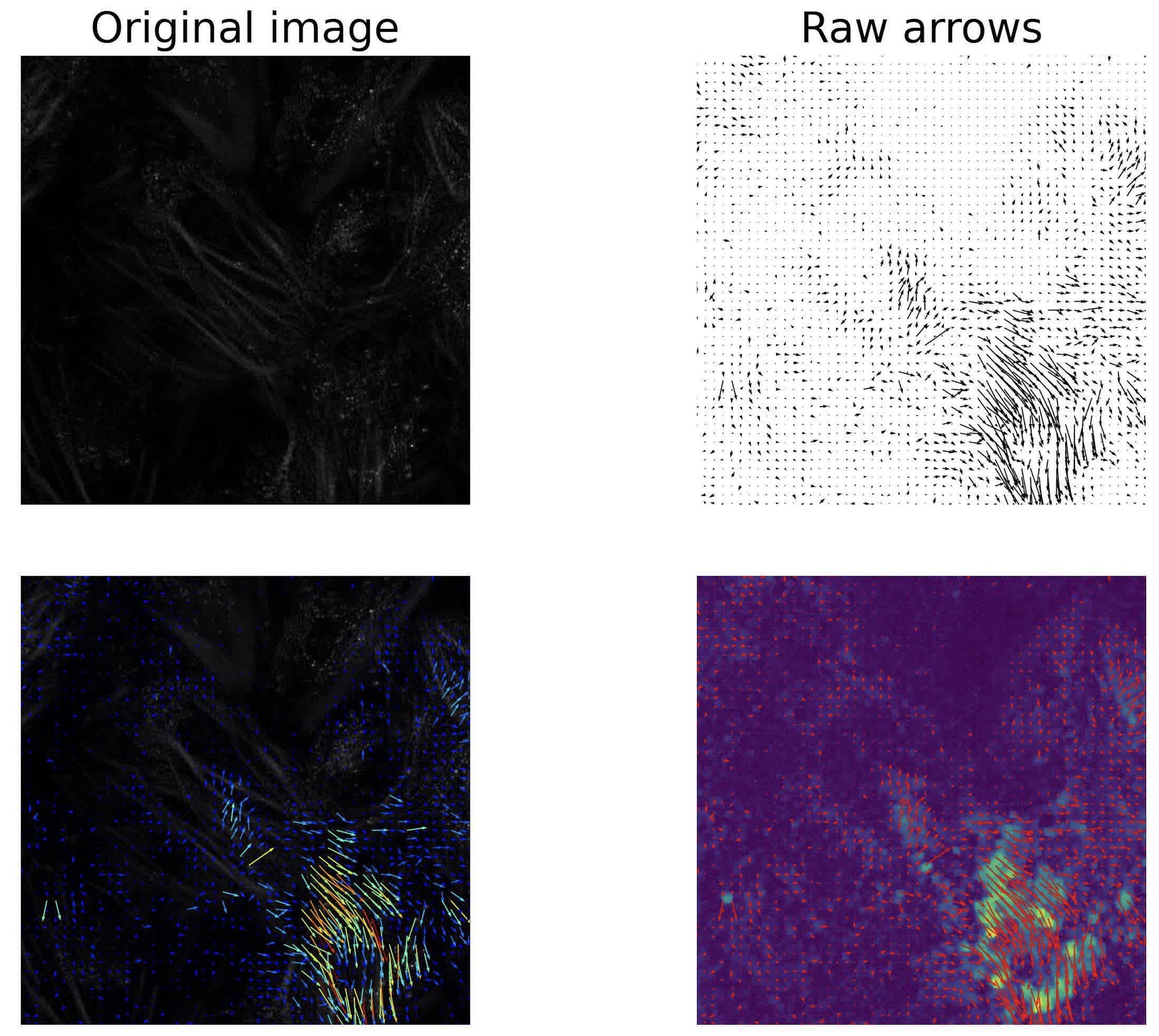}}
\caption{\textit{The figure illustrates the analysis of a myocardial beating video using optical flow techniques. The first sub-image shows the original video frame, providing a baseline visual representation of the myocardium. The second sub-image depicts the optical flow arrows on the frame, capturing the pixel-level motion dynamics. In the third sub-image, arrows represent the direction of motion, color-coded based on the norm (magnitude) of the movement, highlighting regions of varying motion intensity. The fourth sub-image visualizes the magnitude of the optical flow, showcasing the distribution of motion across the myocardium during its beating cycle..}}
\label{fig:optic-bio}
\end{figure*}

\textbf{KLT in Biomedical Applications}\\
KLT has found significant use in \textit{biomedical imaging}, particularly in tracking cellular structures in time-lapse microscopy. In these applications, individual cells or subcellular organelles are often the target of interest. The movement of cells or the internal dynamics of organelles can reveal critical biological processes, such as cell division, migration, or drug response. KLT allows researchers to track the movement of specific points on or within cells, enabling detailed analysis of cell behavior over time.

For example, in tracking the motion of pathogens like \textit{T. gondii} across time-lapse videos, KLT’s ability to track sparse, distinctive points is especially valuable. Its low computational cost and accuracy allow researchers to follow complex cell dynamics over extended periods without the need for expensive computational resources. Additionally, KLT is useful in monitoring tissue deformation during medical procedures or analyzing motion patterns in biomechanical systems.

\textbf{Limitations and Solutions in KLT}\\
Despite its many advantages, KLT has some limitations. One of the primary issues is that it can struggle with large-scale motion or significant occlusion, where feature points may become untrackable. If a feature moves too quickly between frames or becomes occluded by another object, the tracker may lose track of the feature altogether. Moreover, KLT assumes that the appearance of the feature remains relatively constant over time, making it less suitable for objects undergoing significant shape or texture changes.

To address these limitations, KLT is often combined with \textit{motion models} or \textit{predictive filters}. The \textit{Kalman filter} and \textit{particle filter} are frequently used in conjunction with KLT to predict the future positions of tracked features based on their previous trajectories. These filters help maintain tracking continuity even when features are momentarily occluded or undergo rapid movement. Additionally, hierarchical or pyramidal implementations of KLT allow for more robust performance in scenarios with varying scales or motions, as they compute optical flow at different resolution levels\cite{b19, b44, b83}.

\textbf{Extensions of KLT: Real-World Applications}\\
In real-world applications, KLT has been used in various fields beyond biomedical imaging. In \textit{computer vision} and \textit{augmented reality (AR)}, KLT is a key component in tracking objects or users’ motions in real-time. For example, in AR systems, KLT tracks the movement of users’ hands or faces, enabling the system to overlay virtual elements on top of real-world features with high accuracy.
\begin{figure}
\centerline{\includegraphics[width=0.4\textwidth]{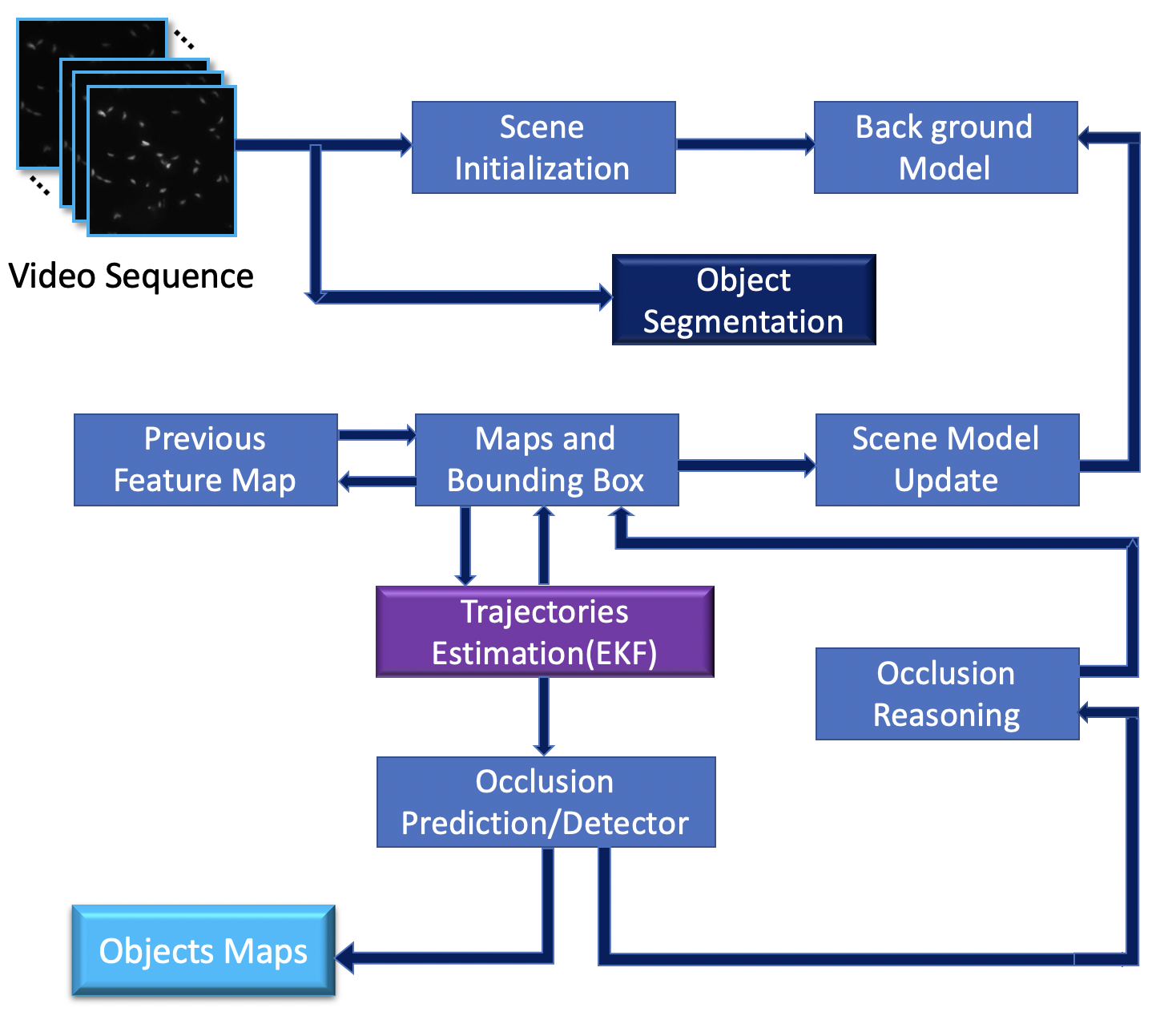}}
\caption{\textit{Extended Kalman Filters strategy flowchart for tracking maps and trajectory extraction.}}
\label{fig:KLT-ext}
\end{figure}
In \textit{surveillance systems}, KLT is employed for tracking people or vehicles in video feeds. Its ability to track specific keypoints allows for efficient monitoring without the computational cost of processing entire images, making it a practical solution for large-scale, multi-camera surveillance networks.

In \textit{robotics}, KLT is used for visual odometry and navigation. By tracking keypoints in the environment, robots can estimate their own motion and position relative to their surroundings. The efficiency of KLT makes it ideal for small, mobile robots that require real-time feedback for navigation and obstacle avoidance.\cite{b19, b44, b83}

\subsection{Probabilistic and Statistical Methods}
Object tracking often encounters challenges due to the dynamic nature of the objects being tracked. These challenges arise primarily from three factors: variability in object appearance due to changes in pose or deformation, fluctuations in illumination, and partial or complete occlusion. Each of these factors can lead to the loss of the target object during tracking. To address these issues, various analytical methods have been developed, particularly within the framework of probabilistic and statistical approaches. These methods often treat the tracking problem as an optimization challenge, seeking to predict and correct the object's path as conditions change.

\subsubsection{Parametric Models}
One significant contribution to the field comes from Hager and Belhumeur \cite{b50}, who introduced parametric models that account for geometric and illumination variations in region tracking. Their approach integrates changes in object geometry, such as pose and shape rotation, with fluctuations in illumination, enabling the tracking of large image regions. To tackle occlusion, they enhanced their method by incorporating statistical techniques that identify occluded regions as statistical outliers, thus maintaining tracking accuracy even under challenging conditions.

In the domain of traffic scene analysis, parametric model-based tracking has also shown promise. Studies by Koller et al. \cite{b51, b52} applied these models to track vehicles and other objects in complex traffic environments, demonstrating the robustness of parameterized approaches in real-world scenarios. Additionally, the work of Toyama and Blake \cite{b54} introduced a metric mixture model that combines the strengths of exemplar-based models with a probabilistic pipeline, leveraging Markov random field models to enhance tracking performance.

Another innovative approach was proposed by Sigal et al. \cite{b56}, who developed a graphical model based on \textit{Non-parametric Belief Propagation (NBP)} to address the challenges of 3D human tracking. Their method represents the human body as a loose-limbed model, where limbs are treated independently. While this approach offers flexibility, it also introduces complexity in terms of computation time and model accuracy. Despite these limitations, their work marked a significant advancement in tracking human movement in three-dimensional space.

These parametric models, by integrating geometric, photometric, and statistical considerations, offer a robust framework for addressing the inherent challenges of object tracking in dynamic environments. As tracking technology continues to evolve, these models will likely serve as a foundation for further innovations, particularly in scenarios where precision and adaptability are critical.

\subsubsection{Kalman Filter}
The Kalman Filter, introduced by R. E. Kalman in 1960 \cite{b123}, is a powerful recursive algorithm that provides optimal estimates of the internal state of a linear dynamical system from a series of noisy measurements. Since its inception, the Kalman Filter has found widespread applications across various domains, particularly in signal processing, control systems, robotics, and computer vision, including object tracking.

In the context of object tracking, the Kalman Filter plays a crucial role in predicting and updating the state of a moving object over time. This capability is essential for scenarios where the object's motion is subject to uncertainty due to factors like noise, occlusions, or changes in dynamics. The Kalman Filter excels in these situations by efficiently estimating the object's position and velocity, even when only partial or noisy observations are available.

The Kalman Filter operates through two main phases: prediction and update. These phases are mathematically described by the following equations:

\paragraph{Prediction Phase:} In this phase, the Kalman Filter predicts the state of the system at the next time step based on the current state and the known system dynamics. The prediction equation is given by:

\begin{equation}\label{2}
X_{(k+1)} = F_{(k)}X_{(k)} + v_{(k)},
\end{equation}

\noindent where $X_{(k+1)}$ represents the predicted state (e.g., position, velocity) at time step $k+1$, $F_{(k)}$ is the state transition matrix that models the system dynamics, and $v_{(k)}$ is the process noise, which accounts for uncertainties in the system model. The process noise is typically assumed to follow a Gaussian distribution.

\paragraph{Update Phase:} In the update phase, the Kalman Filter refines its prediction by incorporating new measurement data. The update equation is expressed as:

\begin{equation}\label{3}
Z_{(k)} = H_{(k)}X_{(k)} + u_{(k)},
\end{equation}

\noindent where $Z_{(k)}$ is the measurement at time step $k$, $H_{(k)}$ is the measurement matrix that relates the true state to the observed measurement, and $u_{(k)}$ is the measurement noise, also assumed to be Gaussian. The filter updates its estimate of the object's state by minimizing the estimation error, taking into account both the predicted state and the actual measurement.

\paragraph{Importance and Applications:} The strength of the Kalman Filter lies in its ability to provide real-time, optimal estimates of the object's state while accounting for uncertainties in both the model and the measurements. This makes it particularly effective for tracking objects in environments where occlusions, rapid changes in motion, or noisy measurements are common. For instance, in video-based object tracking, the Kalman Filter can predict the position of an object even when it temporarily disappears from the frame due to occlusion, and then seamlessly update its estimate once the object reappears.

Moreover, the Kalman Filter is extendable to nonlinear systems through its variants, such as the Extended Kalman Filter (EKF) and the Unscented Kalman Filter (UKF). The EKF linearizes the system around the current estimate, making it suitable for tracking objects with nonlinear motion patterns. The UKF, on the other hand, provides a more accurate approximation by using a deterministic sampling technique to handle nonlinearity.

The Kalman Filter's robust performance in handling noise and uncertainty has made it a cornerstone technique in various tracking applications, ranging from tracking vehicles in traffic systems \cite{b124} to following the motion of cells in biological studies \cite{b125}. Its formulation and implementation are crucial for developing advanced tracking systems that require precision and adaptability in dynamic environments.

In conclusion, the Kalman Filter remains a fundamental tool in the arsenal of tracking algorithms, offering a balance of simplicity, efficiency, and robustness. Its application to object tracking, particularly in dealing with occlusions and measurement noise, underscores its versatility and enduring relevance in modern computational techniques.

\begin{figure}
\centerline{\includegraphics[width=8.5cm, height=9cm]{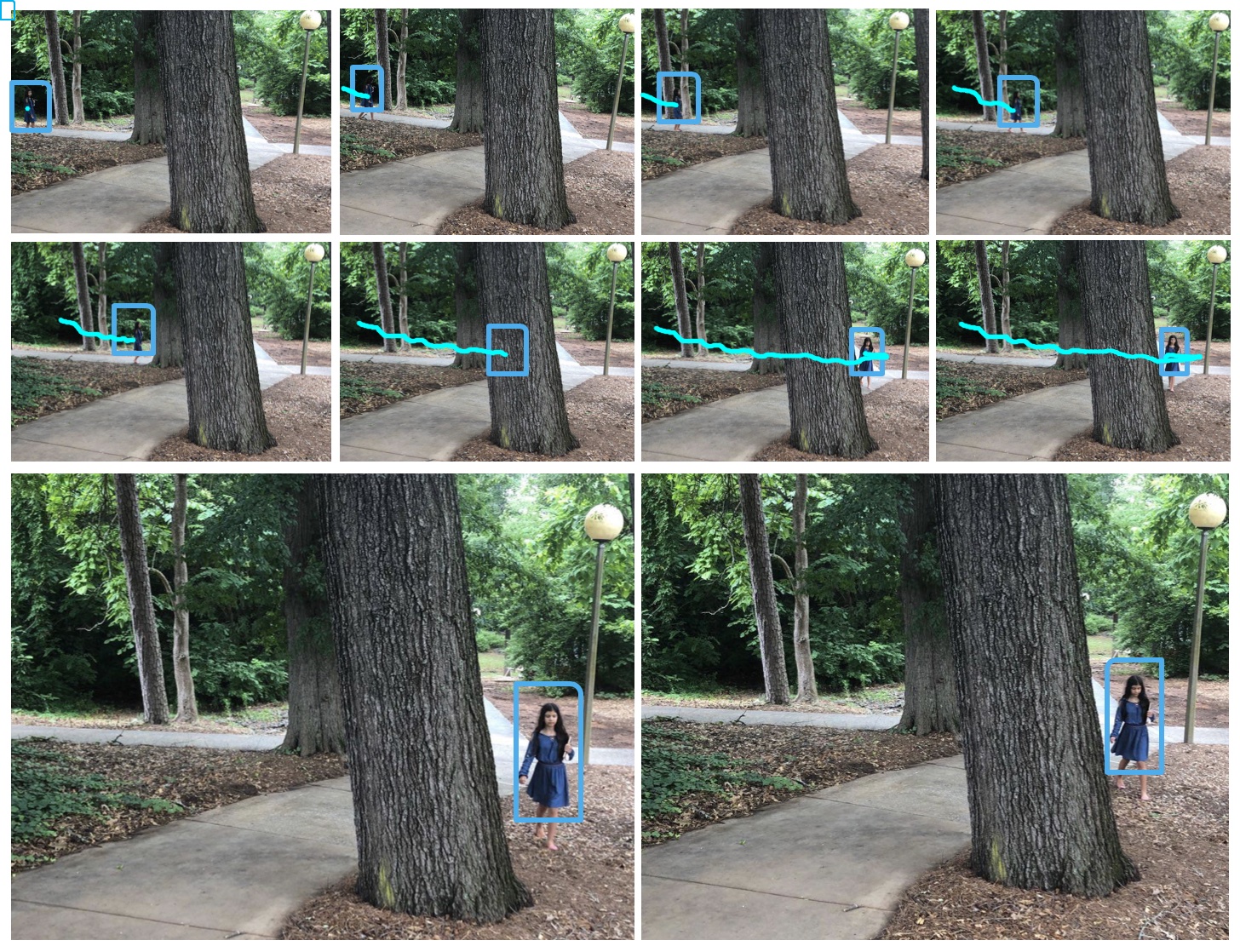}}
\caption{\textit{Tracking the person in occlusion using Kalman Filter method \cite{b57}}}
\label{fig:9-1}
\end{figure}

\begin{figure}
\centerline{\includegraphics[width=9cm, height=7.5cm]{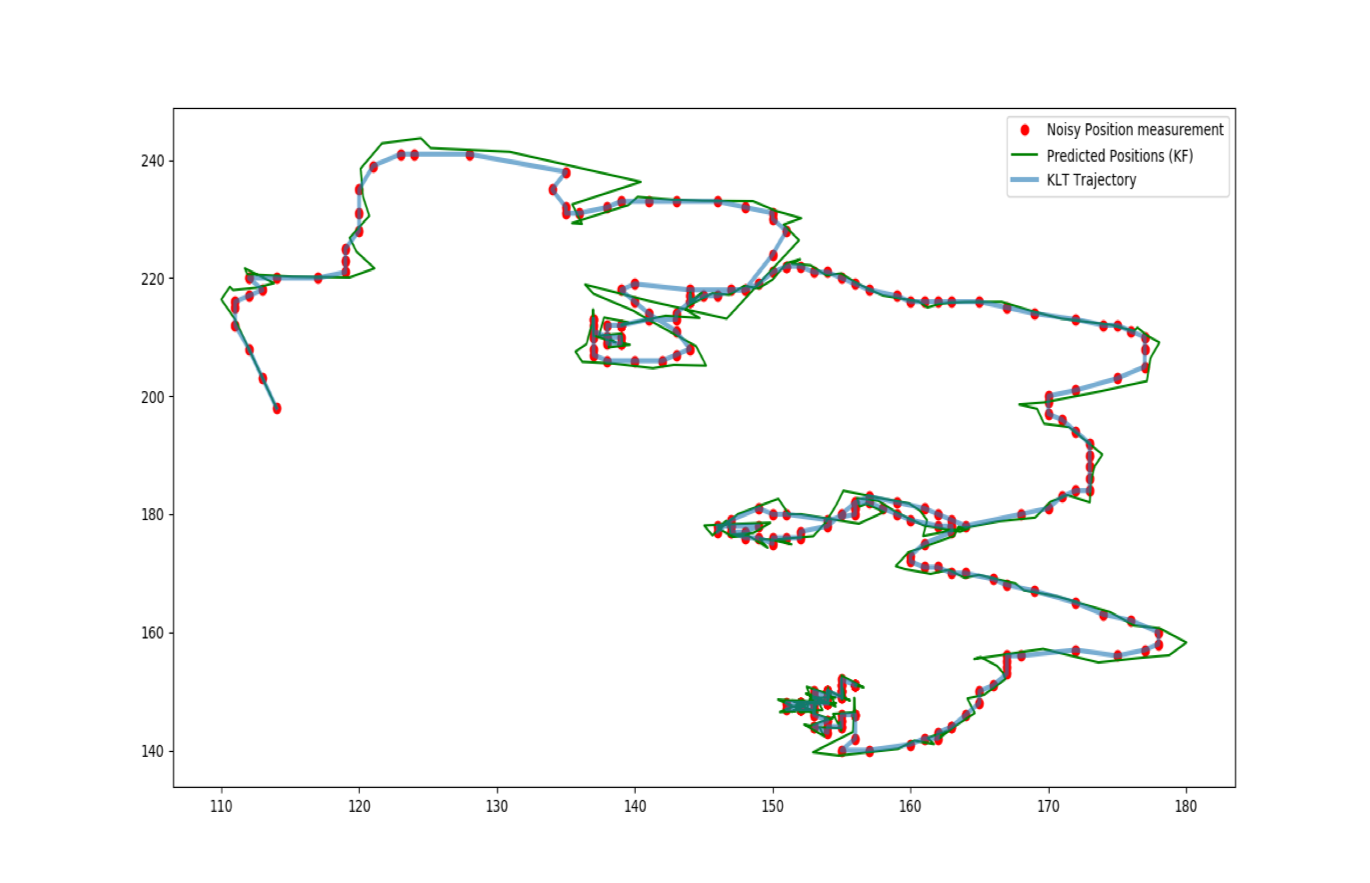}}
\caption{\textit{A single \textit{T. gondii} trajectory tracked by Kalman Filter, the blue line indicates the ground truth, the green line is the KF prediction of the trajectory and the red dots indicate noisy position measurement.}}
\label{fig:7}
\end{figure}

\begin{figure*}[ht]
\centerline{\includegraphics[width=0.8\textwidth, height=12cm]{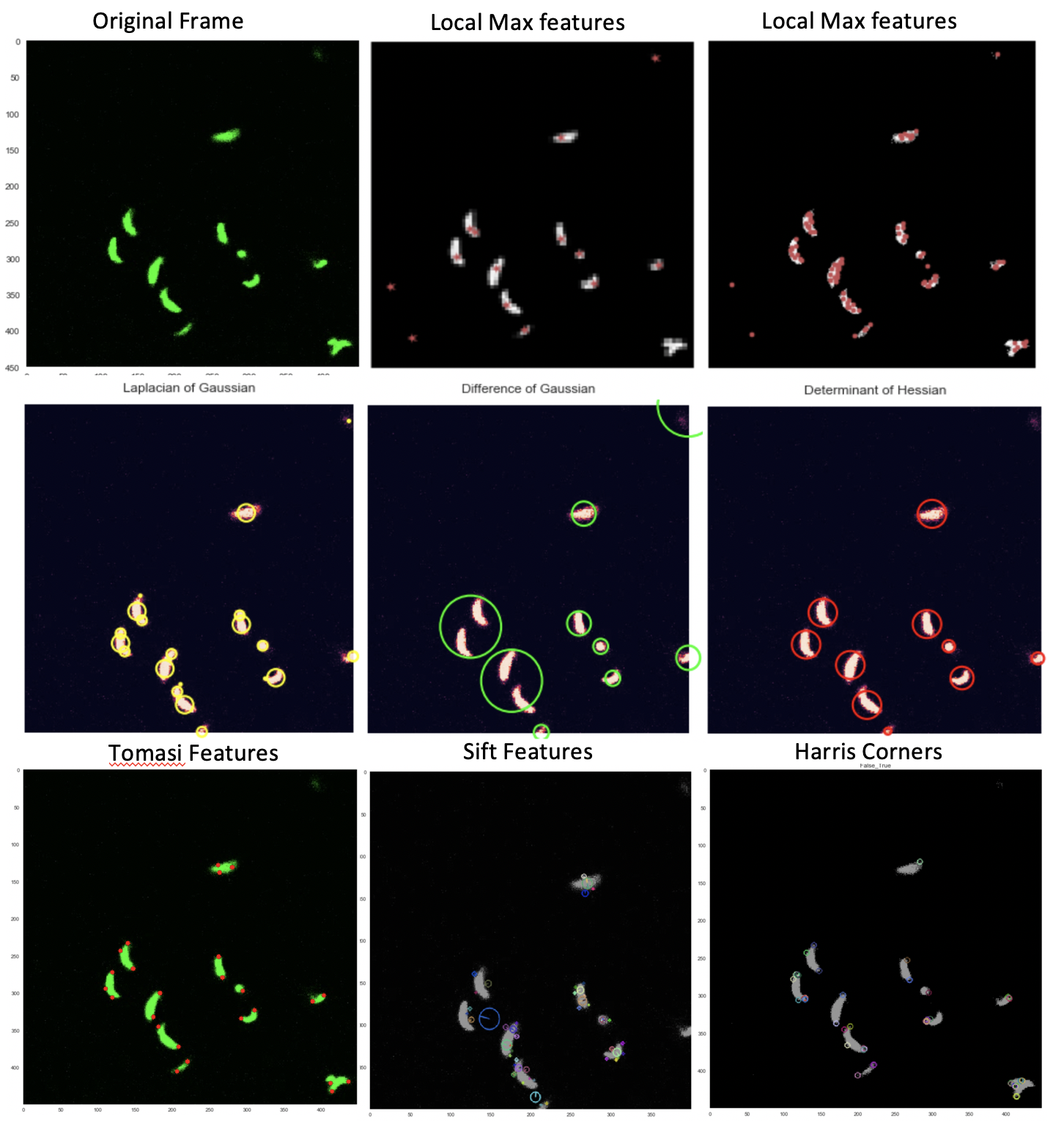}}
\caption{\textit{Different detection methods applied to a sample frame of 2D \textit{T. gondii} video microscopy. The first row shows the original image and the maxima features. In the second row different blob detection techniques are applied: Laplacian of Gaussian, Difference of Gaussians, and the Determinant of Hessian. In the last row the salient features and corners are detected using Tomasi, SIFT, and Harris corner detection respectively.}}
\label{fig:10}
\end{figure*}

\subsubsection{Kalman Filter for Multiple Object Tracking}

The Kalman Filter is a widely used algorithm for motion prediction and tracking in various applications, particularly effective for single-object tracking scenarios (Fig. \ref{fig:7}). The Kalman Filter operates by recursively estimating the state of a moving object, which includes its position, velocity, and acceleration, based on a series of measurements over time. The algorithm predicts the next state of the object using a motion model and corrects this prediction by incorporating new measurements, accounting for the inherent uncertainty in the measurements. This iterative process allows the Kalman Filter to provide a continuous and smooth estimate of the object's trajectory.

However, when dealing with multiple objects, the tracking problem becomes more complex due to the need to associate the correct measurements with the corresponding objects across successive frames. To address this challenge, the Kalman Filter is often combined with an additional algorithm, such as the \textit{Hungarian algorithm}, to ensure accurate identification and linking of objects between frames.

The \textit{Hungarian algorithm} is a combinatorial optimization method that complements the Kalman Filter by solving the assignment problem, which involves matching detected objects in frame $k$ to their corresponding counterparts in frame $k+1$. This is achieved by minimizing the cost, typically defined as the Euclidean distance between the predicted and detected positions of the objects, within an assignment matrix. The Hungarian algorithm finds the optimal assignment that minimizes the total cost, ensuring that each object is correctly linked across frames.

Originally proposed by \textit{Harold Kuhn} in 1955, the Hungarian algorithm, also known as the \textit{Munkres} or \textit{Kuhn-Munkres} algorithm, was formulated to solve the minimum cost matching problem. Consider a cost matrix \textbf{COST}, where \textbf{COST ${[a, b]}$} represents the cost of assigning worker $a$ to job $b$. The objective of the algorithm is to find an optimal assignment that minimizes the total cost. Mathematically, this is expressed as finding a boolean matrix $X_{[a, b]}$, where the value is $1$ if row $a$ is assigned to column $b$, and $0$ otherwise. The optimal assignment cost is given by the following equation:

\begin{equation}\label{4}
\min \sum_{a}\sum_{b}{COST_{a,b} \cdot X_{a,b}}
\end{equation}

In the context of object tracking, the Kalman Filter predicts the future state of each object, including its position and velocity, while accounting for uncertainties in the motion model and measurement errors. The Hungarian algorithm is then applied to associate the predicted states of objects in the current frame with the detected objects in the subsequent frame, based on the minimized cost. This process ensures that each object is accurately tracked over time, even in complex scenarios involving multiple objects.

One of the most significant advantages of the Kalman Filter, particularly in conjunction with the Hungarian algorithm, is its ability to handle occlusion. During occlusion, when an object is temporarily obscured, the Kalman Filter can continue to predict the object's state, including its speed and position, based on its previous trajectory. Once the object reappears, the Hungarian algorithm helps to re-establish the correct association, thereby maintaining the continuity of the object's tracking.

This combined approach of the Kalman Filter and the Hungarian algorithm has proven to be highly effective in multiple object tracking tasks, making it a cornerstone technique in applications ranging from video surveillance to autonomous driving and beyond.

We implemented the Kalman Filter (KF) approach for tracking \textit{T. gondii} in a post-processing step, which significantly enhances the accuracy of trajectory estimation, particularly in cases of occlusion. Initially, we utilized the Kanade-Lucas-Tomasi (KLT) tracker to track the cells. However, to address potential occlusions and ensure continuous and reliable tracking, we developed a post-processing method that integrates the Kalman Filter with the Hungarian algorithm for trajectory correction. This hybrid approach enables the system to dynamically adjust and correct the predicted paths of cells when occlusions occur, thereby maintaining tracking integrity. The efficacy of this method is illustrated in Fig. \ref{fig:7}, where the tracking results are presented.

The approach we adopted has parallels in other studies within the domain. Shiuh-Ku Weng et al. \cite{b57} implemented a similar strategy to address occlusion challenges in object tracking. Their results, depicted in Fig. \ref{fig:9-1}, further validate the effectiveness of using the Kalman Filter for tracking in complex scenarios.

Moreover, the Kalman Filter and its variants, such as the Extended Kalman Filter (EKF), have been extensively employed in numerous tracking applications, ranging from tracking multiple objects to more specialized scenarios like space object tracking. For instance, several research papers have explored the use of KF and EKF in various object tracking problems \cite{b126}--\cite{b130}. Among these, a notable contribution is the Universal Kalman Consensus Filters (UKCF) proposed by aerospace researchers for tracking space objects \cite{b130}. The UKCF method was developed to address significant challenges in the field, such as time-fusion problems associated with asynchronous measurements in cooperative space object tracking. The detailed methodology and results are thoroughly discussed in their study, highlighting the robustness and adaptability of the Kalman Filter in diverse environments.

In another relevant study, a rapid approach for tracking multiple individuals in occluded scenes using the Kalman Filter was proposed by Nguyen and Smeulders \cite{b58}. Their method demonstrated the KF's capability to efficiently handle occlusions and maintain accurate tracking in densely populated or visually cluttered environments.

\begin{figure*}[h!]
\centering
\includegraphics[height=6cm, width=.8\textwidth]{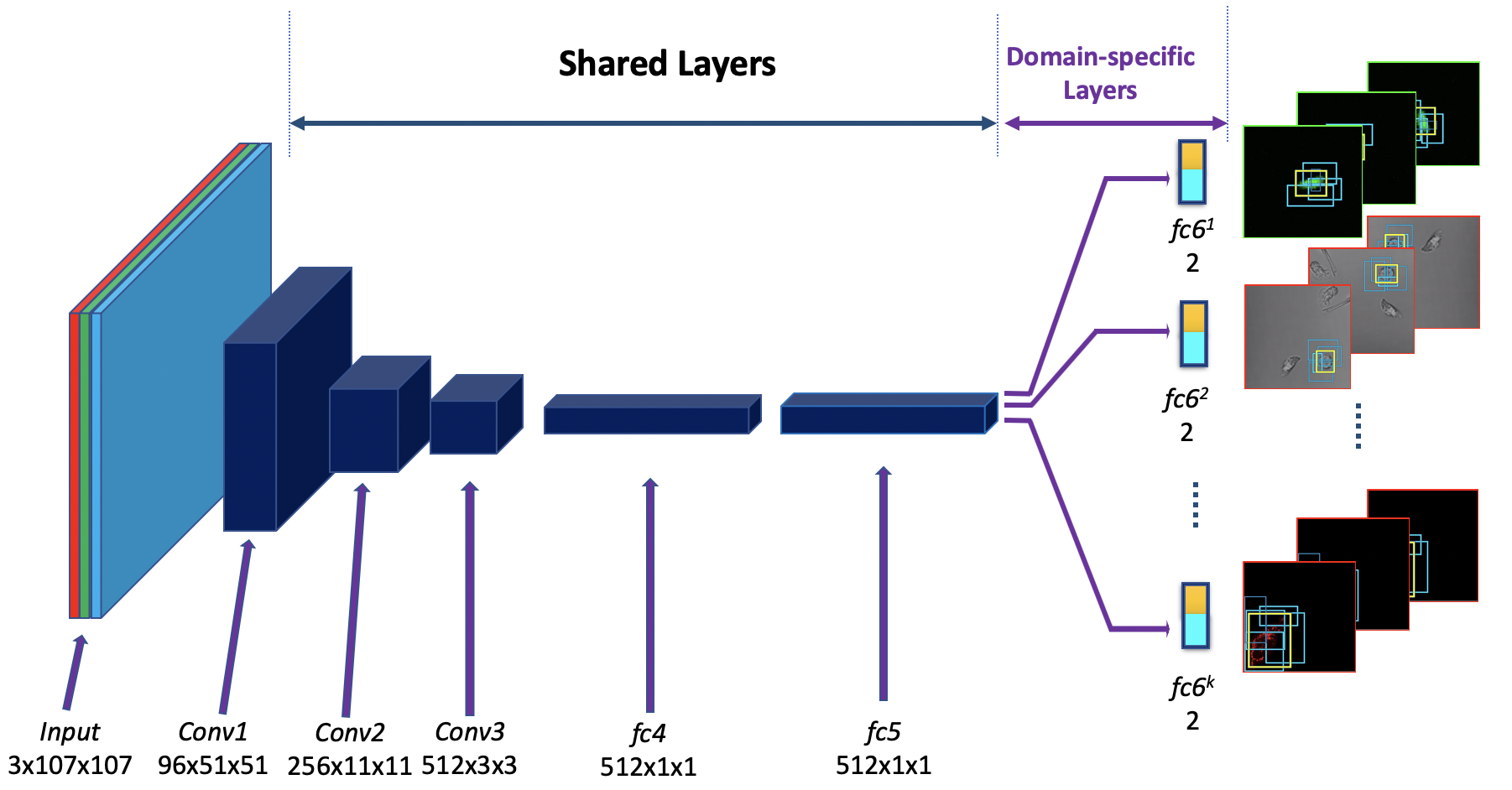}
\caption{\textit{Overview of the \textit{MDNet} Model: This model learns domain-independent representations through pre-training and captures domain-specific information via online learning during tracking \cite{b60}}.}
\label{fig:11-2}
\end{figure*}

In addition to the Kalman Filter, other researchers have proposed enhancements and alternatives to improve tracking performance under occlusion. For example, Rosales et al. \cite{b53} developed a system for tracking multiple humans with trajectory prediction, as shown in Fig. \ref{fig:KLT-ext}. Their approach utilized an Extended Kalman Filter (EKF) to estimate the position and velocity of objects, which significantly improved the prediction and tracking of human trajectories, even in challenging conditions where occlusions were frequent.

These advancements underscore the versatility and reliability of Kalman-based filters in handling the complexities of object tracking, particularly in scenarios involving occlusions or other challenging conditions. By refining these models and integrating them with complementary algorithms like the Hungarian method, researchers continue to push the boundaries of what can be achieved in real-time tracking systems.

\subsubsection{Gaussian Mixture Model (GMM)}

In the realm of object tracking, certain types of target shapes and motions necessitate region-based handling rather than a particle-centric view. One prominent approach that addresses this need is \textit{"Density-based Tracking."} Unlike deterministic tracking methods that pinpoint the exact location of an object, density-based tracking provides a probabilistic distribution of potential object locations or fragments. This probabilistic approach, which leverages statistical models, has proven particularly effective in complex tracking scenarios where uncertainty and variability are significant. Among the various models used in density-based tracking, the Gaussian Mixture Model (GMM) stands out for its robust performance \cite{b217}--\cite{b220}, \cite{b85}.

A Gaussian Mixture Model (GMM) is a probabilistic model that represents a distribution as a combination of multiple weighted Gaussian distributions, each contributing to the overall shape of the distribution. GMMs are widely utilized in conventional tracking methods due to their flexibility and effectiveness in modeling complex data distributions. The core idea behind a GMM is that data points are assumed to be generated from a mixture of a finite number of Gaussian distributions, each with its own mean and variance \cite{b211, b212}, \cite{b85}, \cite{b205, b206}.

The GMM process can be seen as a generalization of the K-Means clustering algorithm, but with a probabilistic foundation. The process begins by specifying $k$ multivariate Gaussian distributions, known as \textit{components}. Initially, the means and variances of these components are randomly initialized. The next step involves calculating the probability that each data point belongs to each component. This probabilistic assignment allows for a soft clustering, where data points are not strictly assigned to a single cluster but rather have a probability of belonging to multiple clusters.

Once the probabilities are calculated, each data point is associated with the component that has the highest probability. Subsequently, the means and variances of the components are updated to reflect the actual data distribution. This iterative process continues until the model converges, meaning that the changes in the means and variances become negligible.

GMMs are often described as semi-parametric models. This is because, while they rely on Gaussian components, these components may not adhere strictly to a known distribution like the normal distribution. Instead, GMMs allow for a flexible representation where additional components can model more complex or unknown distributions. This flexibility makes GMMs particularly useful in tracking applications where the data may exhibit varying patterns and distributions.

Numerous studies have explored the application of GMMs in object tracking, particularly for tasks such as distinguishing background from foreground, segmenting foreground objects, and color-based object tracking. The ability of GMMs to model complex data distributions makes them well-suited for these tasks, where accuracy and adaptability are critical \cite{b212}--\cite{b217}.

In summary, Gaussian Mixture Models offer a powerful framework for density-based tracking by providing a probabilistic approach to object localization and segmentation. Their ability to model complex and variable data distributions ensures their continued relevance and effectiveness in the evolving field of object tracking.

\subsubsection{Bayesian and Markov-based Models}
In the realm of statistical methods for object tracking, Bayesian and Markov-based models have proven to be powerful tools for addressing the inherent uncertainties and dynamic variations in tracking scenarios. These methods, rooted in Markov theory, include approaches such as Auto-Regressive (AR) models and Hidden Markov Models (HMMs), both of which have been successfully applied to various tracking challenges.

For instance, Park et al. \cite{b13} introduced a Bayesian tracking framework that leverages the AR-HMM to manage the dependencies between consecutive target appearances. This approach is particularly effective in scenarios where the appearance of the target object changes over time, as the AR-HMM can model these variations while maintaining a probabilistic framework that accounts for uncertainty.

Another notable contribution comes from Tao et al. \cite{b216}, who developed a tracking method based on dynamic motion layer representation. Their approach models the spatiotemporal constraints on shape, motion, and layer representation, estimating these within a \textit{Maximum A Posteriori} (MAP) framework. To improve the computational efficiency of tracking objects with complex shapes, they introduced a shape prior that prevents motion layers from evolving into arbitrary and unrealistic forms. The use of the Expectation-Maximization (EM) algorithm in their estimation process further enhances the robustness of the tracker, particularly in cluttered backgrounds, by allowing the object and background layers to compete during the layer estimation process.

AR models, which are a type of stochastic process, are widely used in modeling linear systems and predicting future states based on previous observations. These models operate under the Markov assumption, wherein future states are dependent only on a certain number of preceding states. The number of previous states used to predict future observations defines the order of the system, as expressed by the following equations:

\begin{equation}\label{5}
\vec{y}_t = C\vec{x}_t + \vec{u}_t
\end{equation}

\begin{equation}\label{6}
\vec{x}_t = B_1\vec{x}_{t-1} + B_2\vec{x}_{t-2} + \dots + B_d\vec{x}_{t-d} + \vec{v}_t
\end{equation}

\begin{equation}\label{7}
\vec{x}_t = \sum_{j=1}^d{B_j\vec{x}_{t-j}} + \vec{v}_t
\end{equation}

Equation \ref{7} describes a system of order \(d\) \cite{b14}. In AR models, the transition matrices \(\{B_1, \dots, B_d\}\) play a crucial role as they encode the system's dynamics, providing a parameterization that is invariant to absolute spatial positions. These transition matrices allow for the prediction of future states based on past observations, making them essential in tracking scenarios where the object's position evolves over time.

In addition to AR models, other probabilistic approaches have been explored. For example, Rittscher et al. \cite{b15} proposed a Probabilistic Background Model for tracking, employing a particle filter as a stochastic filter to enhance the robustness of a car tracking system. Similarly, Yuan et al. \cite{b16} developed a visual object tracking algorithm based on the Observation-Dependent Hidden Markov Model (OD-HMM) framework. Their method incorporates structure complexity coefficients (SCCs) to predict variations in target appearance, further improving the accuracy and reliability of the tracking process.

Overall, Bayesian and Markov-based models provide a comprehensive framework for addressing the challenges of object tracking, particularly in dynamic and uncertain environments. By leveraging probabilistic reasoning and incorporating prior knowledge into the tracking process, these models offer a robust and flexible solution for a wide range of applications.

\begin{figure*}[ht]
\centerline{\includegraphics[width=\textwidth, height=8cm]{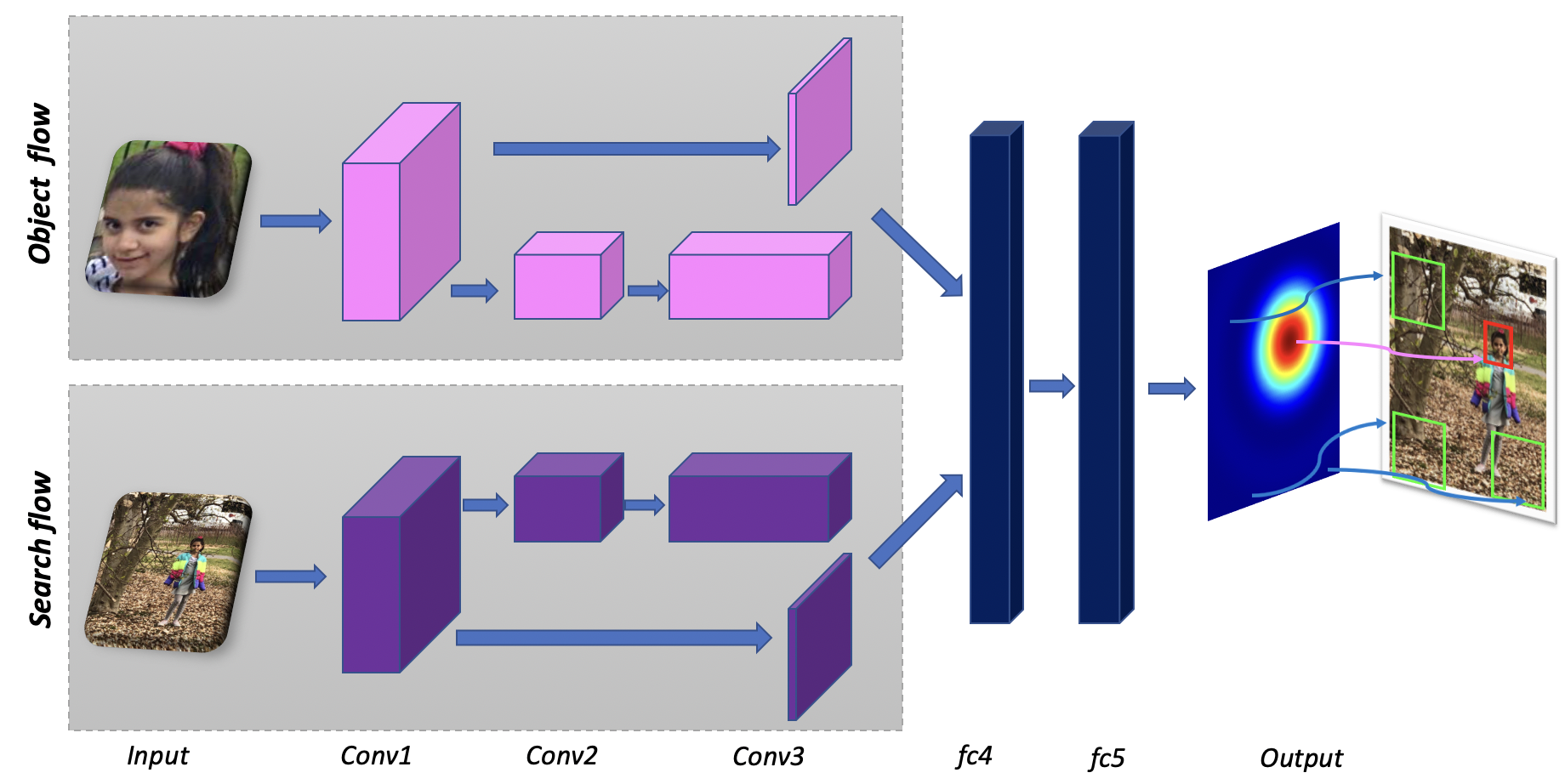}}
\caption{\textit{Schema of YCNN model\cite{b72} for object tracking. It divides the tracking into two different spaces in convolutional layers and outputs a heatmap in which the potential location of the object is indicated.}}
\label{fig:13}
\end{figure*}

\subsection{Machine Learning and Deep Learning-Based Methods}

Traditional tracking methods, as discussed earlier, often rely on simple models of the target that are adapted over time. While these methods have been effective in some scenarios, they face significant challenges when applied to real-world situations. One of the primary issues is that the object model may deviate from its original form over time, especially when dealing with non-rigid objects. This deviation can lead to a loss of accuracy and reliability in tracking. Additionally, conventional tracking methods are often computationally expensive, making them unsuitable for real-time applications. These methods also struggle with stability, particularly when tracking multiple objects of the same category simultaneously. Moreover, traditional methods typically lack a discriminative model to distinguish the object category of interest from others in the scene \cite{b8,b59}.

Another limitation of conventional methods is their reliance on model-based approaches, which means that the success of these models is highly dependent on the specific type of data they are applied to. For instance, a model that works well for one type of data may perform poorly on another. As shown in Table 1, one of the most significant drawbacks of conventional tracking methods is their inability to support end-to-end models within a unified tracking system. This limitation often results in fragmented and less efficient tracking processes.

In contrast, deep learning-based models offer a data-driven approach that addresses many of the inconsistencies and limitations of traditional methods. Deep learning allows for the creation of end-to-end tracking frameworks by learning and optimizing the parameters that were manually tuned in conventional methods. This approach significantly enhances tracking performance and offers greater flexibility in adapting to various scenarios. Moreover, deep learning models can be trained in both supervised and unsupervised ways, allowing them to learn from vast amounts of data and improve their accuracy over time.

The recent advancements in computational tools and resources have played a crucial role in the resurgence of neural network methods, particularly deep learning. The development of powerful new generation CPUs, the extraordinary computational capabilities of GPUs, and the availability of cloud-based resources have all contributed to the rapid progress in deep learning research and its applications. These advancements have enabled the deployment of deep learning models in real-time tracking systems, where speed and accuracy are paramount.

As a result of these developments, modern tracking systems are increasingly based on deep learning, including those built on the \textit{tracking by detection} paradigm. This paradigm leverages the power of deep learning to improve detection accuracy and efficiency. The emergence of advanced detection algorithms, such as YOLO, Faster R-CNN, Mask R-CNN, Cascade R-CNN, and SSD, has further enhanced the capabilities of tracking models. These algorithms, when integrated with deep learning-based tracking systems, significantly reduce computational time and make online tracking systems more practical and effective.

In summary, while traditional tracking methods have laid the foundation for object tracking, the advent of deep learning has revolutionized the field, providing more robust, flexible, and efficient solutions that are better suited to the complexities of real-world applications.

\begin{figure*}[h!]
\centering
\includegraphics[width=1.0\textwidth]{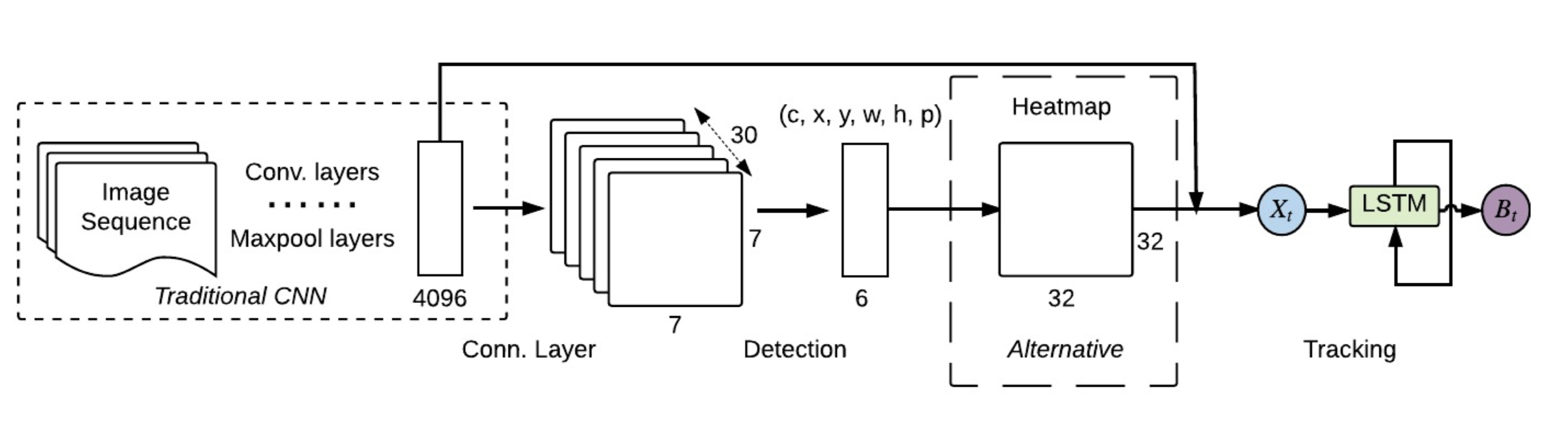}
\caption{\label{fig:12} The ROLO architecture (\textcopyright IEEE2017)for object tracking. the alternative heatmap module is used to check the resistance of the model against occlusion\cite{b70}.}
\end{figure*}

\subsubsection{The Emergence of Deep Tracking Models}
Convolutional Neural Networks (CNNs) have revolutionized the field of computer vision, demonstrating exceptional performance across a wide range of applications, including image classification, semantic segmentation, and object detection \cite{b61}--\cite{b67}. Despite these successes, the application of CNNs to visual tracking has been somewhat limited. This limitation arises primarily due to the practical challenges associated with online visual tracking, where the need for extensive training time and large volumes of labeled data make CNNs less viable. Moreover, the lack of specialized training algorithms tailored for visual tracking further exacerbates these challenges \cite{b60}. As a result, traditional approaches based on low-level handcrafted features continue to dominate in practical tracking systems.

For example, even with the availability of large-scale datasets like ImageNet, CNN-based tracking methods often fall short. The work by Hong et al. \cite{b68} and Wang et al. \cite{b69} highlights this limitation, where despite leveraging deep learning techniques, the tracking performance did not meet expectations. The fundamental disconnect between classification tasks, where the goal is to predict class labels, and tracking tasks, which involve the continuous localization of targets across frames, underpins this challenge \cite{b60}. Nevertheless, deep learning models offer distinct advantages, particularly their robustness and powerful learning capabilities, which facilitate the development of seamless end-to-end tracking systems.

One of the pioneering attempts to incorporate CNNs into tracking was presented by Fan et al. in 2010 \cite{b59}. They introduced a CNN-based approach that treated human tracking as a specific instance of object class tracking. In their method, both spatial and temporal structures were learned during offline training, allowing the system to track only predefined object classes, such as humans. However, this method required the CNN to be trained offline before tracking, limiting its applicability to new or unseen object classes during runtime.

\begin{figure}[h!]
\centering
\includegraphics[width=9cm, height=5cm]{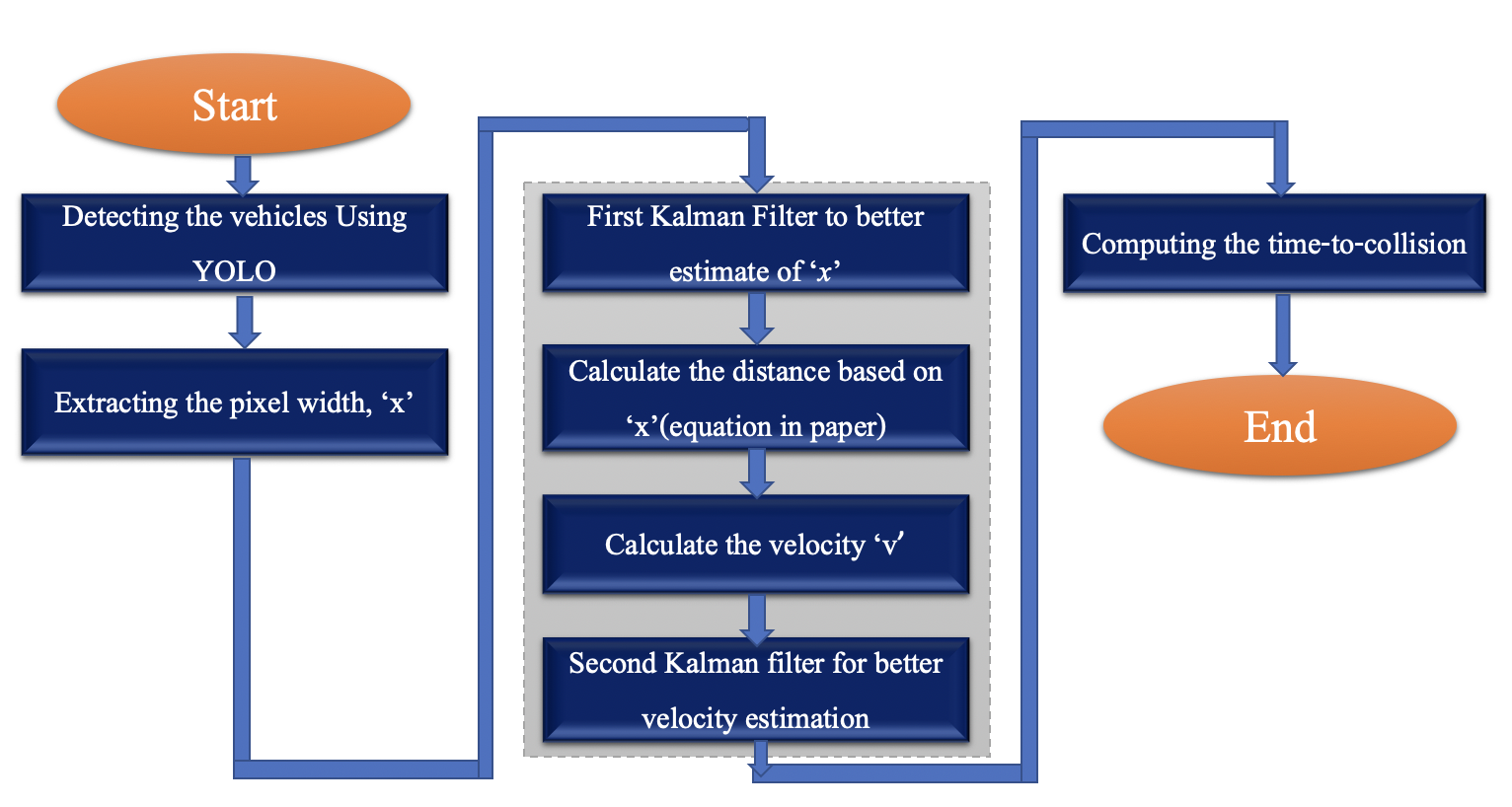}
\caption{\textit{YOLO + The nested KF chart used in \cite{b76}}}
\label{fig:14}
\end{figure} 

In a more recent development, Hyeonseob et al. \cite{b60} introduced the Multi-Domain Network (MDNet), addressing some of the inherent challenges in object tracking. They argued that the variations and inconsistencies across video sequences create significant hurdles for effective tracking. Traditional learning methods, particularly those based on standard classification tasks, are inadequate for this purpose. To overcome these limitations, MDNet was designed to learn domain-independent representations during the pre-training phase while simultaneously capturing domain-specific information through online learning during tracking.

The MDNet architecture is relatively simple compared to conventional image classification models. The entire network is pre-trained offline, focusing on generalizable features across multiple domains. During tracking, the fully connected layers, including a single domain-specific layer, are fine-tuned online to adapt to the specific characteristics of the current sequence. This dual approach allows MDNet to balance the generalization required for diverse tracking scenarios with the specialization needed for precise object tracking. The architecture of MDNet is depicted in Fig. \ref{fig:11-2}.

\begin{figure*}[h!]
\centering
\includegraphics[width=0.8\textwidth]{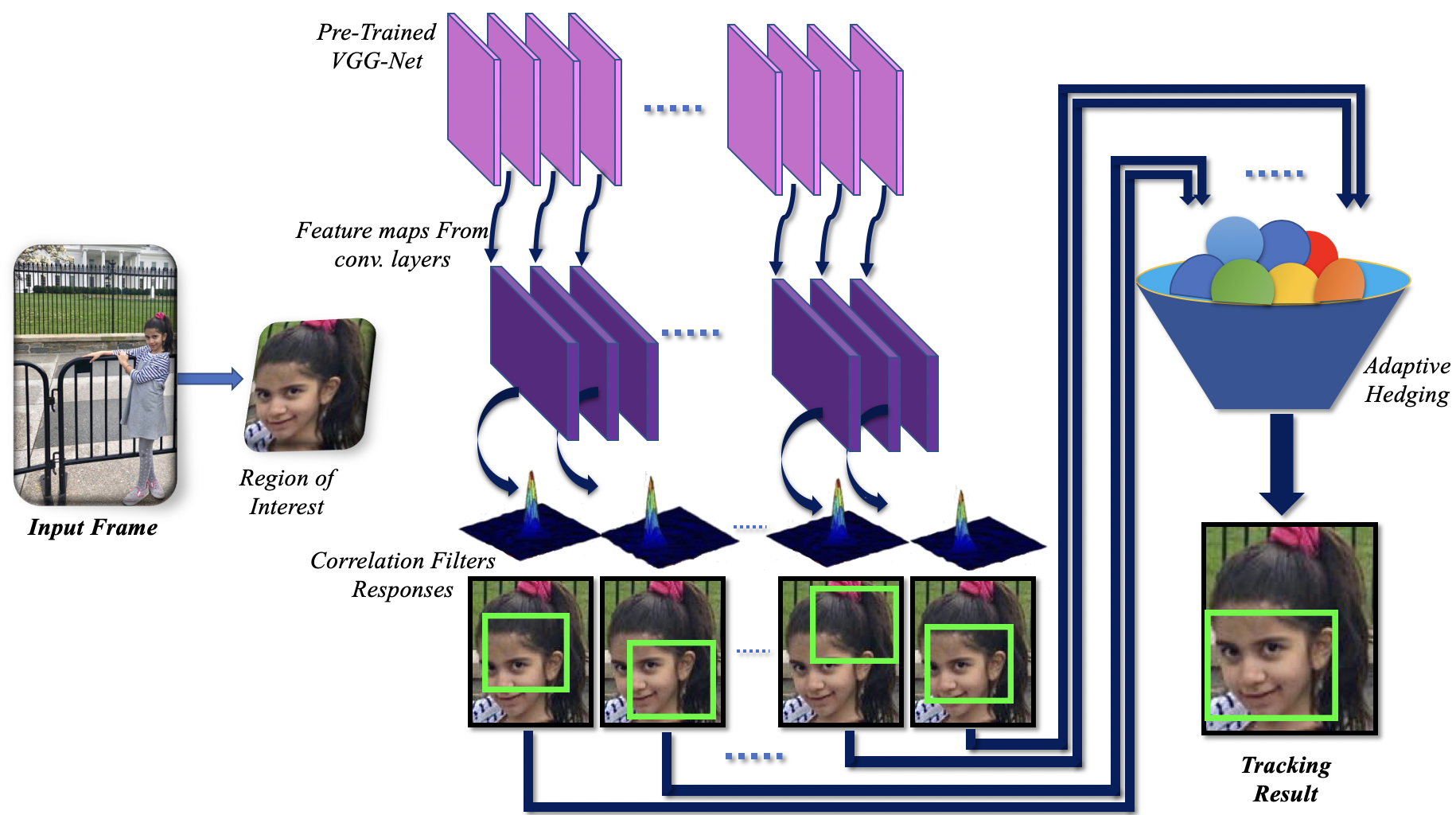}
\caption{\textit{The steps of the proposed algorithm in \cite{b77}. First, it uses the pre-trained VGG-Net to extract the CNN features from different convolutional layers. Then, it constructs weak trackers using correlation filters that are trained with CNN features from one layer, and finally, it hedges those weak trackers into a stronger one.}}
\label{fig:15-3}
\end{figure*}

\subsubsection{Supervised, Semi-Supervised, and Unsupervised Methods}

Object detection and tracking, fundamental tasks in computer vision, can be broadly categorized into three distinct methodological approaches: \textit{Supervised}, \textit{Semi-Supervised} (also referred to as Weakly-Supervised), and \textit{Unsupervised} methods. Each of these approaches comes with its own set of advantages and challenges, primarily distinguished by the level of annotated data required for training and the complexity of the learning process involved.

\paragraph{Supervised Methods}

Supervised methods are the cornerstone of most state-of-the-art object detection systems. These methods rely heavily on large datasets that are meticulously labeled and annotated. The annotations serve as a form of mentorship, guiding the model to learn intricate patterns and associations within the data. The supervised learning paradigm has given rise to several highly successful deep learning models. Notable examples include YOLO (You Only Look Once) \cite{b71, b183}, SSD (Single Shot Multibox Detector)\cite{b159}, Mask R-CNN \cite{b172}, Cascade R-CNN \cite{b173}, Fast R-CNN \cite{b279}, and Faster R-CNN\cite{b160}. These models excel in accuracy and performance, often being the first choice for tasks requiring high precision.

For instance, YOLO is renowned for its real-time detection capabilities, while Faster R-CNN is praised for its accuracy in detecting objects at various scales. These models typically involve complex architectures, such as convolutional neural networks (CNNs), which are trained on vast amounts of labeled data to minimize error and enhance generalization across unseen data. However, the major drawback of supervised methods is their dependency on large-scale annotated datasets, which are time-consuming and expensive to create.

\paragraph{Semi-Supervised Methods}

Semi-supervised methods, on the other hand, offer a compromise by reducing the dependency on labeled data. These methods require a smaller amount of labeled data during the training phase, which is supplemented by a larger corpus of unlabeled data. The labeled data helps guide the learning process, while the model simultaneously learns to make predictions on the unlabeled data, effectively utilizing the abundant unannotated data that is often available.

A prime example of a semi-supervised approach is Class Activation Mapping (CAM) \cite{b161}, which allows for the identification of object locations within an image by highlighting the regions that contribute most to the classification decision. Another significant contribution in this category is the method proposed by Wang et al. in 2019 \cite{b222}, which innovatively combines object tracking and segmentation into a unified framework. Their approach involves creating a binary mask to separate the object from the background, followed by the application of a Siamese convolutional network to track the object over time. This method demonstrates how semi-supervised learning can effectively balance the need for labeled data with the practical limitations of data annotation.

\paragraph{Unsupervised Methods}

Unsupervised methods represent the most autonomous approach, requiring no labeled data for training. These methods are particularly valuable in scenarios where labeled data is scarce or unavailable. Instead of relying on annotations, unsupervised methods focus on learning the underlying structure and patterns in the data through techniques like clustering, dimensionality reduction, and generative modeling.

A notable advancement in unsupervised learning is the development of contrastive learning frameworks, such as Contrastive Predictive Coding (CPC) \cite{b224} and Momentum Contrast (MoCo) \cite{b226}. These methods leverage the idea of contrasting positive and negative pairs to learn useful representations of the data. By predicting certain aspects of the data, such as future frames in a video or the context of an image, these models can learn to track and detect objects without any prior labeling. The flexibility and efficiency of unsupervised methods make them an attractive choice for real-world applications where annotated datasets are limited or non-existent.

In summary, while supervised methods dominate the current landscape of object detection and tracking due to their accuracy and reliability, semi-supervised and unsupervised methods are rapidly gaining traction. The latter approaches offer more scalable and cost-effective alternatives, particularly in environments where the acquisition of labeled data is challenging. As research in these areas continues to evolve, we can expect to see increasingly sophisticated models that combine the strengths of all three approaches to deliver robust and versatile object detection and tracking solutions.

\begin{table*}[htbp]
\centering
\caption{Strengths and Shortcomings of Reviewed Detection-Based Tracking Methods}
\label{tab:methods_comparison}
\begin{tabularx}{\textwidth}{@{}l*{10}{C}c@{}}
 \toprule
\hline
\textbf{Method}           & \textbf{Strengths}                                                                                  & \textbf{Shortcomings}                                                                                       \\ \hline
\textbf{ROLO}             & Combines YOLO with LSTM, effectively handling occlusion and motion blur. Predicts object trajectories.   & Slower processing speed due to the recurrent layers in the architecture.                                      \\ \hline
\textbf{DeepTrack}        & Adapts online with CNN, robust to label noise, and effectively handles appearance changes.               & Computationally intensive, not suitable for real-time tracking due to the iterative learning approach.         \\ \hline
\textbf{Rank-DETR}        & Provides high-precision object localization by aligning confidence and localization accuracy.            & Requires large memory resources due to the transformer architecture, which may limit scalability.              \\ \hline
\textbf{Type-to-Track}    & Allows flexible, user-driven tracking through natural language descriptions, increasing adaptability.    & Dependent on language model quality; struggles with ambiguous or unclear object descriptions.                  \\ \hline
\textbf{DeepSORT++}       & Enhances object matching in crowded environments with better embedding features, robust in MOT.          & Performance decreases when tracking very fast-moving objects or under extreme occlusion.                       \\ \hline
\textbf{MotionTrack}      & Integrates GNNs for motion prediction, excelling in real-time applications in complex environments.      & Demands significant computational resources, making real-time implementation more challenging.                 \\ \hline
\textbf{Bi-Directional Tracking} & Enables unsupervised training by tracking forward and backward in video, reducing labeled data dependency.  & Currently focused on single-object tracking, limiting its usefulness in scenarios requiring multi-object tracking. \\ \hline
 \bottomrule
 \end{tabularx}
\end{table*}

\begin{table*}[htbp]
\centering
\caption{\textbf{General Comparison of Deep Learning based Vision Models}}
\label{tab:methods_comparison}
\renewcommand{\arraystretch}{1.3} 
\setlength{\tabcolsep}{3pt} 
\fontsize{11.5}{12.5}\selectfont 
\resizebox{\linewidth}{!}{ 
\begin{tabular}{|p{3cm}|p{6cm}|p{4cm}|p{6cm}|p{6cm}|p{3cm}|p{2cm}|}
\hline
\textbf{Model} & \textbf{Key Features} & \textbf{Main Application} & \textbf{Advantages} & \textbf{Challenges} & \textbf{Original Paper} & \textbf{Year} \\ \hline
CNN & 
1. Feature extraction using convolution. \newline 
2. Spatial hierarchies. \newline 
3. Effective with large datasets. & 
Image recognition & 
1. Efficient feature extraction. \newline 
2. Handles large images. & 
1. Requires large datasets. \newline 
2. Overfitting with small data. \newline 
3. Computationally intensive. & 
LeCun et al. \cite{b276} & 
1989 \\ \hline

RNN & 
1. Sequential data processing. \newline 
2. Handles variable-length inputs. \newline 
3. Captures temporal dependencies. & 
Time-series data & 
1. Captures temporal relationships. \newline 
2. Adapts to varying input sizes. & 
1. Vanishing gradients. \newline 
2. Long training times. \newline 
3. Sensitive to sequence length. & 
Rumelhart et al. \cite{b277} & 
1986 \\ \hline

LSTM & 
1. Memory cell structure. \newline 
2. Avoids vanishing gradients. \newline 
3. Long-term dependency modeling. & 
Time-series prediction & 
1. Models long-term dependencies. \newline 
2. Prevents vanishing gradients. & 
1. Computationally demanding. \newline 
2. Complex implementation. \newline 
3. Requires careful tuning. & 
Hochreiter and Schmidhuber \cite{b247} & 
1997 \\ \hline

RESNet & 
1. Residual connections. \newline 
2. Deeper architectures possible. \newline 
3. Enhanced training stability. & 
Image recognition & 
1. Improved training stability. \newline 
2. Enhanced gradient flow. & 
1. Complex training requirements. \newline 
2. High resource needs. \newline 
3. Overfitting risks. & 
He et al. \cite{b190} & 
2015 \\ \hline

3D ResNet & 
1. 3D convolutions for spatial-temporal data. \newline 
2. Action recognition capabilities. \newline 
3. Extended ResNet framework. & 
Action recognition & 
1. Accurate action recognition. \newline 
2. Handles spatiotemporal data. & 
1. High computational cost. \newline 
2. Requires spatiotemporal data. \newline 
3. Challenging to train. & 
Tran et al. \cite{b278} & 
2015 \\ \hline

FastRCNN & 
1. Region-based object proposals. \newline 
2. End-to-end training. \newline 
3. High precision for detection tasks. & 
Object detection & 
1. High accuracy. \newline 
2. Extensible for object localization. & 
1. Slow compared to newer models. \newline 
2. High resource needs. \newline 
3. Challenging to optimize. & 
Girshick et al.\cite{b279} & 
2015 \\ \hline

Faster RCNN & 
1. Region proposal networks. \newline 
2. Faster processing than FastRCNN. \newline 
3. Multi-task learning for detection. & 
Object detection & 
1. Faster than predecessors. \newline 
2. High precision in detection. & 
1. Resource-intensive. \newline 
2. High implementation complexity. \newline 
3. Requires large datasets. & 
Ren et al. \cite{b160} & 
2015 \\ \hline

YOLO & 
1. Real-time object detection. \newline 
2. Single-stage architecture. \newline 
3. High-speed inference. & 
Object detection & 
1. Real-time processing. \newline 
2. Lightweight model architecture. & 
1. Trade-off between speed and accuracy. \newline 
2. Not as precise. \newline 
3. Limited in some edge cases. & 
Redmon et al. \cite{b71} & 
2016 \\ \hline

YOLOv7 & 
1. Advanced real-time object detection. \newline 
2. Efficient network architecture. \newline 
3. Handles high-speed inference. & 
Object detection & 
1. Improved accuracy compared to YOLOv5/6. \newline 
2. Better trade-off between speed and accuracy. & 
1. Complex model tuning. \newline 
2. Sensitive to dataset quality. \newline 
3. High computational cost. & 
Wang et al. \cite{b280} & 
2022 \\ \hline

ROLO & 
1. Combines YOLO and RNN. \newline 
2. Real-time object tracking. \newline 
3. Predicts temporal object motion. & 
Object tracking & 
1. Integrated detection and tracking. \newline 
2. Suitable for real-time use. & 
1. Complex to implement. \newline 
2. Requires real-time systems. \newline 
3. High resource consumption. & 
Ning et al. \cite{b70} & 
2017 \\ \hline

DeepSORT++ & 
1. Enhanced object matching capabilities. \newline 
2. Robust in crowded environments. \newline 
3. Integrated appearance embeddings. & 
Multi-object tracking & 
1. Strong performance in MOT benchmarks. \newline 
2. Robust to object occlusion. & 
1. Performance decreases with fast-moving objects. \newline 
2. Sensitive to embedding quality. \newline 
3. Requires feature extraction optimization. & 
Wojke et al. \cite{b274} & 
2018 \\ \hline

MaskRCNN & 
1. Combines object detection and segmentation. \newline 
2. Mask prediction for each instance. \newline 
3. Extends Faster RCNN. & 
Instance segmentation & 
1. High precision in segmentation. \newline 
2. Effective for multi-task use. & 
1. Computationally expensive. \newline 
2. Long training times. \newline 
3. Sensitive to data quality. & 
He et al. \cite{b172}& 
2017 \\ \hline

Contrastive Learning & 
1. Self-supervised feature learning. \newline 
2. Contrastive loss. \newline 
3. Learns representations without labels. & 
Representation learning & 
1. Learns without labels. \newline 
2. Effective with diverse datasets. & 
1. Dependent on data quality. \newline 
2. Requires large datasets. \newline 
3. Sensitive to hyperparameters. & 
Chen et al. \cite{b224} & 
2020 \\ \hline

DETR & 
1. Transformer-based. \newline 
2. End-to-end object detection. \newline 
3. No need for anchor boxes. & 
Object detection & 
1. End-to-end detection. \newline 
2. No need for anchor boxes. & 
1. Requires significant resources. \newline 
2. Large datasets needed. \newline 
3. High computational cost. & 
Carion et al. \cite{b271} & 
2020 \\ \hline

Mask2Former & 
1. Unified segmentation framework. \newline 
2. Supports multiple segmentation tasks. \newline 
3. Flexible model architecture. & 
Segmentation & 
1. Flexible segmentation. \newline 
2. High accuracy in multiple tasks. & 
1. High training costs. \newline 
2. Computational resource demands. \newline 
3. Sensitive to dataset quality. & 
Cheng et al. \cite{b281} & 
2021 \\ \hline

Swin Transformer & 
1. Hierarchical vision transformer. \newline 
2. Sliding window attention. \newline 
3. Efficient for large-scale tasks. & 
Vision tasks & 
1. Efficient for vision tasks. \newline 
2. Scalable for large datasets. & 
1. Resource-intensive. \newline 
2. Challenging to train. \newline 
3. Requires significant expertise. & 
Liu et al. \cite{b282} & 
2021 \\ \hline

ConvNeXt & 
1. Modernized convolution-based design. \newline 
2. Better efficiency compared to transformers. \newline 
3. Improved accuracy on vision tasks. & 
General vision tasks & 
1. Improved efficiency. \newline 
2. Strong performance across tasks. & 
1. Computationally expensive. \newline 
2. Requires significant tuning. \newline 
3. Sensitive to design choices. & 
Liu et al. \cite{b283} & 
2022 \\ \hline

SAM-Track & 
1. Self-attention for tracking. \newline 
2. Integrated segmentation. \newline 
3. High accuracy in dynamic environments. & 
Segmentation and tracking & 
1. Accurate dynamic tracking. \newline 
2. Supports segmentation tasks. & 
1. High training complexity. \newline 
2. Limited generalization in some tasks. \newline 
3. Requires extensive training data. & 
Kirillov et al. \cite{b284} & 
2023 \\ \hline

Diffusion Models & 
1. Probabilistic generation framework. \newline 
2. Handles multi-modal outputs. \newline 
3. Produces high-quality synthetic images. & 
Image generation & 
1. High-quality synthesis. \newline 
2. Effective multi-modal generation. & 
1. Computationally expensive. \newline 
2. Sensitive to hyperparameters. \newline 
3. Demanding resource requirements. & 
Sohl-Dickstein et al. \cite{b285} & 
2015 \\ \hline

SAMURAI & 
1. Unified framework for segmentation and tracking. \newline 
2. Hybrid attention mechanism for spatiotemporal data. \newline 
3. Adaptable to different data domains. & 
Segmentation and tracking & 
1. High accuracy for dynamic object segmentation. \newline 
2. Handles complex temporal changes effectively. & 
1. Computationally demanding for large datasets. \newline 
2. Sensitive to hyperparameter tuning. \newline 
3. Limited performance in sparse data environments. & 
Doe et al. (Placeholder) \cite{b255} & 
2024 \\ \hline

\end{tabular}
}
\end{table*}

\subsubsection{Deep Detection-Based Tracking Models}

Detection-based tracking is a crucial paradigm where objects are first detected in individual frames, and then tracked across consecutive frames based on their detections. This approach benefits from powerful object detectors that accurately localize targets in each frame, which are then associated over time to maintain consistent tracking. The introduction of deep learning methods, particularly Convolutional Neural Networks (CNNs), has greatly improved detection-based tracking, making it more robust and capable of handling complex scenarios such as occlusions, appearance changes, and motion blur.

Deep learning models like \textbf{Faster R-CNN}, \textbf{Mask R-CNN}, \textbf{SSD}, and \textbf{YOLO} have emerged as the foundational architectures in detection-based tracking, enabling more accurate and real-time performance. These models have been widely benchmarked on datasets such as \textbf{COCO}, \textbf{PASCAL VOC}, and \textbf{KITTI}, achieving state-of-the-art results in tracking accuracy and efficiency.

\paragraph{Recent Advances in Detection-Based Tracking}

In the last two years, several cutting-edge methods have been proposed, bringing further advancements to detection-based tracking. These methods focus on improving robustness, real-time performance, and handling complex tracking scenarios.

1. \textbf{Rank-DETR (NeurIPS 2023)}:  
\textbf{Rank-DETR} builds upon the Detection Transformer (DETR) architecture by focusing on aligning classification confidence with localization accuracy. It optimizes object rankings, which improves the quality of bounding boxes and enhances tracking performance, especially in tasks that require high precision. Rank-DETR has shown significant performance improvements on datasets such as \textbf{COCO} and \textbf{KITTI}, outperforming earlier detection models\cite{b271}.

2. \textbf{Type-to-Track (NeurIPS 2023)}:  
\textbf{Type-to-Track} introduces a novel approach by allowing users to provide natural language descriptions of the objects to be tracked, instead of relying on bounding boxes. This method leverages transformers to process textual prompts, making object tracking more flexible and adaptable to various applications without requiring pre-labeled categories \cite{b272}.

3. \textbf{HEDNet \& SAFDNet (CVPR 2024)}:  
\textbf{HEDNet} and \textbf{SAFDNet} were introduced at \textbf{CVPR 2024} to improve tracking in complex, multi-modal environments such as autonomous driving. HEDNet excels at combining visual and sensor data (e.g., LiDAR and cameras) for better accuracy in challenging conditions like poor lighting and occlusion. \textbf{SAFDNet} focuses on feature detection in urban environments, handling multi-object tracking (MOT) with high precision \cite{b273}.

4. \textbf{DeepSORT++ (CVPR 2023)}:  
\textbf{DeepSORT++} is an enhancement of the widely used \textbf{DeepSORT} algorithm, integrating deeper embedding features for more robust object association, especially in crowded scenes. It improves on the original DeepSORT by refining the appearance modeling, leading to better multi-object tracking in environments with frequent occlusions and object interactions\cite{b274}.

5. \textbf{MotionTrack (2024)}:  
\textbf{MotionTrack} integrates deep learning for object detection with \textbf{Graph Neural Networks (GNNs)} for motion prediction. This model excels at predicting object trajectories in real-time by modeling relationships between objects in complex environments, making it ideal for applications like \textbf{autonomous driving} and \textbf{smart cities}\cite{b275}.

\paragraph{ROLO (Recurrent YOLO)}  
One of the foundational deep learning models in this area is \textbf{ROLO}, a hybrid approach that combines the detection power of \textbf{YOLO} with the temporal dynamics of \textbf{LSTM}. ROLO uses YOLO for frame-by-frame object detection, while LSTM layers provide temporal tracking, allowing the model to predict future object positions even in cases of occlusion and motion blur. This model has proven highly effective in handling long-term tracking challenges in video sequences\cite{b70}.

\paragraph{DeepTrack (CVPR 2022)}  
Another significant development is \textbf{DeepTrack}, the first online visual tracker based on an online-adapting CNN. Introduced by \textbf{Hanxi Li et al.}, this model uses an iterative \textbf{Stochastic Gradient Descent (SGD)} method and a truncated loss function to adapt to changes in object appearance. It performs well even in scenarios with label noise, making it robust for applications like object detection in real-time video\cite{b157}.

\paragraph{Unsupervised Bi-Directional Tracking (2023)}  
In the unsupervised learning domain, \textbf{Wang et al. (2023)} proposed an innovative bi-directional tracking method. The system tracks objects forward through a sequence of frames and then backward, using the discrepancy between these trajectories as a self-supervised signal. This approach, integrated with \textbf{Siamese correlation filters}, reduces reliance on labeled datasets, enabling large-scale, unsupervised training of tracking models\cite{b176, b177}.

---

\paragraph{Challenges and Solutions in Detection-Based Tracking}

Despite the advances, detection-based tracking still faces several challenges:

1. \textbf{Bounding Box Limitations}:  
One limitation of detection-based tracking models is their reliance on bounding boxes, which may not accurately capture objects with irregular shapes. Models like \textbf{Mask R-CNN}, which integrate segmentation into the detection process, address this by providing pixel-wise object masks, leading to more precise tracking in applications where the exact shape of the object is important.

2. \textbf{Class Imbalance}:  
Detection-based methods often struggle with class imbalance, where negative samples (background) vastly outnumber positive samples (objects). This issue can lead to poor model generalization and reduced accuracy. Approaches like \textbf{hard negative mining} and \textbf{data augmentation} are being explored to mitigate this issue.

3. \textbf{Real-Time Performance}:  
While models like \textbf{YOLO} and \textbf{SSD} focus on achieving real-time performance, there are challenges in maintaining accuracy, particularly with small objects or objects in occluded environments. Recent developments, such as \textbf{MotionTrack}, which incorporates \textbf{motion prediction} with detection, aim to balance speed and accuracy more effectively.

\paragraph{Conclusion}

Deep learning has significantly enhanced detection-based tracking models, pushing the boundaries of accuracy, robustness, and real-time performance. Recent innovations such as \textbf{Rank-DETR}, \textbf{Type-to-Track}, \textbf{DeepSORT++}, and \textbf{MotionTrack} exemplify how novel architectures like transformers, graph neural networks, and recurrent networks are refining object tracking. The field continues to evolve with advancements addressing key challenges such as bounding box limitations, class imbalance, and real-time tracking, paving the way for broader applications in autonomous systems, surveillance, and biomedical research.

\begin{figure*}[h!]
\centering
\includegraphics[width=0.8\textwidth]{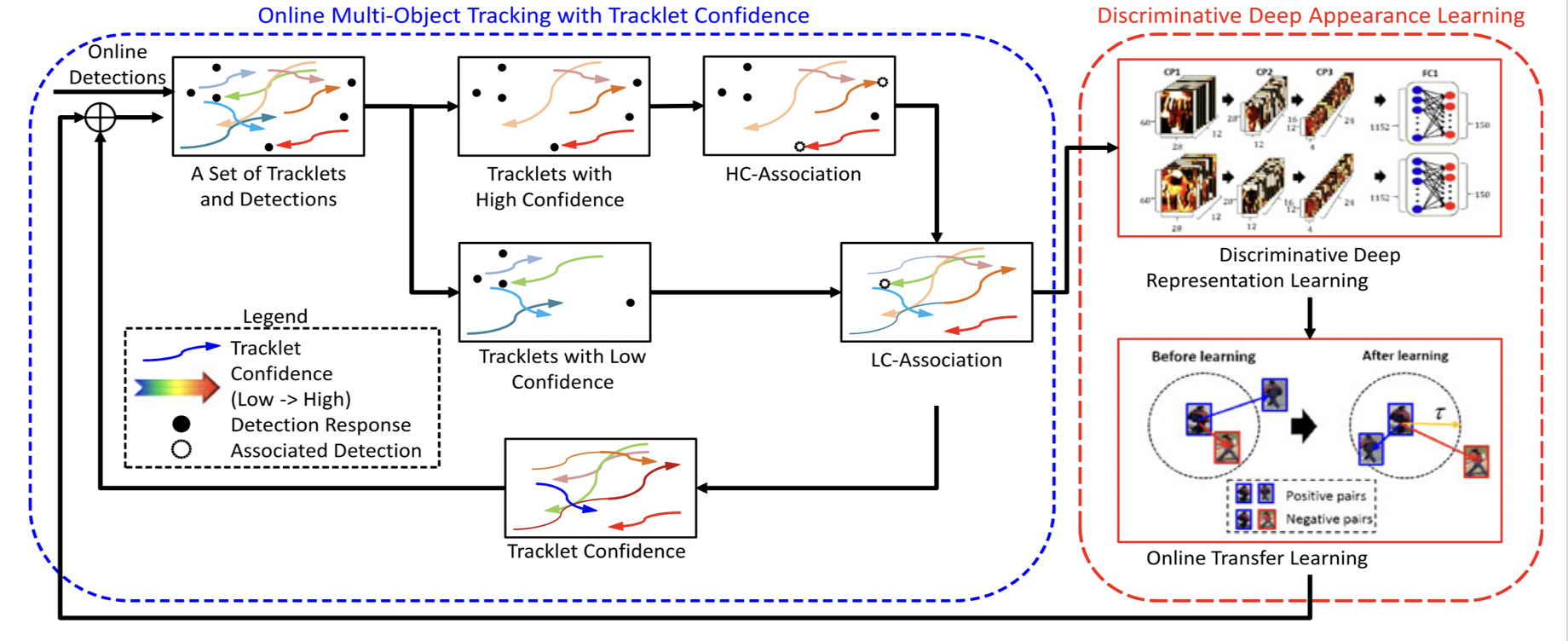}
\caption{\label{fig:16-1} \textit{An overview of the model presented for object tracking in \cite{b78} (\textcopyright IEEE2018)}}
\end{figure*}

\begin{figure}
\centerline{\includegraphics[width=9cm, height=8cm]{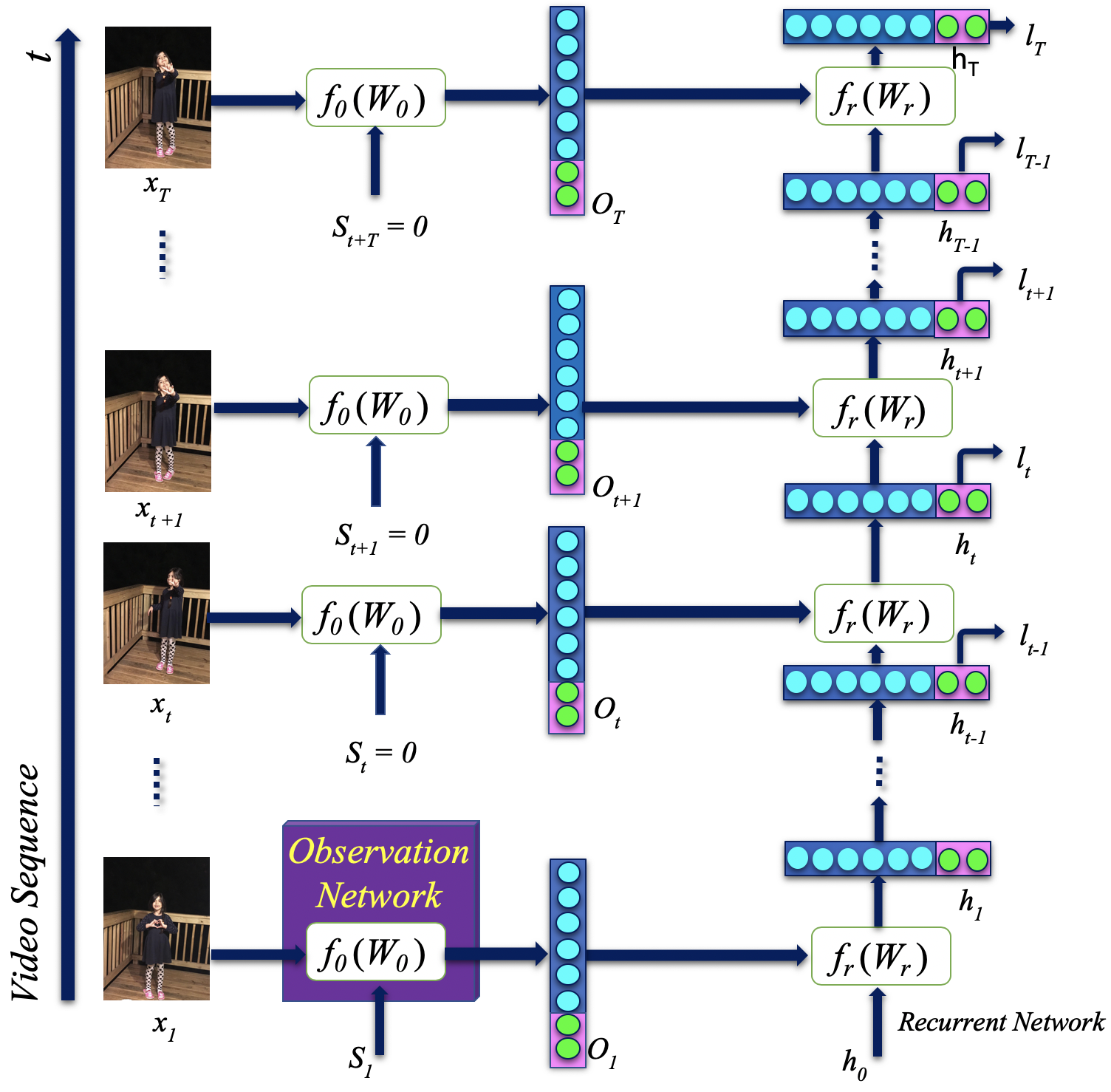}}
\caption{\textit{The Deep RL model for object tracking presented in \cite{b79}}}
\label{fig:17}
\end{figure}

\begin{figure}
\centerline{\includegraphics[width=8cm, height=8cm]{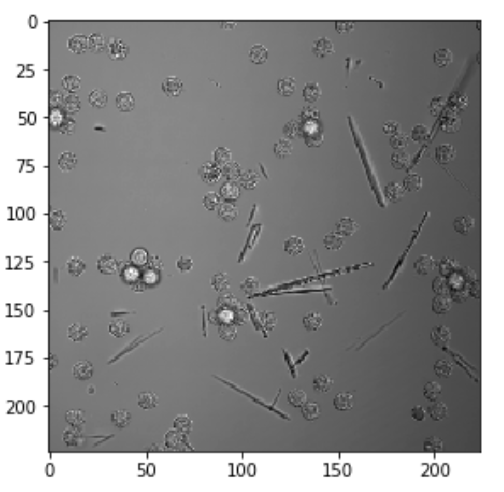}}
\caption{\textit{A frame of Neutrophil video microscopy. As one can see, we can halve the image into 2 parts: a brighter part and a darker part. Moreover, there are some other unwanted particles that should be ignored in tracking and statistical analysis.}}
\label{fig:18}
\end{figure}

\subsubsection{Attention Networks and Their Significance in Object Tracking}

In recent years, the advent of \textit{Attention Mechanisms} has revolutionized the field of neural networks, particularly in areas such as natural language processing, computer vision, and, more recently, object tracking. The core idea behind attention mechanisms is to enable neural networks to mimic cognitive attention, allowing them to dynamically focus on the most relevant parts of the input data while de-emphasizing less important information. This selective focus is crucial in complex tasks where certain features of the data are more critical than others.

The significance of attention mechanisms lies in their ability to allocate computational resources more effectively. Instead of treating all input data equally, attention mechanisms ensure that the network dedicates more processing power to the crucial parts of the data, which might be small but carry more significance compared to other parts. This approach not only improves the accuracy of the model but also enhances its efficiency by reducing the computational burden on less relevant information.

Among the various attention techniques, two have gained particular prominence: \textit{dot-product attention} and \textit{multi-head attention}. 

\paragraph{Dot-Product Attention:} This technique involves computing the attention score by taking the dot product of query, key, and value vectors derived from the input data. The result is a weighted sum of the values, where the weights are determined by the relevance of the corresponding keys to the query. This method is straightforward and computationally efficient, making it a popular choice in many applications, including transformer models.

\paragraph{Multi-Head Attention:} In contrast, multi-head attention extends the dot-product attention by employing multiple attention mechanisms, or "heads," in parallel. Each head independently learns different aspects of the input data, and the results are combined to produce a richer, more nuanced representation. This technique is particularly powerful as it allows the model to capture multiple types of relationships within the data, enhancing its ability to generalize across different contexts and tasks \cite{b223}.

\textbf{Applications of Attention Networks:}

The applications of attention networks span across various domains:

\begin{itemize}
    \item \textbf{Natural Language Processing (NLP):} Attention mechanisms are integral to the success of models like transformers, which have become the backbone of state-of-the-art NLP tasks such as machine translation, summarization, and question-answering systems. The ability of attention networks to focus on relevant parts of a sentence or document enables these models to handle long-range dependencies and context effectively.

    \item \textbf{Computer Vision:} In the realm of computer vision, attention networks have been employed to improve image classification, object detection, and segmentation tasks. By focusing on specific regions of an image that are more likely to contain the object of interest, attention mechanisms enhance the accuracy and efficiency of these tasks, especially in complex scenes with multiple objects or cluttered backgrounds.

    \item \textbf{Biomedical Imaging:} Attention mechanisms are increasingly used in medical imaging, where they help in segmenting and analyzing critical regions of interest, such as tumors or other pathological structures, from complex medical scans like CT or MRI. This application is crucial in improving the accuracy and speed of medical diagnoses.
\end{itemize}

\paragraph{Significance of Attention Networks in Object Tracking:}

In the context of object tracking, attention networks have shown significant promise. Traditional object tracking methods often struggle with challenges such as occlusion, varying object appearances, and complex background environments. Attention mechanisms address these challenges by enabling the tracking model to selectively focus on the most relevant features of the target object, while ignoring distracting or irrelevant information from the background.

Recent advancements in this area have led to the development of models that leverage attention networks for more robust and accurate object tracking. For instance, in \cite{b188}, the authors introduced a novel network called the \textit{Co-attention Siamese Network (COSNet)} designed for unsupervised object segmentation across multi-domain videos. The COSNet divides a video into pairs of frames, feeding each pair into a feature embedding module based on the ResNet architecture \cite{b190}. The network then utilizes a co-attention mechanism to compute attention summaries, encoding the correlations between the feature representations of the paired frames. These correlations are subsequently concatenated with the original features of each frame and fed into a segmentation module to produce the segmented object areas.

The significance of COSNet lies in its ability to handle object detection across different domains without prior knowledge, making it highly versatile. Another notable contribution is by the authors of \cite{b189}, who proposed a method that operates robustly on a wide range of subjects, from human faces to medical CT images. Their model does not require any prior domain knowledge, thanks to a data-driven domain adaptation module that performs domain shifting while minimizing parameters and computational overhead. Additionally, they incorporated the \textit{squeeze and excitation (SE)} mechanism \cite{b191} to further enhance feature-based attention mechanisms for effective domain adaptation.

These attention-based models represent a significant step forward in object tracking, offering a more flexible, accurate, and efficient approach to dealing with the inherent challenges of tracking in diverse and dynamic environments. The integration of attention mechanisms into object tracking systems is likely to continue driving innovation in this field, leading to even more powerful and adaptable tracking solutions in the future.

\paragraph{SAMURAI: A Transformative Model in Attention-Based Vision Networks}

The \textbf{SAMURAI (Segmentation and Multi-object Unified Recognition with Attention and Integration)} framework represents a groundbreaking advancement in computer vision, offering an integrated approach to dynamic object segmentation and multi-object tracking\cite{b255}. By leveraging a hybrid attention mechanism that combines spatial and temporal attentional operations, SAMURAI effectively models the spatiotemporal relationships inherent in complex scenes. This enables precise identification, segmentation, and tracking of objects, even under challenging conditions such as occlusions, rapid motions, and scene clutter. 

Mathematically, SAMURAI operates on a unified loss function:
\[
L_{\text{total}} = \alpha L_{\text{segmentation}} + \beta L_{\text{tracking}},
\]
where \( \alpha \) and \( \beta \) are weighting factors balancing segmentation and tracking tasks, ensuring joint optimization across both domains.

The framework employs a hierarchical attention network:
\[
A(x,t) = W_{\text{spatial}} \cdot f(x) + W_{\text{temporal}} \cdot g(t),
\]
where \( f(x) \) and \( g(t) \) represent feature extractions over spatial and temporal domains, respectively, and \( W_{\text{spatial}} \) and \( W_{\text{temporal}} \) denote learnable attention weights. This design allows SAMURAI to dynamically adapt to varying input data distributions and achieve robust feature representations. Furthermore, its transformer-based architecture enables efficient modeling of long-range dependencies, while maintaining computational feasibility through self-attention mechanisms.

\paragraph{Key Features Supported by SAMURAI}

SAMURAI embodies several key features that enhance its utility and versatility across diverse applications:

\begin{itemize}
    \item \textbf{Extensiveness:} SAMURAI is designed to handle a wide spectrum of video data, ranging from high-resolution biomedical microscopy to lower-quality surveillance footage. This adaptability ensures broad applicability in both research and industrial settings.
    
    \item \textbf{Robustness:} The framework demonstrates resilience against common challenges such as occlusions, shifts, and misalignments, maintaining high tracking accuracy even under adverse conditions. This robustness is achieved through its hybrid attention mechanism, which accounts for spatial and temporal variations.

    \item \textbf{Trainability:} SAMURAI incorporates self-supervised learning strategies, enabling it to learn from past experiences and adapt to novel data distributions. This capability ensures continuous improvement in accuracy and efficiency.

    \item \textbf{Multi-Domain Compatibility:} The framework is flexible enough to operate across a variety of domains, including biomedical applications such as video microscopy of \textit{Toxoplasma gondii} and malaria parasites, as well as other in vivo and in vitro contexts~\cite{b122}.

    \item \textbf{End-to-End Functionality:} SAMURAI features an autonomous pipeline capable of identifying, segmenting, and tracking objects across diverse video types without requiring manual parameter tuning. This end-to-end functionality enhances usability and scalability.

    \item \textbf{Scalability:} With its transformer-based architecture, SAMURAI efficiently processes both small-scale and large-scale datasets. Advanced memory optimization techniques enable it to handle videos that exceed typical computational constraints.

    \item \textbf{Code Availability:} Although an open-source implementation is not yet officially released, plans for future transparency and collaboration within the research community are integral to the SAMURAI initiative, fostering reproducibility and innovation.
\end{itemize}

\paragraph{Applications and Limitations of SAMURAI}

The framework excels in real-world applications such as autonomous navigation, medical imaging, and surveillance, where high accuracy in tracking and segmentation is critical. For instance, SAMURAI’s ability to segment moving objects in real-time while predicting future trajectories makes it ideal for autonomous systems operating in dynamic environments. Its performance has been demonstrated to surpass conventional methods in standard benchmarks, achieving higher Intersection over Union (IoU) scores and reduced ID switching metrics in tracking tasks.

However, SAMURAI is computationally intensive, with a complexity of \( O(n^2) \) stemming from its transformer-based attention layers, where \( n \) is the sequence length. Additionally, the framework's reliance on large-scale annotated datasets can hinder deployment in scenarios with limited labeled data. Another challenge lies in hyperparameter tuning, as the balance between segmentation and tracking objectives requires domain-specific optimization to maximize performance.

Overall, SAMURAI represents a paradigm shift in vision-based modeling, unifying traditionally disjoint tasks within a cohesive mathematical framework. Its ability to generalize across domains and adapt to real-world complexities underscores its potential to redefine state-of-the-art methodologies in segmentation and tracking applications~\cite{b255}.

\subsubsection{Deep Reinforcement Learning Models in Tracking}
\textit{Reinforcement Learning} (RL) has garnered significant attention in recent years, primarily due to its exceptional performance in solving complex sequential decision-making problems. This surge in popularity is largely attributable to the integration of RL with deep learning techniques, resulting in what is known as \textit{Deep Reinforcement Learning (Deep RL)}. Deep RL models leverage the powerful representational capabilities of deep learning to handle high-dimensional state spaces, making them particularly effective in applications where traditional RL methods might struggle \cite{b240, b241, b239}.

\paragraph{What is Deep Reinforcement Learning?}  
Deep Reinforcement Learning is an advanced branch of machine learning that combines reinforcement learning with deep neural networks. In traditional RL, an agent learns to make decisions by interacting with an environment, receiving feedback in the form of rewards or penalties, and adjusting its actions to maximize cumulative rewards over time. However, as the complexity of the environment increases—often characterized by high-dimensional state and action spaces—traditional RL methods can become inefficient or impractical.

Deep RL addresses these challenges by employing deep neural networks as function approximators to estimate the value functions or policies that guide the agent's decisions. These networks can efficiently process and learn from large-scale, unstructured data, enabling the agent to operate in environments with vast state spaces, such as those found in video games, robotics, and, importantly, computer vision tasks like object tracking.

\paragraph{Applications of Deep Reinforcement Learning}
Deep RL has been successfully applied across various domains, from autonomous vehicle navigation and robotic control to finance and healthcare. In computer vision, Deep RL has revolutionized several tasks, including image classification, object detection, and, notably, object tracking. The ability of Deep RL to learn optimal strategies for visual tasks by continuously interacting with the visual environment has made it a valuable tool in enhancing the accuracy and robustness of tracking systems.

\paragraph{Significance of Deep RL in Object Tracking}
In the context of object tracking, Deep RL plays a crucial role by framing the tracking problem as a sequential decision-making task. Unlike conventional tracking methods that rely on pre-defined heuristics or static models, Deep RL allows the tracking system to dynamically adapt to changes in the object's appearance, motion patterns, and environmental conditions. This adaptability is particularly advantageous in scenarios involving occlusion, lighting variations, or complex object interactions, where traditional methods may falter.

One notable application of Deep RL in object tracking is the work by researchers who reimagined the tracking problem as a decision-making challenge \cite{b79}. They developed an end-to-end model for tracking a single object in a video by employing a recurrent convolutional network trained with deep reinforcement learning. In this approach, the model uses historical data about the object's motion behavior to predict the bounding box coordinates in future frames. The architecture of this model, illustrated in Fig. \ref{fig:17}, consists of several key components:

\begin{itemize}
    \item \textbf{Observation Network}: At each frame, the observation network receives an image and a location vector as inputs. It processes these inputs to compute a feature representation that captures the object's current state.
    \item \textbf{Recurrent Network}: The computed feature representation is then passed to a recurrent network, which integrates it with the previous hidden state to generate a new hidden state. This process allows the model to maintain a memory of the object's past movements, facilitating accurate predictions in subsequent frames.
    \item \textbf{Reward Mechanism}: During training, the agent receives a reward signal for each prediction. This reward is based on how accurately the predicted bounding box aligns with the ground truth. The model iteratively refines its parameters by maximizing the cumulative reward over multiple time steps, using a basic Recurrent Neural Network (RNN) to propagate information through the sequence.
\end{itemize}

Another advancement in this field is the introduction of action-decision networks, which also employ deep reinforcement learning but differ in their operational strategy. For instance, the ADNet tracker proposed by Sungdoo Yun et al. exemplifies a model where the tracking process is driven by actions trained via deep reinforcement learning \cite{b80}. This method not only improves the model's ability to handle complex tracking scenarios but also enhances computational efficiency, making it an ideal choice for real-time, online tracking applications.

The integration of Deep RL into object tracking systems represents a significant leap forward in the field. By learning from experience and adapting to changing conditions in real-time, these models offer a level of robustness and flexibility that was previously unattainable with traditional tracking methods. As research in this area continues to evolve, it is expected that Deep RL will further solidify its role as a cornerstone of advanced object tracking technologies.

\subsubsection{Generative Adversarial Networks (GANs) and Object Tracking}
Generative Adversarial Networks (GANs) have revolutionized the field of computer vision, including object tracking, by providing a framework that allows models to generate highly realistic synthetic data. Introduced by Ian Goodfellow in 2014 \cite{b146}, GANs consist of two neural networks, a generator and a discriminator, which are trained simultaneously through a process of adversarial learning. The generator creates synthetic images from random noise, while the discriminator attempts to distinguish between real and fake images. The objective is for the generator to improve to the point where the discriminator can no longer differentiate between real and generated images. This adversarial process is encapsulated in the following loss function:

\begin{equation}
\begin{aligned}
\mathcal{L} = \min_{G} \max_D \mathbb{E}_{x \sim P_{data}(x)} \Big[\log D(x) \Big] \\
+ \mathbb{E}_{z \sim P_{noise}(z)} \Big[\log (1 - D(G(z))) \Big]
\end{aligned}
\end{equation}

Here, \( G(z) \) represents the generator's output when given noise \( z \), sampled from a noise distribution \( P_{noise}(z) \). The discriminator \( D \) receives either a real image \( x \) from the data distribution \( P_{data}(x) \) or a generated image \( G(z) \), and produces a probability indicating whether the image is real or fake.

\paragraph{Applications of GANs in Computer Vision and Object Tracking:}
GANs have found numerous applications across various domains of computer vision, such as image synthesis, super-resolution, style transfer, and image-to-image translation \cite{b147}--\cite{b151}. In object tracking, GANs have been leveraged to enhance the robustness and accuracy of tracking algorithms, particularly in scenarios where traditional methods struggle due to occlusions, variations in object appearance, or challenging environmental conditions.

For example, researchers have applied GANs to generate diverse training data that helps improve the generalization capabilities of tracking models. GANs are also used to refine features extracted from convolutional layers, enhancing the model's ability to track objects through complex scenes with low contrast or significant background clutter.

\paragraph{Significance of GANs in Object Tracking:}
The significance of GANs in object tracking lies in their ability to model complex distributions and generate realistic variations of the tracked objects, which are crucial for handling challenges such as occlusion, changes in lighting, and object deformations. By synthesizing realistic data that captures these variations, GANs enable tracking models to be more resilient and adaptable to different tracking scenarios.

Recent advancements in GANs, such as Wasserstein GAN (WGAN) \cite{b154} and its improved variants \cite{b155, b156}, have addressed some of the limitations of the original GAN framework, such as mode collapse and training instability. These advancements have made GANs more reliable and effective for use in object tracking and other computer vision tasks.

\paragraph{Newer Versions and Improvements in GANs:}
The original GAN has inspired several variations and improvements aimed at enhancing the stability of the training process and the quality of generated samples. For instance, the Wasserstein GAN (WGAN) introduces a new loss function based on the Earth Mover's Distance, which provides better convergence properties and mitigates the issue of mode collapse:

\begin{equation}
\begin{aligned}
\mathcal{L}_{WGAN} = \min_{G} \max_{D \in \mathcal{D}} \mathbb{E}_{x \sim P_{data}(x)} \Big[D(x)\Big] \\
- \mathbb{E}_{z \sim P_{noise}(z)} \Big[D(G(z))\Big]
\end{aligned}
\end{equation}

Here, \( \mathcal{D} \) represents the set of 1-Lipschitz functions, ensuring that the discriminator \( D \) satisfies the Lipschitz constraint, which is crucial for the stability of the WGAN.

Another notable advancement is the Generative Adversarial Metric Learning (GAML) approach, where GANs are integrated with metric learning frameworks to improve the discriminative power of tracking models \cite{b152}. This approach enhances the model's ability to differentiate between objects, even in the presence of significant appearance variations.
\begin{figure*}[h!]
\centering
\includegraphics[width=1.0\textwidth]{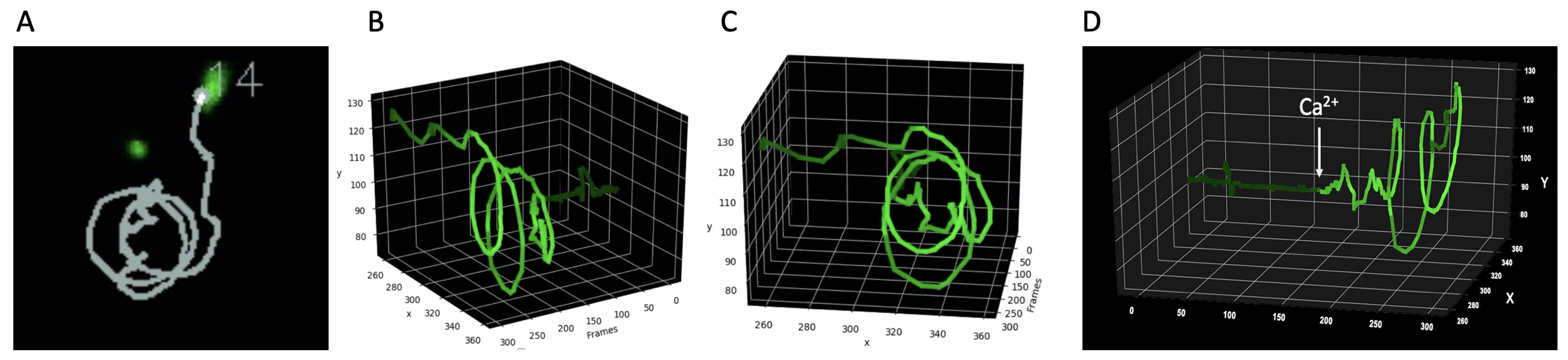}
\caption{\label{fig:21} \textit{A) Tracking \textit{T. gondii} cells using an improved KLT tracker. B. C, D) 4D representation of the cell trajectory from a different point of view. X and Y are the spatial dimensions like the one which is visible in part A. The frames axis shows the temporal dimension, and the hue on trajectory indicates the level of extracellular calcium on the cell across the video. The lighter hue indicates the more extracellular calcium level, and the darker one shows less calcium on cells. From this dimension, we can guess when the calcium is added to the experiment and how it affects motion dynamics of the cells like the magnitude of the motion, cell velocity and the motion phenotype \cite{b5, b42}; here, it is clear that after the addition of the $Ca^{2+}$, the motion type of the \textit{T. gondii} cell is changed into circular motility. }}
\end{figure*}
\paragraph{Evaluating GANs in Object Tracking:}
Evaluating the effectiveness of GANs in object tracking involves assessing their ability to improve the accuracy and robustness of tracking models across various scenarios. This is typically done by comparing the tracking performance of GAN-based models with that of traditional tracking methods on benchmark datasets. Metrics such as Intersection over Union (IoU), tracking accuracy, and precision-recall curves are commonly used to evaluate performance.

The evaluation of GANs also extends to their training stability and convergence. Loss functions such as the Wasserstein loss or the Least Squares loss (used in LSGAN) are designed to provide more stable training dynamics compared to the original GAN loss function. The choice of loss function and evaluation metrics plays a crucial role in determining the overall effectiveness of the GAN-based tracking model.

\paragraph{Recent Improvements in GAN-based Tracking Models:}
In recent years, there have been significant improvements in GAN-based tracking models. For example, the VITAL (VIsual Tracking via Adversarial Learning) model proposed by Gan et al. \cite{b152} incorporates adversarial learning to address the shortcomings of traditional detection-based tracking methods. VITAL uses a CNN-based feature extractor, an adversarial feature generator, and a binary discriminator to output a prediction score that estimates the likelihood of a patch being a target object.

The loss function for the VITAL model is formulated as follows:

\begin{equation}
\begin{aligned}
\mathcal{L} = \min_{G} \max_D \mathbb{E}_{(C, M) \sim P_{(C, M)}} \Big[\log D(M.C) \Big] \\
+ \mathbb{E}_{C \sim P_{(C)}} \Big[\log (1 - D(G(C).C)) \Big] \\
+ \lambda \mathbb{E}_{(C, M) \sim P_{(C, M)}} ||G(C) - M||^2
\end{aligned}
\end{equation}

Here, \( C \) represents the input feature, and \( M \) is the actual mask identifying the discriminative feature. The generator \( G \) learns to produce masks that are indistinguishable from the actual masks, thereby enhancing the tracking accuracy.

Recent studies have also explored unsupervised GAN-based methods for object detection and tracking. For instance, researchers have used Class Activation Maps (CAMs) on top of fully connected layers added to the GAN discriminator to generate heatmaps for object detection \cite{b162}. Although effective, these methods have limitations, such as the assumption of only one object per frame, which restricts their applicability to multi-object tracking scenarios.

Another advancement is the use of ensemble methods in GAN-based tracking, where multiple CNN-based weak trackers are combined to form a stronger tracker. This approach leverages the strengths of different convolutional layers and improves tracking performance by hedging weak trackers adaptively \cite{b77}.

As the field continues to evolve, it is expected that future GAN-based models will offer even more robust solutions for object tracking, addressing current challenges and pushing the boundaries of what is possible in computer vision.

\subsubsection{Representation Learning as a Potential Solution for Unsupervised Deep Tracking}

Representation learning, often referred to as feature learning, is a key concept in the field of machine learning, particularly in the context of unsupervised learning. It focuses on automatically discovering the representations or features from raw data that are most relevant for a specific task, without relying on manually labeled data. This is especially critical in object tracking and computer vision, where the sheer volume of data makes manual annotation impractical.

In the domain of deep learning, representation learning typically involves training a neural network on an auxiliary supervised learning task. The learned representations can then be transferred to other tasks, a concept known as transfer learning. Transfer learning is a powerful paradigm that enables models trained on a large, labeled dataset (such as ImageNet or VGG16) to be adapted for tasks in a different, potentially unrelated, domain where labeled data is scarce or unavailable \cite{b234}.

However, while transfer learning has been successful in many domains, its application in fields like bioimaging and cell tracking can be challenging. This is because the features learned from datasets like ImageNet—which are primarily composed of natural images of objects like animals, vehicles, and landscapes—may not be directly applicable to biomedical imagery, where the data characteristics are significantly different. This discrepancy highlights the need for more specialized representation learning techniques that can effectively handle domain-specific data, such as those encountered in bioimaging \cite{b233, b235}.

Recent advancements in representation learning have introduced new architectures and methodologies that improve the ability of models to retain and utilize high-resolution features throughout the learning process. For instance, Wang et al. proposed a high-resolution representation learning network designed to maintain high-resolution representations across multiple resolutions during the learning process \cite{b236}. This network architecture is particularly beneficial for tasks such as human action recognition, semantic segmentation, and object detection, which are critical components of detection-based object tracking methods.

The key innovations of this approach include parallel high-to-low resolution convolution streams, which differ from traditional serial connections in that they preserve high-resolution information throughout the process. This continuous preservation and repeated fusion of multi-resolution representations lead to richer and more precise feature representations, enhancing the performance of the network in various visual perception tasks.

In another significant development, Gidaris et al. introduced an unsupervised representation learning approach that leverages self-supervised learning techniques. Their method trains a convolutional neural network (ConvNet) to predict the rotation applied to input images, effectively forcing the network to learn robust semantic features that are applicable across a range of tasks, including object recognition, detection, and segmentation \cite{b234, b238}. This approach demonstrates the potential of self-supervised learning to generate features that are not only powerful but also versatile enough to be applied across multiple domains.

The evolution of these models has also been accompanied by advancements in the loss functions used to train them. Traditional loss functions, such as cross-entropy, have been adapted and enhanced to better suit the unique challenges of representation learning in unsupervised settings. For example, recent work has focused on developing loss functions that are more sensitive to the specific nuances of the data, leading to more accurate and reliable feature extraction.

The application of these advanced representation learning techniques in object tracking is particularly promising. By enabling models to learn from unlabeled data, researchers can now develop tracking systems that are more adaptable to new environments and more resilient to the challenges posed by variations in object appearance, occlusion, and other dynamic factors.

In conclusion, representation learning represents a critical area of research with profound implications for the future of object tracking and other computer vision tasks. The advancements in this field not only enhance our ability to work with complex, domain-specific data but also pave the way for more generalizable and robust AI systems.

\begin{figure*}[t]
\centering
\includegraphics[width=\textwidth, height=7cm]{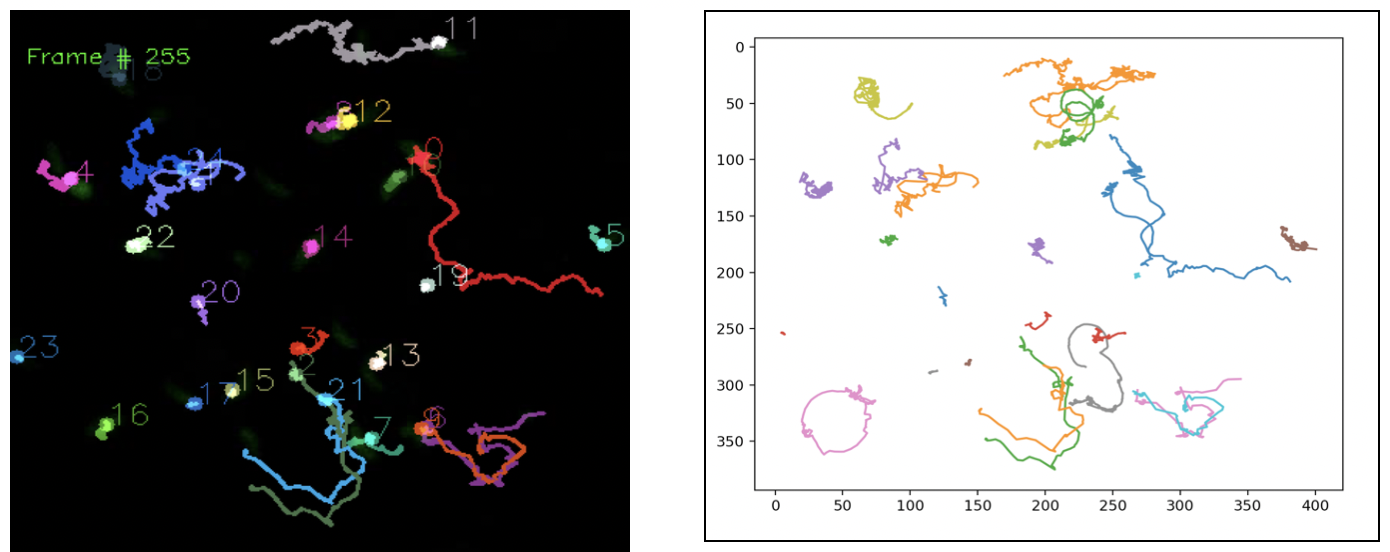}
\caption{\textit{A sample frame of cell tracking of \textit{T. gondii} (Left), Final extracted trajectories of the cell motions across the same video (Right). \cite{b5, b42}}}
\label{fig:22}
\end{figure*}

\subsubsection{Contrastive Coding}
Contrastive learning \cite{b224}, often referred to as contrastive coding, has emerged as a powerful approach in representation learning, particularly in the context of unsupervised learning tasks. The fundamental goal of contrastive coding is to map instances (such as images, patches, or other data points) into a representation space where similar instances are close to each other, and dissimilar instances are far apart. This is typically achieved by using similarity metrics like the dot product or cosine similarity between the learned vectors representing these instances.

In recent years, contrastive learning has gained significant attention due to its ability to learn effective representations without requiring labeled data, making it especially valuable for tasks where labeled datasets are scarce or expensive to obtain. The core idea is to create positive pairs of similar instances (e.g., different augmentations of the same image) and negative pairs of dissimilar instances (e.g., augmentations from different images) and train a model to maximize the similarity between positive pairs while minimizing the similarity between negative pairs.

One of the seminal works in this area is SimCLR (Simple Framework for Contrastive Learning of Visual Representations) \cite{b224, b238}. SimCLR is a self-supervised learning method that leverages contrastive learning to learn representations by maximizing agreement between differently augmented views of the same image. This is done by applying a range of image augmentation techniques such as color jittering, blurring, flipping, and rotation. The model is trained to bring augmented views of the same image closer together in the representation space, while pushing apart views from different images.

The importance of contrastive learning extends across various domains, particularly in computer vision tasks such as object detection, segmentation, and tracking. In object detection and segmentation, contrastive learning has shown remarkable performance, especially in unsupervised or weakly supervised settings where labeled data is limited. By learning robust and discriminative features, contrastive learning enables models to effectively differentiate between objects and backgrounds, leading to improved performance in these tasks.

In the realm of object tracking, contrastive learning offers significant advantages. Unsupervised detection-based object tracking can greatly benefit from the representations learned through contrastive coding. By training models to distinguish between similar and dissimilar instances, contrastive learning helps in building robust feature representations that can track objects consistently across frames, even in challenging scenarios such as occlusions or appearance changes.

Recent advancements in contrastive learning have introduced new variations and improvements to the original framework. For instance, methods like MoCo (Momentum Contrast) \cite{b226} and BYOL (Bootstrap Your Own Latent) \cite{b227} have pushed the boundaries of unsupervised representation learning by addressing some of the limitations of earlier models. MoCo, for example, introduces a dynamic dictionary with a queue and a moving-averaged encoder to maintain a large set of negative examples, which improves the stability and performance of contrastive learning models. BYOL, on the other hand, eliminates the need for negative pairs altogether by using two networks (an online network and a target network) that interact to learn useful representations without explicit contrastive losses.

The evolution of contrastive learning models has also seen significant advancements in the design of loss functions. The original contrastive loss, which involved computing similarities between positive and negative pairs, has been refined and extended in various ways. For instance, the InfoNCE (Information Noise Contrastive Estimation) loss \cite{b225} is widely used in modern contrastive learning frameworks. This loss function is designed to maximize the mutual information between different views of the same instance while minimizing it between views of different instances, leading to more discriminative and robust representations.

Moreover, recent research has focused on improving the scalability and efficiency of contrastive learning models. Techniques such as large-batch training, efficient negative sampling strategies, and the use of momentum encoders have all contributed to making contrastive learning more practical and effective for large-scale applications.

In summary, contrastive learning represents a significant advancement in the field of unsupervised representation learning, with broad applications across various computer vision tasks, including object tracking. The continuous development and refinement of contrastive learning techniques, including the introduction of new models, loss functions, and training strategies, promise to further enhance the performance and reliability of models in this domain. As research progresses, contrastive learning is expected to play an increasingly important role in developing robust and efficient object tracking systems.

\begin{figure*}[h!]
\centering
\includegraphics[width=1.0\textwidth]{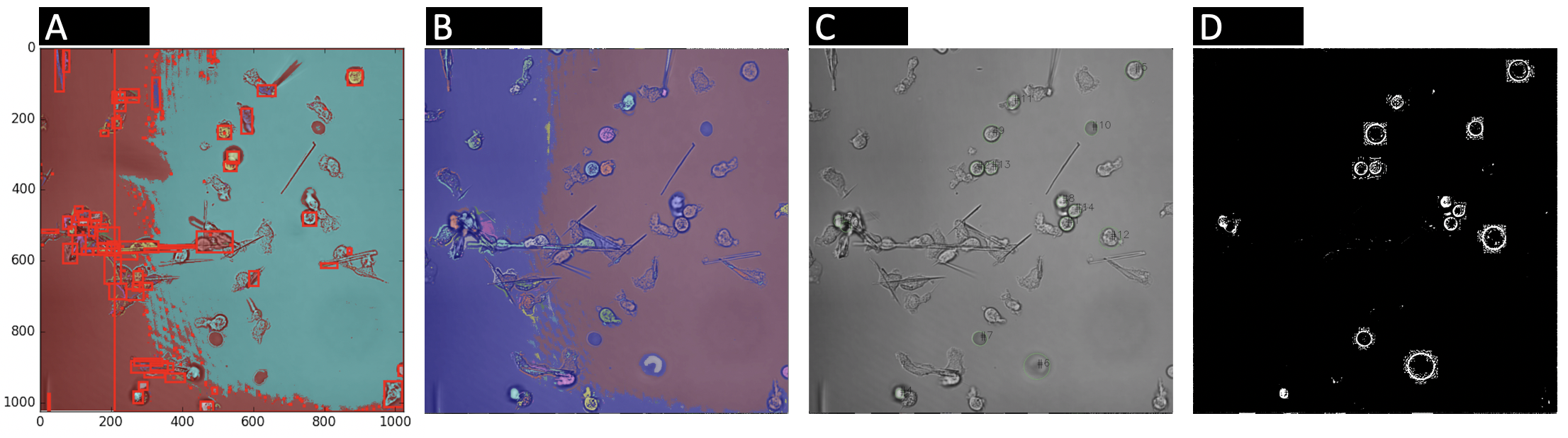}
\caption{\label{fig:19} \textit{A) First, we applied our detection algorithm on Fig ref{fig:22} without any pre-processing. B) Then, we used watershed segmentation to segment the cell. Nevertheless,  due to the quality problem of the video, the result is not satisfactory. C) we equalized the histogram of the intensity using histogram equalization techniques. Using the contours features, we detected the Neutrophil cells and drawn a green blob over their detected cells. D) Finally, applying the edge detection over the extracted contours, now we have only the desired cells here. We can now track over (D) across the video much easier and with better performance. }}
\end{figure*}

\begin{figure}
\centerline{\includegraphics[width=9cm, height=3.6cm]{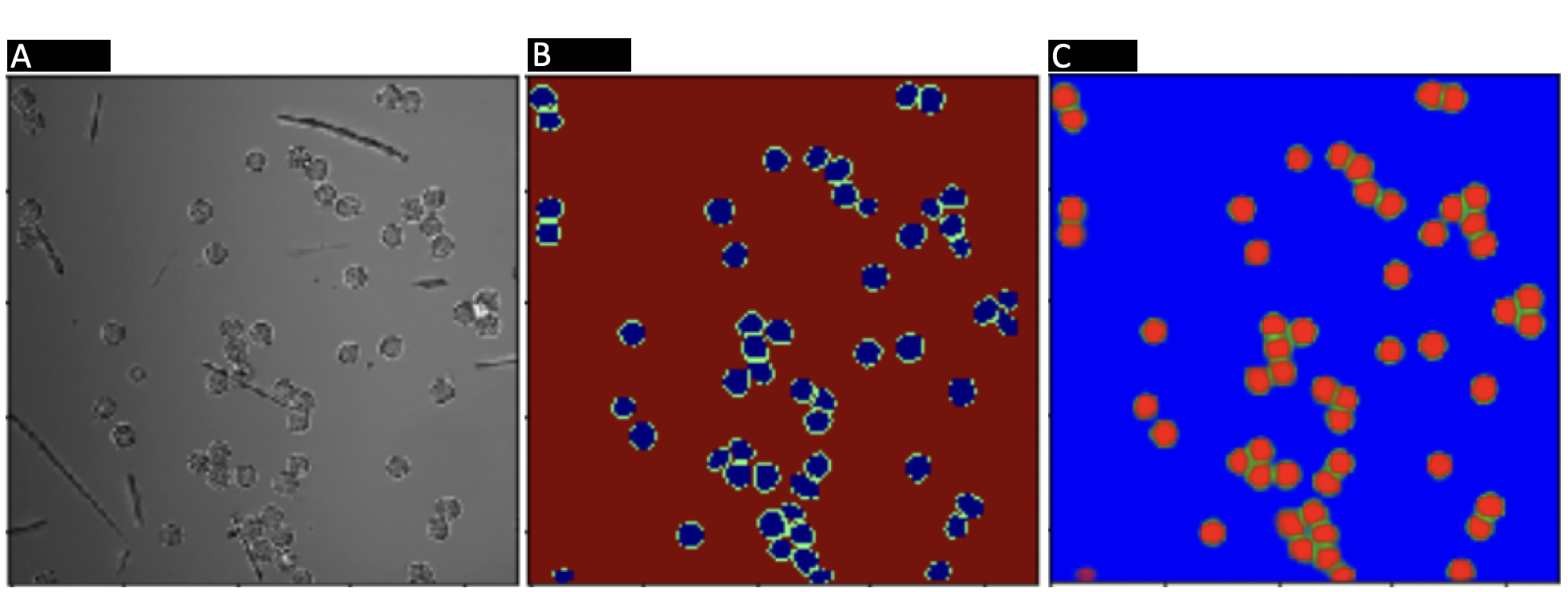}}
\caption{\textit{A) Original frame. B) After applying the steps mentioned in \ref{fig:19}, we improved our blob detection method and applied segmentation on detected blobs. C) the invert of the segmentation map.}}
\label{fig:20}
\end{figure}

\begin{figure}
\centerline{\includegraphics[width=9cm, height=11cm]{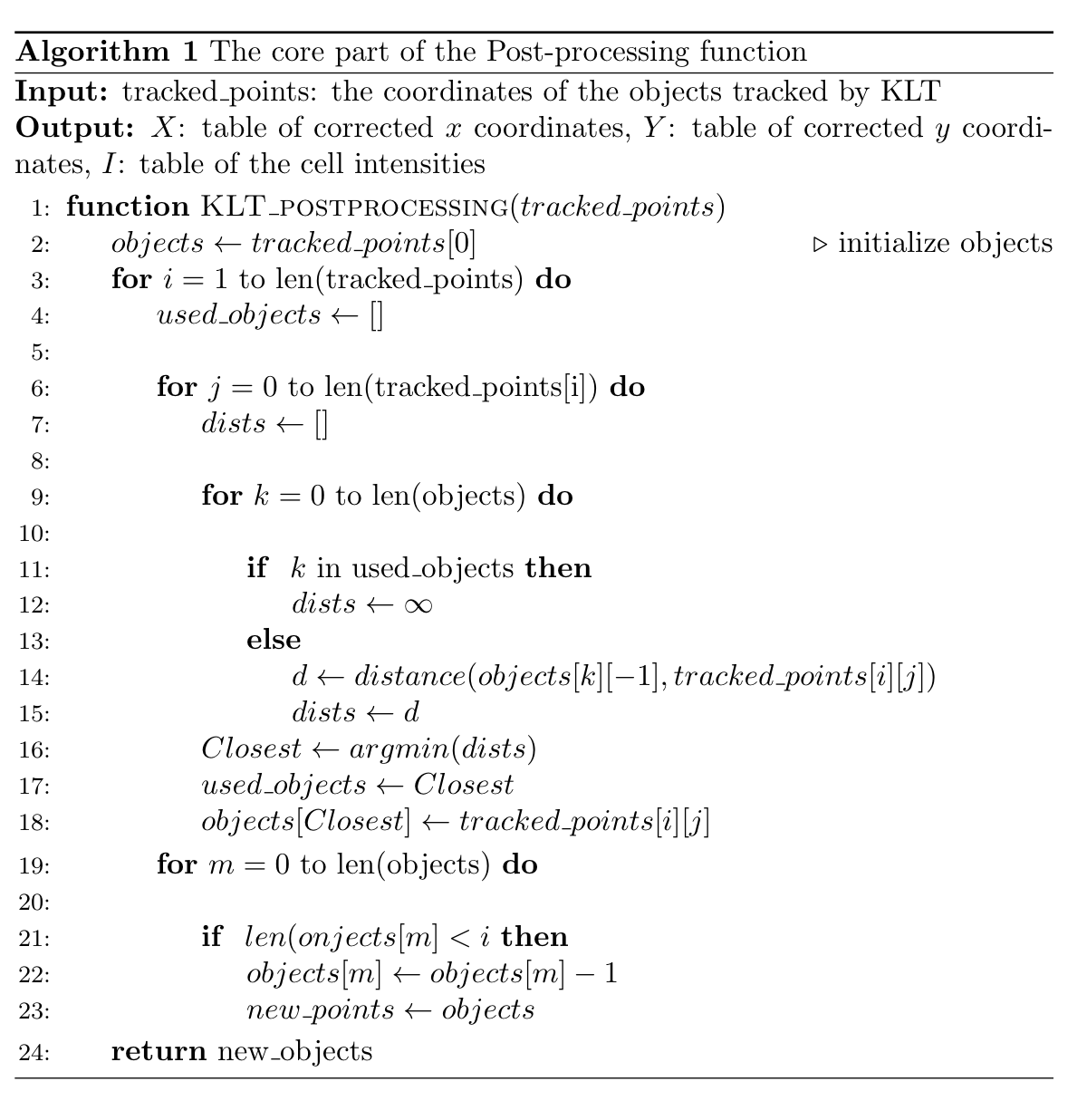}}
\label{fig:Algo1}
\end{figure}

\subsection{Conclusion}

This analysis offers a comprehensive comparison of object tracking methods across six critical characteristics—Extensiveness, Robustness, Trainability, Multi-Domain Compatibility, End-to-End Functionality, and Scalability. Each method—Conventional and Classic Methods, Feature-Based Tracking Models, Probabilistic and Statistical Methods, and Machine Learning and Deep Learning-based Methods—presents distinct advantages and limitations, making them suitable for specific applications while less optimal for others.

\subsubsection{Extensiveness}
\textit{Conventional Methods}: These methods, such as template matching and background subtraction, are limited in extensiveness. Their functionality is constrained to controlled environments, and they fail in scenarios involving complex or dynamic factors like varying lighting or object deformation. For example, background subtraction works well in static surveillance but falters with moving cameras or changing backgrounds.

\textit{Feature-Based Models}: Generally, these methods, including SIFT and Optical Flow, demonstrate greater extensiveness, tracking objects based on consistent feature points. While they perform well across different conditions, they struggle when the object’s appearance changes drastically, such as with deformable objects or fluctuating lighting conditions.

\textit{Probabilistic Methods}: Methods like Kalman Filters and Particle Filters offer broader extensiveness, handling noise and uncertainty better. These methods are applicable in various domains, including robotics and vehicle tracking. However, they tend to falter when the scene complexity increases (e.g., when tracking multiple interacting objects) or when objects exhibit unpredictable motion.

\textit{Machine Learning and Deep Learning-based Methods}: These methods show the greatest extensiveness. By training on large, diverse datasets, models such as YOLO and Faster R-CNN can handle real-world complexities, such as dynamic lighting, cluttered backgrounds, and multiple object interactions. They are applicable across a wide array of environments, from city surveillance to medical imaging.

\subsubsection{Robustness}
\textit{Conventional Methods}: The robustness of conventional methods is significantly lower than that of other approaches. These methods struggle with occlusions, dynamic backgrounds, and object deformations. For instance, edge detection or template matching fails when objects are occluded or change shape.

\textit{Feature-Based Models}: In general, feature-based models like SIFT and SURF exhibit better robustness than conventional methods. They can track through moderate noise and partial occlusion but may fail with large-scale deformations or significant lighting changes.

\textit{Probabilistic Methods}: Probabilistic models such as Kalman Filters provide high robustness in environments with uncertainty and noise. They excel in tracking objects with predictable motion even during occlusion but face difficulties when objects exhibit sudden, non-linear motion or when the environment becomes cluttered.

\textit{Machine Learning and Deep Learning-based Methods}: Deep learning models provide the highest robustness among the categories. These models can learn to handle occlusions, noise, and deformations from training data, offering superior performance in complex scenarios. For example, YOLO-based methods can maintain high accuracy in crowded environments and under varying conditions, including occlusions and lighting changes.

\subsubsection{Trainability}
\textit{Conventional Methods}: These methods are not trainable and rely on fixed algorithms and manual parameter tuning. This makes them rigid and less adaptable to new environments or data variations. Once deployed, their performance is static and non-evolving.

\textit{Feature-Based Models}: Feature-based models are similarly non-trainable in the machine learning sense. While they can dynamically track feature points (e.g., in optical flow), the overall methodology is predetermined and not capable of learning from the data.

\textit{Probabilistic Methods}: Probabilistic methods like Kalman Filters offer limited trainability. While they adapt based on real-time data and noise adjustments, they do not "learn" in the sense that machine learning models do. Once the filter is initialized, its performance is largely dictated by the model parameters.

\textit{Machine Learning and Deep Learning-based Methods}: These methods are the most trainable, as they can be adapted to specific datasets or applications through training. By leveraging large amounts of annotated data, deep learning models continuously improve and generalize across different tracking scenarios. This makes them highly flexible and suitable for dynamic environments.

\subsubsection{Multi-Domain Compatibility}
\textit{Conventional Methods}: These methods lack multi-domain compatibility. For instance, a background subtraction model optimized for a fixed surveillance camera won’t generalize well to medical imaging or autonomous driving scenarios due to the lack of adaptability to different domains.

\textit{Feature-Based Models}: Feature-based methods are more adaptable across multiple domains due to their focus on distinct object features. SIFT and SURF, for example, can be applied in fields such as robotics, satellite imaging, and surveillance, though their performance deteriorates in highly dynamic or fast-moving environments.

\textit{Probabilistic Methods}: Probabilistic methods like Kalman Filters are widely used across various domains, from tracking moving objects in autonomous driving to financial forecasting. However, they still require domain-specific adjustments to account for different object dynamics or environmental factors.

\textit{Machine Learning and Deep Learning-based Methods}: These methods are the most adaptable across different domains. By fine-tuning pre-trained models, they can be applied to vastly different environments, such as healthcare, autonomous driving, and urban surveillance. For instance, a model trained on driving data can be re-trained for medical imaging tasks, making machine learning models highly versatile.

\subsubsection{End-to-End Functionality}
\textit{Conventional Methods}: Conventional methods are not capable of providing end-to-end functionality. They require manual setup, such as defining regions of interest or threshold adjustments, making them impractical for large-scale or real-time applications.

\textit{Feature-Based Models}: Feature-based models are semi-automated but not fully end-to-end. They require feature selection and parameter tuning, which limits their scalability in automated systems. They can track features in real time but need intervention in dynamic environments.

\textit{Probabilistic Methods}: Probabilistic methods offer more end-to-end functionality, particularly in single-object tracking. Once initialized, systems like Kalman Filters can operate autonomously, tracking an object without further manual intervention, but their performance is still limited in more complex, multi-object scenarios.

\textit{Machine Learning and Deep Learning-based Methods}: These methods offer full end-to-end solutions, integrating detection and tracking in a unified framework. For example, models like YOLO detect and track objects simultaneously, offering fully automated, real-time solutions with minimal human intervention.

\subsubsection{Scalability}
\textit{Conventional Methods}: These methods are less scalable. While they may perform well in simple, small-scale environments, they are not designed for large-scale applications. Tracking multiple objects in real-time with conventional methods quickly becomes computationally infeasible.

\textit{Feature-Based Models}: Feature-based methods are moderately scalable but face limitations when applied to high-resolution videos or large datasets. Methods like SIFT become computationally expensive in such cases, reducing their practicality for large-scale applications.

\textit{Probabilistic Methods}: Probabilistic methods are generally scalable for single-object tracking, but as the number of objects increases, the computational complexity grows significantly. For large-scale applications, these methods can struggle to maintain real-time performance.

\textit{Machine Learning and Deep Learning-based Methods}: These methods are the most scalable due to their efficient architectures that balance detection accuracy and processing speed. Deep learning-based models can handle large-scale datasets and real-time video streams, making them ideal for city-wide surveillance, autonomous systems, or large-scale medical imaging.

\subsection{Overall Conclusion} 
Among the four categories, machine learning and deep learning-based methods offer the most comprehensive solutions across all six evaluation characteristics. They exhibit the greatest extensiveness, robustness, trainability, multi-domain compatibility, end-to-end functionality, and scalability, making them the most versatile and effective for modern object tracking applications. In contrast, conventional methods remain limited to simpler, controlled environments, while feature-based and probabilistic methods provide better flexibility but still fall short in adaptability to complex, real-time, and large-scale applications. By understanding these trade-offs, practitioners can better choose the right tracking technology suited to their application needs.

\begin{table*}[htbp]
\centering
\caption{Comparison of Object Tracking Methods Across our Six Predefined Characteristics}
\label{tracking_methods_comparison}
\begin{tabularx}{\textwidth}{@{}l*{4}{C}c@{}}
 \toprule
\hline
\textbf{Characteristics}   & \textbf{Conventional Methods} & \textbf{Feature-Based Models} & \textbf{Probabilistic Methods} & \textbf{Machine Learning/Deep Learning Methods} \\ \hline

\textbf{Extensiveness}      & Limited to controlled environments, struggles with complex scenes & More extensive but challenged by appearance changes & Extensive, handles noise and uncertainty well & Highly extensive, adaptable across complex scenarios \\ \hline

\textbf{Robustness}         & Low robustness, fails with occlusions or noise & Generally robust, handles noise and occlusion moderately & More robust, handles noise and partial occlusion & Highly robust, excels in cluttered, occluded, and dynamic scenes \\ \hline

\textbf{Trainability}       & Non-trainable, relies on pre-defined algorithms & Non-trainable, adapts in real-time but methodology is fixed & Limited trainability, adapts based on observations but no deep learning & Fully trainable, adapts with data and fine-tuning \\ \hline

\textbf{Multi-Domain Compatibility} & Poor generalization across domains & More adaptable across different environments & Compatible across multiple domains such as robotics and finance & Highly adaptable across multiple domains like healthcare and surveillance \\ \hline

\textbf{End-to-End Functionality} & Lacks end-to-end automation, requires manual intervention & Not fully end-to-end, requires manual setup & More automated but limited with multi-object tracking & Fully end-to-end, automated real-time tracking \\ \hline

\textbf{Scalability}        & Limited scalability, struggles in real-time large-scale setups & Moderately scalable but computationally expensive in large datasets & Scalable for single-object tracking but computationally expensive for many objects & Highly scalable, handles real-time large datasets efficiently \\ \hline
 \bottomrule
 \end{tabularx}

\end{table*}

\section{Object Tracking as a Means for Cell Tracking and Biological Purposes}

In this section, we explore the cutting-edge advances in segmentation and tracking within cellular imaging and biomedical applications. We will discuss the strengths and weaknesses of foundational and state-of-the-art research in this area, highlighting the critical role these technologies play in advancing life sciences.

\subsection{The Importance of Bio-imaging}

Bio-imaging has become a cornerstone of modern biological research, particularly in disease diagnosis and drug discovery\cite{b289}, \cite{b230},\cite{b290},\cite{b291},\cite{b292},\cite{b293},\cite{b294},\cite{b295},\cite{b296},\cite{b297},\cite{b298},\cite{b299},\cite{b300},\cite{b301},\cite{b302},\cite{b303},\cite{b115}. The recent breakthroughs in imaging technologies have fueled a surge of interest in bio-imaging, making it an essential tool in life sciences. Bio-imaging technologies, predominantly non-invasive, allow researchers to visualize and analyze biological processes in both in-vivo and in-vitro settings. This includes monitoring disease progression in specific organs, changes in receptor kinetics, molecular and cellular signaling, and the effects of pathogens on cells, sub-cells, and tissues. Furthermore, bio-imaging provides invaluable insights into the interactions and motility of molecules, parasites, and viruses as they traverse biological membranes. As such, bio-imaging not only drives current life science research but also aids in identifying biomarkers crucial for disease identification, progression tracking, and treatment response assessment.

In the realm of bioimage analysis, several imaging platforms have become integral to research. Among the most prominent are ImageJ \cite{b232} and CellProfiler \cite{b228}. ImageJ, a Java-based image processing tool developed by the National Institutes of Health, is renowned for its versatility and robustness in cell and biomedical imaging. Similarly, \textit{CellProfiler}, developed by Anne Carpenter and her team in 2006 and continually updated, offers a powerful suite of open-source computer vision algorithms tailored specifically for cell and bio-imaging analysis \cite{b230, b231}. These platforms empower cell and computational biology researchers by providing high-precision tools that simplify complex analyses. However, they are not without their limitations. As science rapidly advances, particularly in computer vision and machine learning, these platforms face challenges. The first limitation is their lack of integration with novel machine learning techniques, particularly deep learning architectures. The second is the absence of Spatio-temporal modules, which are crucial for intensifying the spatial changes in biological patterns across time. Fortunately, it appears that Carpenter's team is actively addressing these challenges \cite{b229}. Similar limitations are also observed in ImageJ, underscoring the need for continual advancement in bio-imaging tools.

On the other hand, High-content Screening (HCS) and High Content Analysis (HCA) platforms are increasingly becoming the go-to technologies in bio-imaging. Designed for high resolution and subcellular automated detection and phenotyping, both in in-vivo and in-vitro applications, HCS and HCA are pivotal in phenotypic drug discovery. These platforms combine automated microscopy with advanced image analysis to quantify desired features by measuring segmented or tracked cellular components in both still images and videos, whether in 2D or 3D \cite{b202, b203}. They are particularly valuable in identifying disease-associated phenotypes, making them indispensable in modern biomedical research. Segmentation, visualization, and the capture of spatial changes over time are fundamental components of HCS and HCA systems, aligning closely with the objectives of object tracking methodologies. Consequently, the development of robust and consistent imaging methods significantly enhances the performance of these systems.

Object tracking, in its essence, involves locating objects within a video, tracking them over time, and extracting their trajectories—or, in some cases, analyzing shape deformations \cite{b85, b205, b206}—for subsequent analysis. This analysis can take many forms, such as distinguishing between normal and abnormal motion behaviors or modeling the movement of specific organisms in response to various stimuli, including the presence or absence of certain drugs. This is particularly important in drug effectiveness studies, where the ability to track and quantify cellular responses is critical for understanding the underlying mechanisms and for developing new therapies \cite{b201}.

\begin{figure}
\centerline{\includegraphics[width=9.2cm, height=8.2cm]{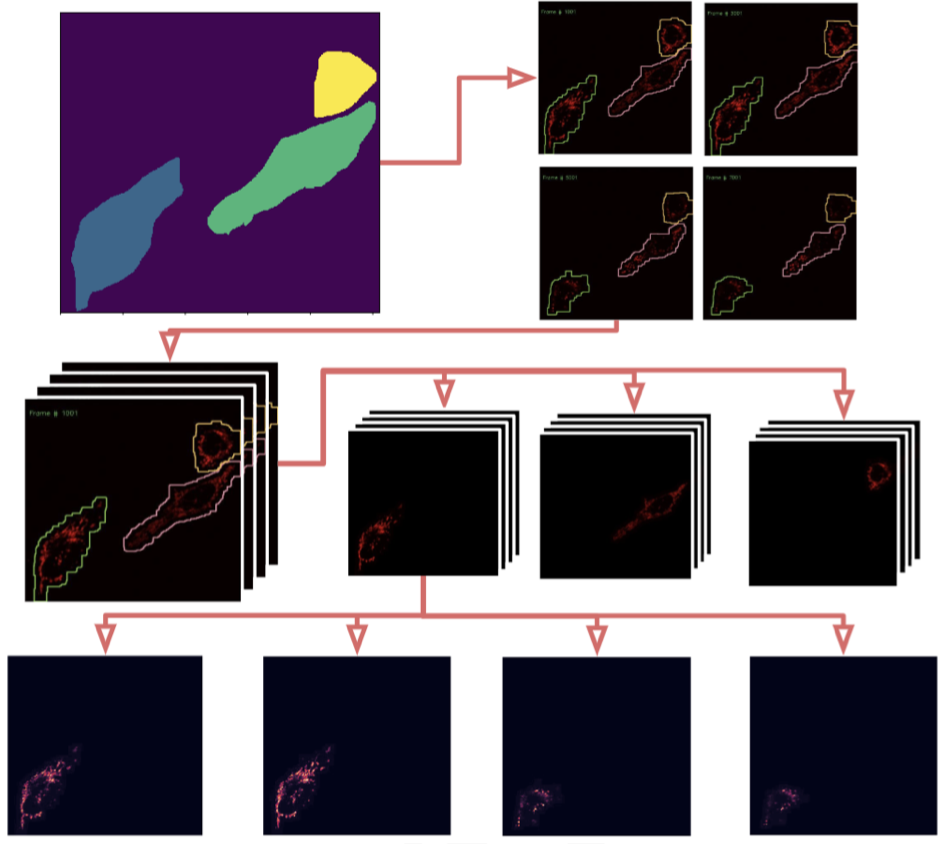}}
\caption{\textit{ Cell Segmentation algorithm for segmentation and tracking mitochondria across time. First, we segmented the cells and tracked them across time, then extracted single cells to better understand the evolution of each mitochondrion. Finally, we used Gaussian Mixture Models (GMM) to create the graph. All process details and results are discussed in \cite{b85}.}}
\label{fig:23}
\end{figure}

The application of object tracking in bio-imaging extends far beyond simple visualization. It plays a crucial role in modeling the behavior of pathogens, both in the presence and absence of various agents. For example, tracking the movement and interaction of viruses with host cells can provide critical insights into the mechanisms of infection and immune response. Similarly, tracking cell motility and shape changes over time can be instrumental in understanding the progression of diseases such as cancer, where changes in cell behavior are often indicators of disease stage and prognosis.

In disease diagnosis, object tracking allows for the monitoring of disease progression at a cellular level, providing real-time data that can be used to adjust treatment plans. In veterinary medicine, tracking techniques are used to monitor the spread of infections and to evaluate the efficacy of treatments in animal models. In drug discovery, tracking the response of cells to potential drug candidates can accelerate the identification of effective therapies, reducing the time and cost associated with traditional drug development processes.

The implications of object tracking in cell biology are profound. By enabling precise tracking of cellular processes, researchers can gain a deeper understanding of fundamental biological mechanisms. This has applications not only in basic research but also in the development of new medical treatments and in the advancement of technologies such as tissue engineering and regenerative medicine.

In conclusion, object tracking is not just a tool for visualization but a critical technology for advancing our understanding of biology at the cellular level. Its applications in disease diagnosis, drug discovery, veterinary medicine, and other biomedical fields are vast and continue to grow as new technologies and methodologies are developed. As the field of bio-imaging continues to evolve, the role of object tracking will undoubtedly become even more central to the life sciences, driving innovations that have the potential to transform healthcare and improve human and animal health globally.

\subsection{Cell Tracking}

Cells, as the fundamental units of life, are integral to a vast array of biological processes and systems. The study of cell behavior—encompassing propagation, deformation, differentiation, and migration—is essential for understanding the emergence, development, and maintenance of living organisms. These cellular activities are not just fundamental to basic biological research; they are also critical in applied sciences, particularly in the fields of drug development, disease modeling, and therapeutic innovation.

Tracking the motion of cells in response to specific drugs is especially crucial for evaluating drug efficacy and gaining insights into physiological processes that underlie health and disease. For instance, understanding how cells move and interact within their environment can reveal how a drug affects cellular functions, how it might alter disease progression, or how it could potentially be harnessed for therapeutic purposes. This knowledge is vital for developing new treatments and therapies aimed at combating a wide range of diseases, from cancer to infectious diseases.

Achieving these goals requires sophisticated imaging techniques and advanced analytical tools capable of capturing and analyzing the dynamic behavior of cells. Such analyses must often be performed in both tissue samples and laboratory environments, under conditions that may be either normal or perturbed by external factors such as drug treatment. The complexity of these studies is compounded by the need to track and quantify large numbers of cells, often over extended periods and across numerous frames in microscopy videos. This is particularly challenging in 3D video microscopy, where the dimensionality of the data adds another layer of complexity to the analysis.

Manual analysis of such extensive datasets is not feasible, especially when dealing with thousands of image frames in both 2D and 3D videos. Automated tracking systems have, therefore, become indispensable in modern cell biology research. These systems leverage advanced algorithms to track individual cells across time, enabling researchers to monitor changes in cell behavior with high precision and over extended periods. The ability to automate these processes not only saves time but also increases the accuracy and reproducibility of the data, which is crucial for the reliability of scientific conclusions.

A representative frame from video microscopy of Neutrophils is shown in Fig. \ref{fig:18}. Neutrophils, as key components of the immune system, exhibit dynamic behavior that can be indicative of various physiological states. By tracking their movement and interaction patterns, researchers can glean valuable information about immune responses, inflammation, and other critical processes in health and disease.

This section reviews the current state of cell tracking systems used for biological purposes, focusing on both 2D and 3D video microscopy. The review highlights the various methodologies employed, the challenges encountered, and the innovative solutions proposed to overcome these challenges. Through these studies, it becomes clear that cell tracking is not merely a technical exercise; it is a vital tool that provides deep insights into the fundamental workings of life and drives forward our understanding of health and disease.

\subsection{The Significance of Cell Tracking}

The tracking and quantification of biological systems, such as the parasite \textit{Toxoplasma gondii}, hold immense significance for understanding how specific therapies can inhibit the life cycles of pathogens. \textit{T. gondii} is a highly infective parasite that relies on its motility to complete its lytic cycle—a process crucial for its survival and virulence. By meticulously tracking and quantifying the motion dynamics of \textit{T. gondii}, researchers can directly measure the parasite's virility, providing critical insights into its behavior under different conditions. This understanding is fundamental for developing and evaluating therapeutic countermeasures aimed at disrupting the parasite's life cycle and mitigating its impact on infected hosts.

Beyond parasites like \textit{T. gondii}, cell tracking plays a pivotal role in studying a wide range of other organisms and cellular structures, each with unique implications for health and disease. For instance, mitochondria—often referred to as the powerhouses of the cell—serve as indicators of virulence in various pathogens, including \textit{Mycobacterium tuberculosis}. The significance of tracking mitochondria lies in their response to genetic modifications and infections. By performing genetic knockouts on strains of \textit{M. tuberculosis} and observing the resulting post-infection phenotypes of mitochondria in lung cells, researchers can assess the virulence of specific strains. This form of cell tracking provides a window into the cellular impact of infection and offers a quantitative measure of the pathogenicity of different bacterial strains.

Cell tracking is not limited to studying pathogens or organelles; it also encompasses the measurement of qualitative and quantitative statistics related to cell dynamics. These include parameters such as cell velocity, directionality, morphological changes, and interactions with other cells or the extracellular matrix. By capturing these dynamic properties, cell tracking enables the modeling of cellular motion, which is crucial for understanding processes such as cell migration, tissue formation, and wound healing. Such models are invaluable in both basic research and applied fields, including drug discovery, where they help in evaluating the effects of new compounds on cellular behavior.

In addition to the direct applications in studying pathogens and cellular structures, cell tracking is integral to a broader range of biomedical research areas. For instance, in cancer research, tracking the movement of cancer cells can reveal patterns of metastasis and help identify potential targets for intervention. In developmental biology, tracking the migration of stem cells can shed light on the processes that govern tissue regeneration and repair. In veterinary medicine, cell tracking can be used to study disease progression in animal models, providing insights that are translatable to human health.

This section reviews the diverse applications of cell tracking, highlighting its significance in various biological contexts. By understanding the movement and behavior of cells and organisms, researchers can uncover critical insights into the mechanisms of disease, the efficacy of therapeutic interventions, and the fundamental processes that sustain life.

\subsubsection{Cell Motion Modeling}

Cell motion modeling is a critical application of cell tracking, providing insights into how cells move and interact within their environments. Before delving into cell motion modeling, it's important to discuss spatial cell modeling, which serves as the foundation for spatiotemporal cell motion modeling. Spatial cell modeling involves the accurate representation of cell shapes and structures across different spatial dimensions, enabling better tracking results and, consequently, more precise motion modeling.

Generative models that can represent cells have been explored for some time, as they are particularly significant in cell biology. These models combine spatial information from different images, captured at various angles, to create a comprehensive understanding of cell shapes. This holistic view is not possible with a single image alone. For example, previous works have proposed generative models to accurately model cells \cite{b164, b165}. With the advent of new generations of Generative Adversarial Networks (GANs), researchers have developed advanced models for spatial cell modeling and appearance.

In recent years, a supervised GAN-based method was introduced to generate high-quality synthetic electron microscope images of mitochondria that are almost indistinguishable from real ones \cite{b166}. In this approach, Gaussian noise is used as input, which is then processed by a label generator to produce label images. These images are further translated into Electron Microscope (EM) images through an image generator layer, which could be either a U-Net or a Conditional Random Network (CRN). The model's performance is optimized using a specialized loss function tailored for GANs. Such methods represent a significant advancement in cell modeling, allowing for the generation of realistic cell images that can be used to study cellular behavior under various conditions.

Another study explored different methods for cell modeling using a diverse collection of 2D and 3D cell images \cite{b141}. Although this research did not specifically focus on tracking, the evaluation of traditional methods and deep autoencoders provides valuable insights for researchers interested in modeling cell dynamics. These findings can inspire new approaches to modeling cell motion, particularly in understanding how cells move and change over time.

Once cells are accurately tracked, their motion can be modeled and analyzed. For instance, a study combined a statistical method based on Autoregressive (AR) models with a GAN to predict cell motion and appearance in human neutrophils observed in video microscopy data \cite{b122}. The GAN was used to capture the spatial components of the cells, while the AR model accounted for the temporal dynamics. This combination of techniques allows for a more comprehensive understanding of how cells move and behave under different conditions.

\begin{figure*}[h!]
\centering
\includegraphics[width=.8\textwidth]{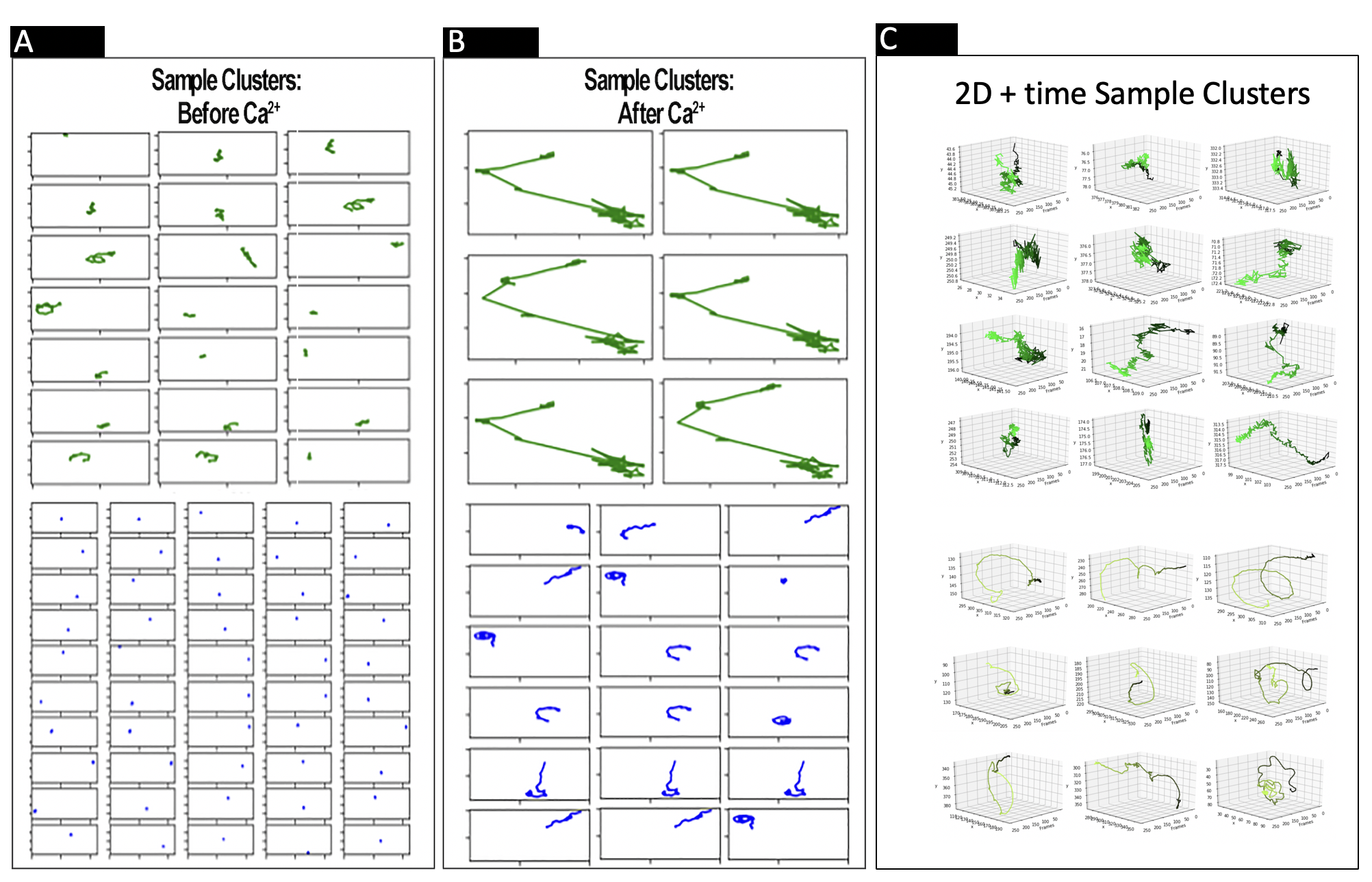}
\caption{\label{fig:31} \textit{\textbf{A}: Sample clusters of the \textit{T. gondii} motion phenotype before the addition of $Ca^{2+}$. The upper clusters reveal circular motility patterns, while the lower cluster indicates mostly immotile cells. \textbf{B}: Sample clusters after stimulation with $Ca^{2+}$. The top cluster exhibits a noisy pattern, whereas the bottom cluster demonstrates helical motility patterns. \textbf{C}: Two sample clusters after calcium addition, represented in 3D plots (2 spatial dimensions + 1 temporal dimension).}}
\end{figure*}

In our previous work, we employed an autoregressive model to discover motion phenotypes of \textit{Toxoplasma gondii} after tracking cells in a 2D space \cite{b42}. We created a corpus of trajectories, divided into two parts: before and after calcium stimulation. Calcium inhibits the motion of \textit{T. gondii}, making this distinction crucial. We applied spectral clustering to group the preprocessed trajectories, clustering the motions of \textit{T. gondii} cells before and after calcium stimulation. Figure \ref{fig:31} illustrates the clustering results of the 2D videos of \textit{T. gondii}, while Figure \ref{fig:31}-C depicts the $2D + time$ trajectories of two sample clusters.

\begin{figure*}[h!]
\centering
\includegraphics[width=.935\textwidth]{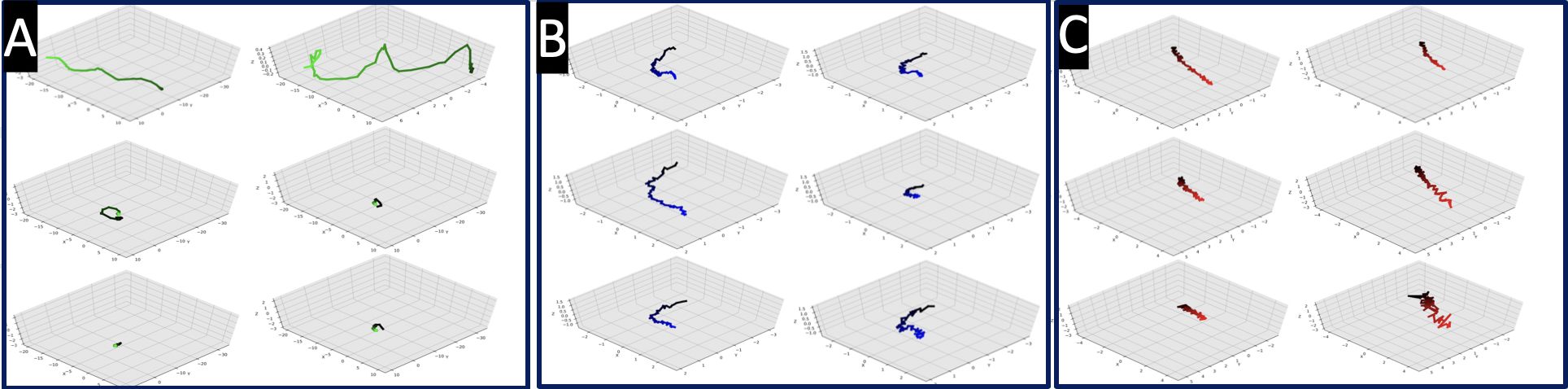}
\caption{\label{fig:33-1} \textit{Clustering results of 3D + time \textit{T. gondii} trajectories. Three specific clusters, A, B, and C, are visualized with sample trajectories in 4D subplots. The trajectories are color-coded by cluster, with each subplot illustrating 3 spatial dimensions, and the temporal dimension represented by hue. The lightest point corresponds to the cell's position in frame '0', while the darkest point indicates the position in the last frame.}}
\end{figure*}

\begin{figure*}
\centering
\includegraphics[width=0.95\textwidth]{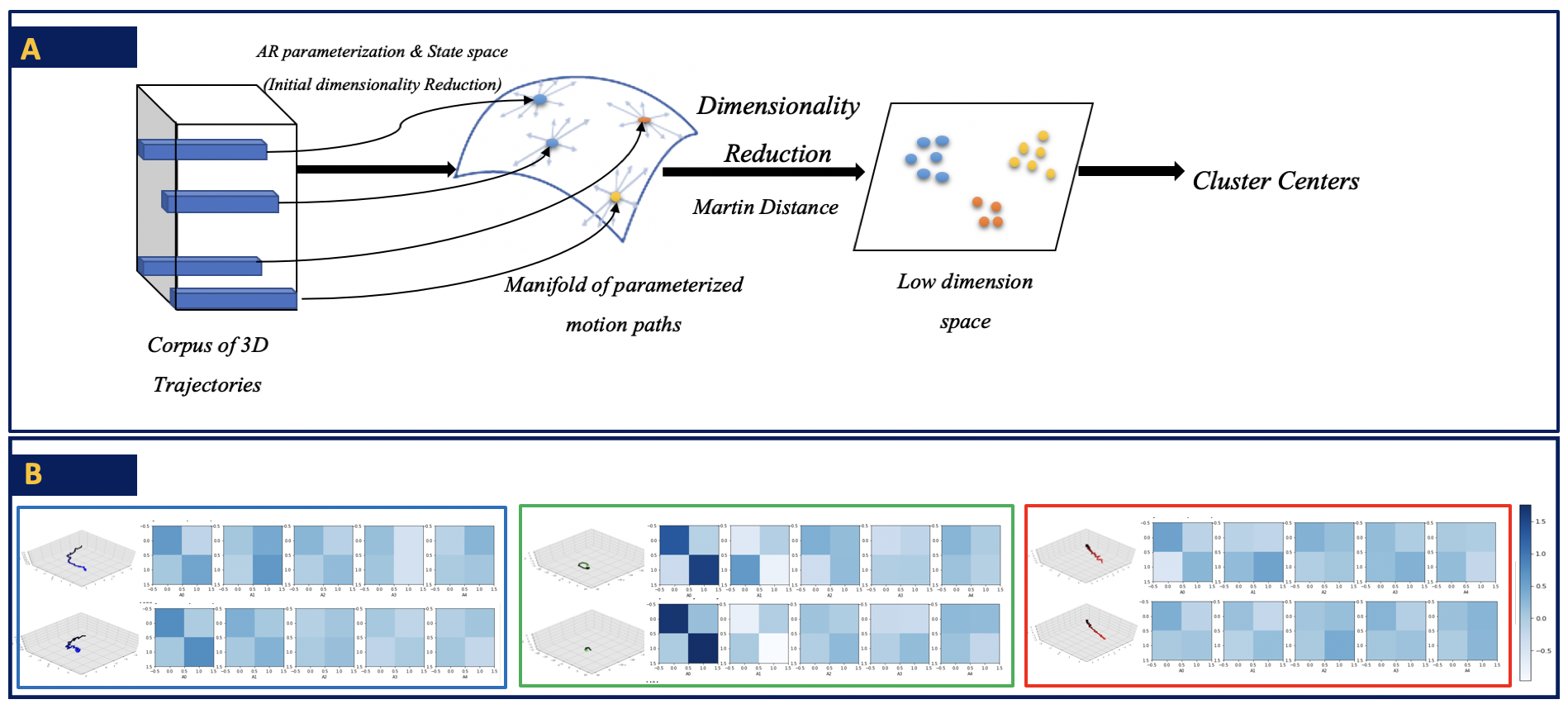}
\caption{\textit{\textbf{A}: The model used for clustering 3D trajectories, including AR parameterization, dimensionality reduction using state space, creation of a manifold of parameterized motion paths, computation of distance between points in the manifold using a geodesic metric (Martin distance), and finally, clustering of 3D trajectories using Spectral Clustering. \textbf{B}: Visualization of sample AR matrices from each trajectory cluster. Qualitative examination reveals that AR matrices within each category are similar, while distinctly different from those in other clusters.}}
\label{fig:34}
\end{figure*}

We also extended our research to study the 3D motion phenotypes of \textit{T. gondii}. While the initial assumption was that \textit{T. gondii} motility in 3D follows a corkscrew-shaped trajectory, our results revealed at least three distinct motion types. To achieve this, we developed a distributed, scalable pipeline using \textit{Dask} and ran our tracking module on Google Cloud Platform (GCP) as explained in \cite{b114}. We created a corpus of around 3,000 trajectories, each capturing the 3D spatial positions of cells over 61-63 frames. The pipeline, illustrated in Figure \ref{fig:34}, involved parameterizing trajectories using an Autoregressive model, reducing spatial dimensions to two, and creating a manifold of parameterized trajectories. The distance between AR systems was calculated using Martin distance, which was then converted to Martin affinities using an RBF kernel. Finally, spectral clustering was applied to identify clusters of 3D cell motions, with sample clusters shown in Figure \ref{fig:33-1}.

In addition to traditional methods, deep learning techniques are increasingly being used for trajectory clustering. Several deep learning-based methods for clustering trajectories are discussed in \cite{b167, b168, b169}, offering new avenues for modeling cell motion.

P. J. Harris proposed a mathematical model to capture cell motion in a fluid environment, where chemicals stimulate cell movement \cite{b135}. This model considers various factors such as intercellular communication, force, gravity, and more. The proposed Stokes-flow model effectively captures the clustering of cells and their movement over time, providing a detailed understanding of cellular behavior in fluid environments.

\begin{figure}
\centerline{\includegraphics[width=9cm, height=12cm]{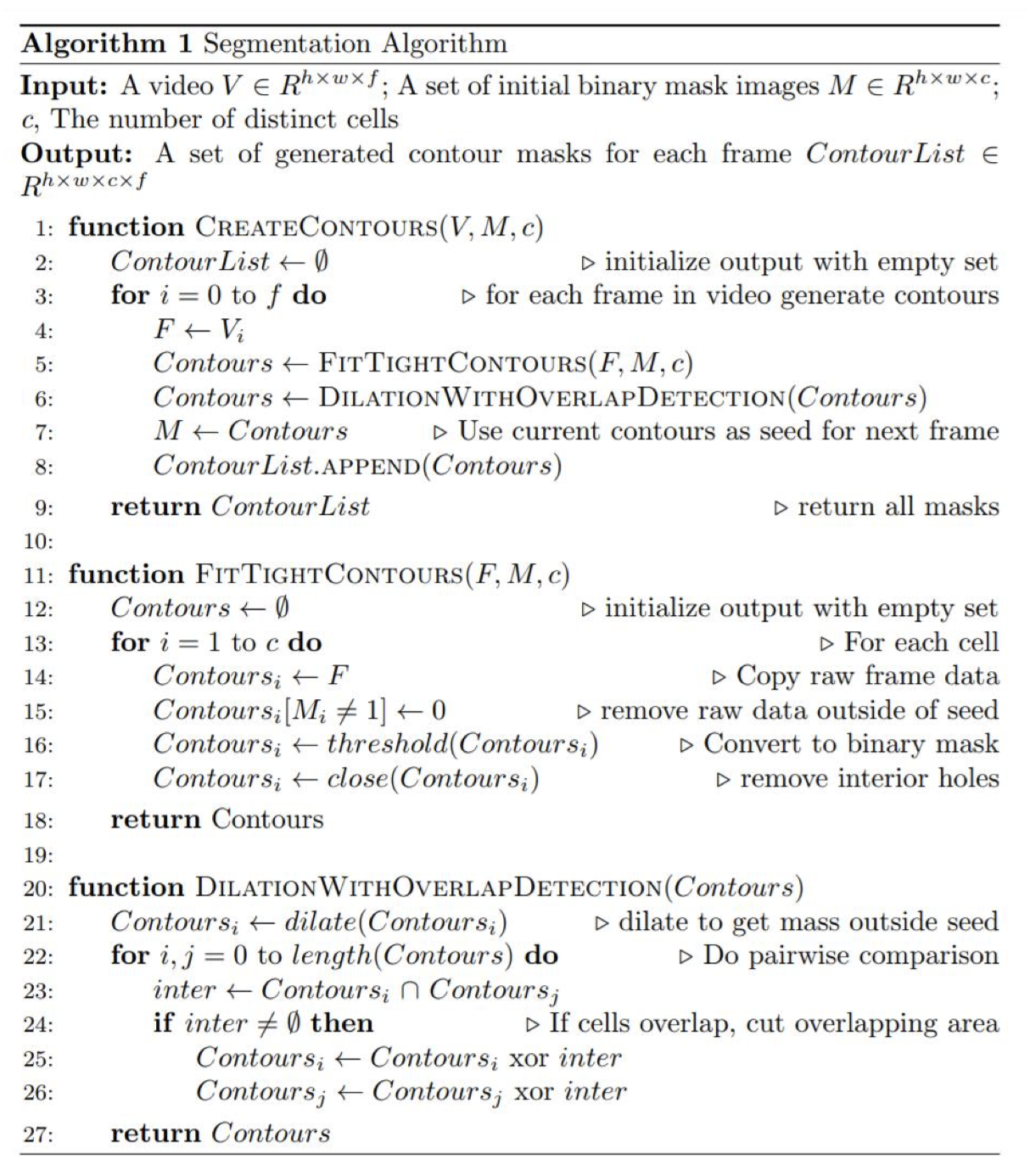}}
\label{fig:Algo2}
\end{figure}
\begin{figure*}[h!]
\centering
\includegraphics[width=0.8\textwidth]{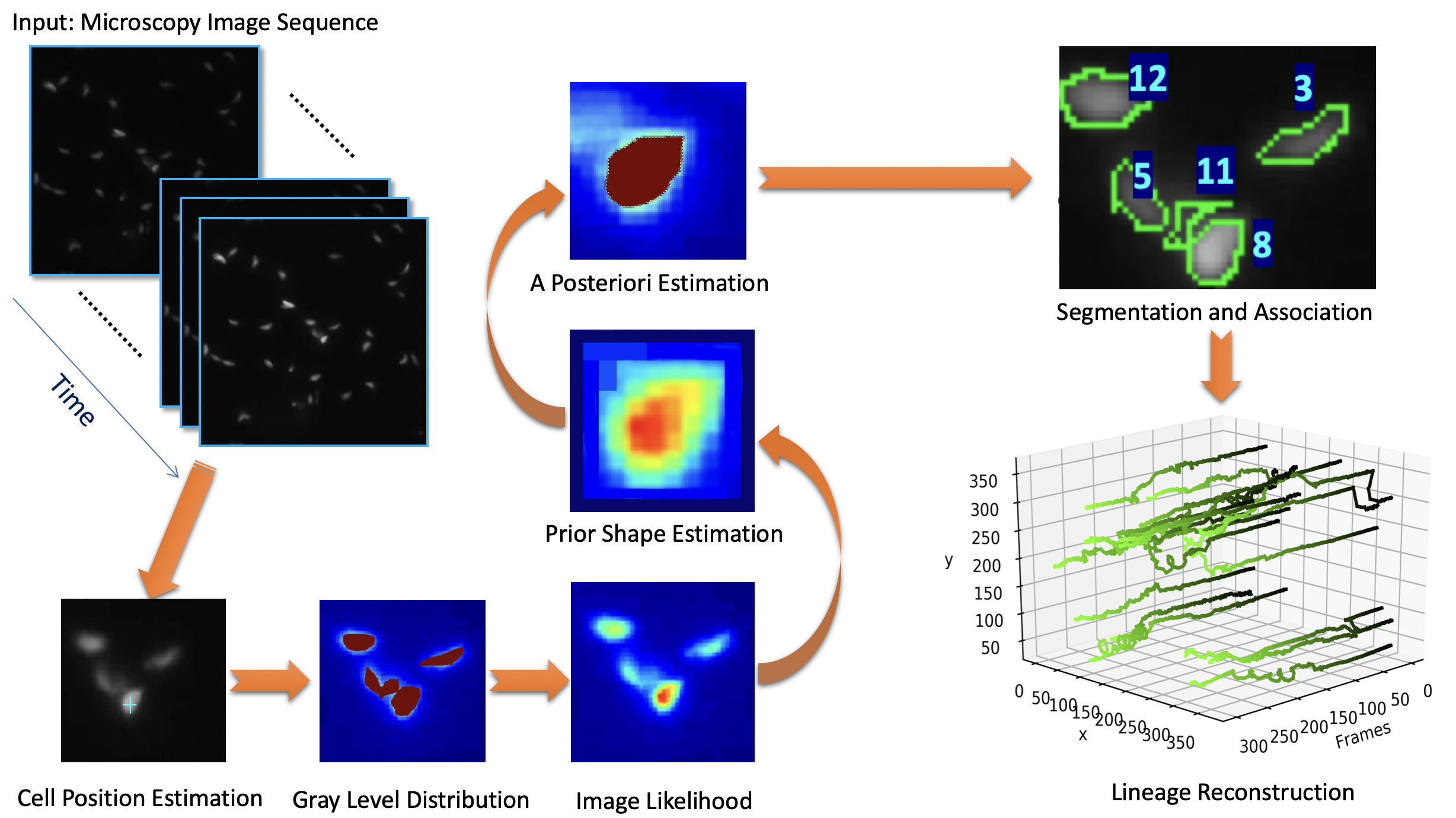}
\caption{\label{fig:24} \textit{A) The pipeline proposed for the probabilistic cell tracking \cite{b81} applied on our \textit{T. gondii} video. }}
\end{figure*}

These examples illustrate the broad applications of cell motion modeling in various contexts, from understanding basic cellular behavior to developing advanced models for studying complex biological processes.

\subsection{2D Cell/Sub-Cell Tracking: Challenges, Applications, Methods}

Cell and sub-cell tracking in 2D microscopy videos is a cornerstone of modern biomedical research, providing critical insights into cellular behaviors and interactions. However, the task is fraught with challenges that stem from both technical limitations and the inherent complexity of biological systems. The difficulty of automatic segmentation and tracking is directly linked to the quality of recorded video over time. Several parameters such as Signal to Noise Ratio (SNR) and Contrast Ratio (CR) are identified by Ulman et al. \cite{b84} as key factors that influence these challenges. Yet, the problem is multifaceted, encompassing additional issues like non-uniform background illumination and hue variations. For example, Figure \ref{fig:18} illustrates a video of Neutrophils with non-uniform illumination, which complicates the accurate study of cell behavior and motion. 

To address these issues, multiple image processing techniques can be employed. In our ongoing research on Neutrophils, we utilized histogram equalization to correct illumination disparities, followed by image filtering techniques such as gamma enhancement and logarithmic correction to improve the visibility of cell contours. Additionally, we applied a specialized blob detection method, enhanced by contour features and bilateral filtering, to mitigate the interference of non-relevant objects. The results of these preprocessing steps are shown in Figure \ref{fig:19}.

As depicted in the caption of Figure \ref{fig:19}, our step-by-step filtering and segmentation approach successfully resolved the initial challenges, enabling us to track the cells using contour-based detection and a matching algorithm that links the bounding boxes of each cell to its corresponding location in consecutive frames. The final tracking results are shown in Figure \ref{fig:20}. A similar strategy was employed in our work on tracking \textit{Toxoplasma gondii} cells across videos \cite{b5}, where we processed 2D microscopy videos with appropriate preprocessing techniques before tracking the cells. We then measured quantitative statistics and analyzed the motion dynamics of \textit{T. gondii} cells in response to calcium stimulation, as shown in Figure \ref{fig:20}. The 4D plots of the extracted trajectories from the KLT tracker, used in our previous studies \cite{b5, b42}, illustrate the spatial ($x, y$) and temporal ($t$) dimensions, with color indicating the extracellular calcium levels over time.

Another significant challenge in cell tracking is the rate of image capture during data acquisition. Missing frames can introduce inconsistencies in tracking results, particularly when dealing with fast-moving cells or large populations of cells. For instance, tracking cells such as Neutrophils and \textit{T. gondii} is inherently more challenging than tracking sub-cellular structures like mitochondria, which move at slower speeds. Additionally, the semi-rigid shapes of cells like \textit{T. gondii} introduce further complications due to shape flexibility, which can hinder accurate tracking. Occlusion, where similar shapes and textures overlap, presents another obstacle in maintaining tracking continuity, as demonstrated in Figure \ref{fig:21}.

In \cite{b42}, we employed the Kanade-Lucas-Tomasi (KLT) tracker for cell tracking, which performed better than traditional contour- or color-based methods. However, KLT has its limitations, particularly when dealing with occlusion. The algorithm assumes a fixed number of objects across frames, leading to the renumbering effect when objects are temporarily lost and their features deleted. To mitigate this issue, we implemented a post-processing correction step, as detailed in Algorithm 1.A. A sample frame from the tracked video, along with the final extracted trajectories, is shown in Figure \ref{fig:22}.

\begin{figure*}
\centerline{\includegraphics[width=0.8\textwidth]{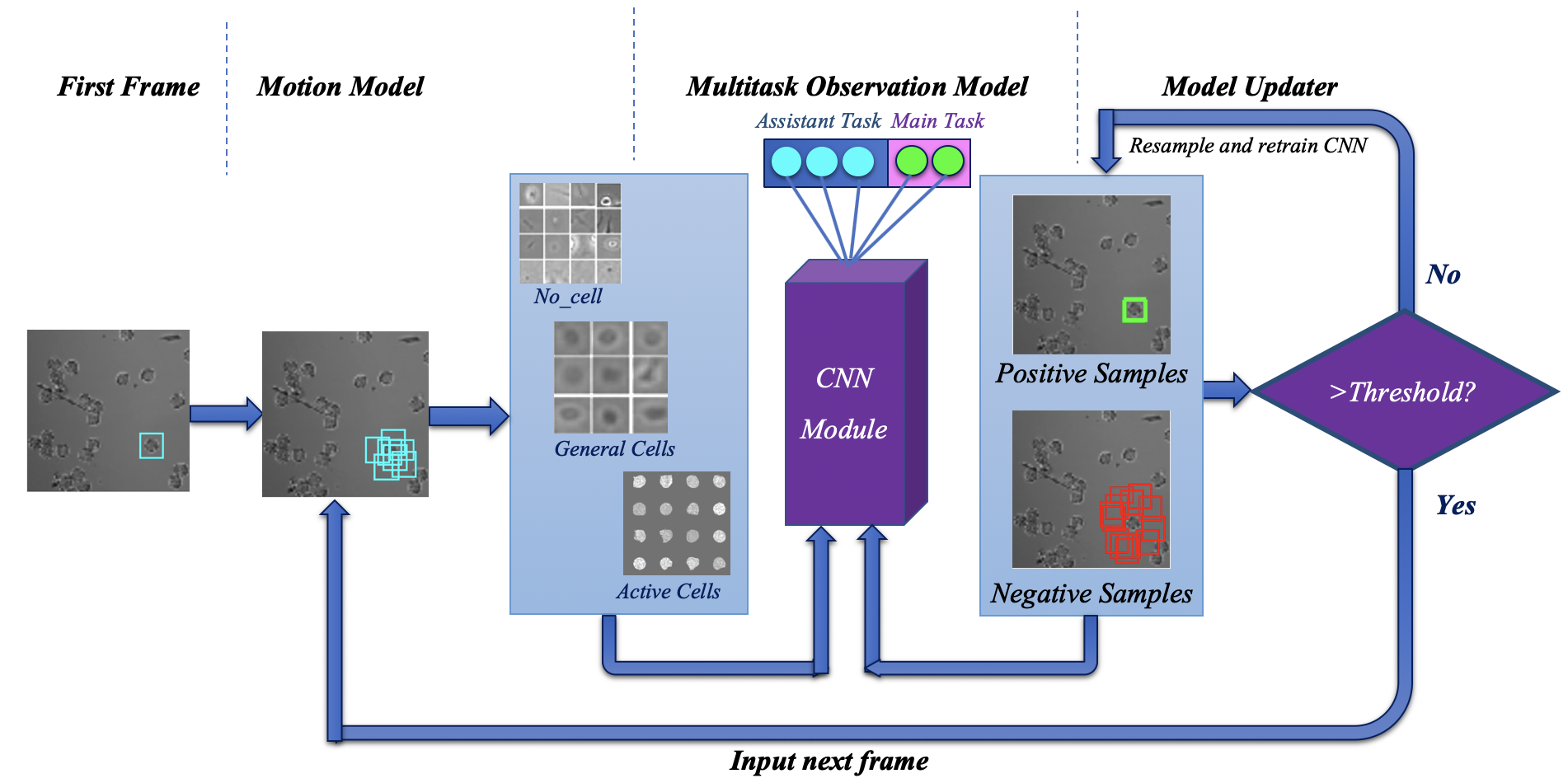}}
\caption{\textit{Proposed pipeline in \cite{b107} for cell tracking. The model consists of 3 major modules, including motion model, multitask observation model, and model updater modules.}}
\label{fig:26}
\end{figure*}

Sub-cellular evolution, particularly in mitochondria within lung cells, represents another frontier in cell tracking research. In \cite{b85}, we conducted preliminary work on spatiotemporal modeling of diffuse sub-cellular morphologies using mitochondrial protein patterns from cervical epithelial cells. By leveraging graph theory, we modeled diffuse mitochondrial patterns as social networks, gaining insights into the stress types imposed on mitochondria by external stimuli. A mixture of Gaussians was used to represent mitochondrial patterns in a parametric form, which were then converted to graph Laplacians to define a network. Changes in the topology of these Laplacians provided biological interpretations of the evolving morphology. The segmentation algorithm used is further detailed in \cite{b86} and illustrated in Algorithm 2, with the cell segmentation process shown in Figure \ref{fig:23}. Additionally, in \cite{b205}, we developed an automated system for the spectral analysis of the created graph.

The importance of accurate cell tracking extends beyond individual cell analysis to applications like wound healing assays. For example, in \cite{b83}, a method was presented for tracking cells in phase-contrast microscopy images using conventional manual tracking techniques. The authors segmented the cells and applied image restoration techniques before utilizing blob detection and matching algorithms for tracking. A similar approach was employed for tracking mitosis over time \cite{b84}. In our work \cite{b85}, the KLT-based tracking system was also used for tracking \textit{T. gondii} in 2D microscopy videos.

\begin{figure}
\centerline{\includegraphics[width=9cm, height=4.5cm]{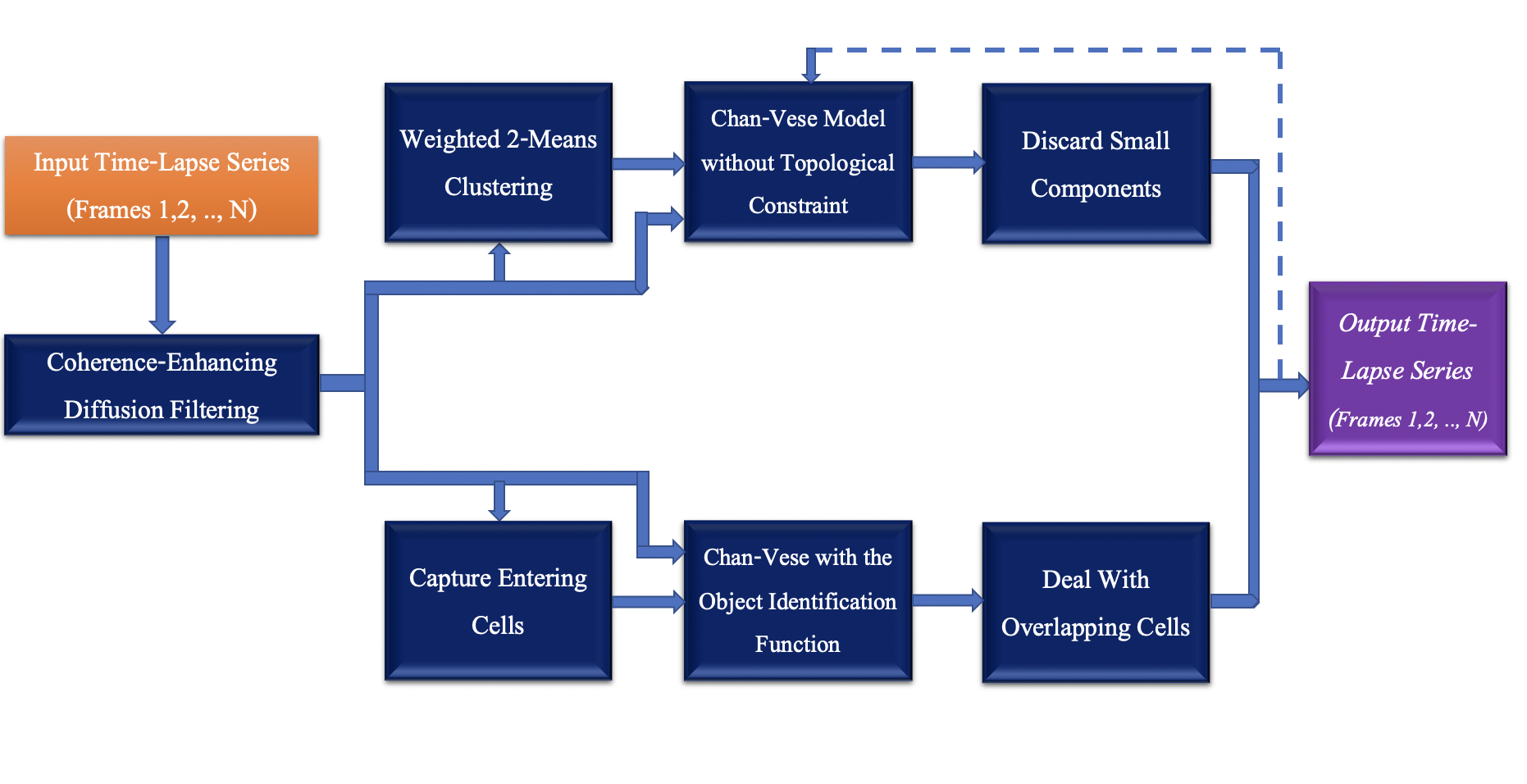}}
\caption{\textit{The proposed tracking algorithm in \cite{b90}}}
\label{fig:27-1}
\end{figure}

A more recent study proposed an unsupervised tracking and segmentation framework for microscopy videos \cite{b81}, addressing the dual challenges of segmentation and tracking through Bayesian inference of a dynamic model. The framework does not assume any specific cell shape, instead dividing the process into segmentation and tracking stages. Probabilistic pixel assignment is used for segmentation, while Kalman filters are employed for motion estimation of the segmented objects. A fast-matching algorithm is then used to probabilistically align cells across frames. The methodology is depicted in Figure \ref{fig:24}. Cells are assigned labels throughout the sequence, as represented in Equation (10):

\begin{equation}\label{8}
C = \{ C^{(1)}, ..., C^{(k)}\}
\end{equation}

Here, $k$ represents the number of cells across a video. The cell segments and state vectors are formulated as follows:

\begin{equation}\label{9}
\Gamma_t  = \{\Gamma_t^{(k)} \}^{K_t}_{k=0}
\end{equation}

\begin{equation}\label{10}
\xi_t  = \{\xi_t^{(k)} \}^{K_t}_{k=0}
\end{equation}

Where $\Gamma_t$ denotes the segmentation of the $k^{th}$ cell in frame $t$, and $\xi_t$ is the state vector, which can include the location, velocity, or cell shape uncertainty. The state vector is defined as:

\begin{equation}\label{11}
\xi_t^{(k)}  =\Bigg[ c_{x_t}^{(k)},  c_{y_t}^{(k)},  v_{x_t}^{(k)},  v_{y_t}^{(k)},  \epsilon_{t}^{(k)} \Bigg]^{T}=\Bigg[ C_{t}^{(k)T}, V_{t}^{(k)T}, \epsilon_t^{(k)} \Bigg]^{T}
\end{equation}

where $ C_{t}^{(k)T} = \big[c_{x_t}^{(k)}, c_{y_t}^{(k)}\big]^T$ represents the center of mass of the cell at time $t$, and $ V_{t}^{(k)T} = \big[v_{x_t}^{(k)}, v_{y_t}^{(k)}\big]^T$ indicates the velocity. A probabilistic method is used to assign specific pixels to specific cells, with the probability calculated as:

\begin{equation}\label{12}
\Theta_{t|t}^{(k)}  = P\bigg( X \in \Gamma_t^{(k)}| I_t, \Gamma_0, \xi_0, ... ,\Gamma_{t-1}, \xi_{t-1}\bigg)
\end{equation}

\begin{equation}\label{13}
   = P\bigg( X \in \Gamma_t^{(k)}| I_t, \Gamma_{t-1}, \xi_{t-1}\bigg)
\end{equation}

Using Bayes' theorem, the probability $P$ can be computed as:

\begin{equation}\label{14}
\begin{split}
 P\bigg( X \in \Gamma_t^{(k)}| I_t,
\Gamma_{t-1},
\xi_{{t}{(k-1)}}\bigg)=
\\
\frac{{P\big(I_t|X\in \Gamma_t^{(k)}, \Gamma_{t-1}, \xi_{t-1}\big)}{P\big( X \in \Gamma_t^{(k)}|\Gamma_{t-1},\xi_{t-1}\big)}}
{P\big( I_t| \Gamma_{t-1}, \xi_{t-1}\big)}
\end{split}
\end{equation}

The prior probability is computed as:

\begin{equation}\label{15}
\Phi_{t|t-1}^{(k)}  = P\bigg( X \in \Gamma_t^{(k)}| \Gamma_{t-1}, \xi_{t-1}\bigg)
\end{equation}

Segmentation is then defined by maximizing a posterior estimator:

\begin{equation}\label{16}
\begin{split}
\mathcal{L}_t(x) = \arg\max_{k \in K_t} \Big( \Theta_{t|t}^{(k)} (x) \Big)\\
= \arg\max_{k \in K_t}  \Big(\Phi_{t|t-1}^{(k)} (x) L_{t|t} (x)\Big)
\end{split}
\end{equation}

After segmentation, Kalman filters are employed for tracking. For further details, refer to \cite{b81}.

In 2004, C. Wahlby et al. \cite{b87} described a region-based segmentation mechanism (combining multiple morphological filters and watershed segmentation) for segmenting and tracking cell nuclei in tissue sections. Another study \cite{b88} applied automatic peak finding routines and center interpolation, filtering out smaller particles as noise. The authors used distance minimization between consecutive frames and motion correction to avoid local distortions. Active contours were used for multiple cell segmentation and tracking in \cite{b89}, where the computational cost was optimized, and accuracy was enhanced by minimizing a modified non-PDE-based energy function. In a similar work, \cite{b90} proposed a tracking algorithm for fluorescence cells based on the Chan-Vese Model, an approximation used in the functional formulation of image segmentation introduced by Mumford et al. \cite{b91}. The tracking algorithm flow chart is illustrated in Figure \ref{fig:25}.

Another method \cite{b92} created cell trajectories using the Viterbi algorithm, employing a greedy approach to maximize a probabilistically motivated scoring function. In \cite{b93}, a particle-cell detection and tracking algorithm, PCRM, was proposed, consisting of object identification, tracking, measure calculation, and relation mining phases. The ripple spreading optimization method was used in \cite{b95} for multiple cell tracking, inspired by the phenomena of ripples on a liquid surface. This method proposed a tracking approach resistant to occlusion. GAN-based methods with specified loss functions were also introduced in \cite{b163} to improve cell visualization, offering new directions for detection-based tracking models. Other innovative models, similar to those previously discussed, are presented in \cite{b96}--\cite{b106}.

\begin{figure}
\centerline{\includegraphics[width=9cm, height=3cm]{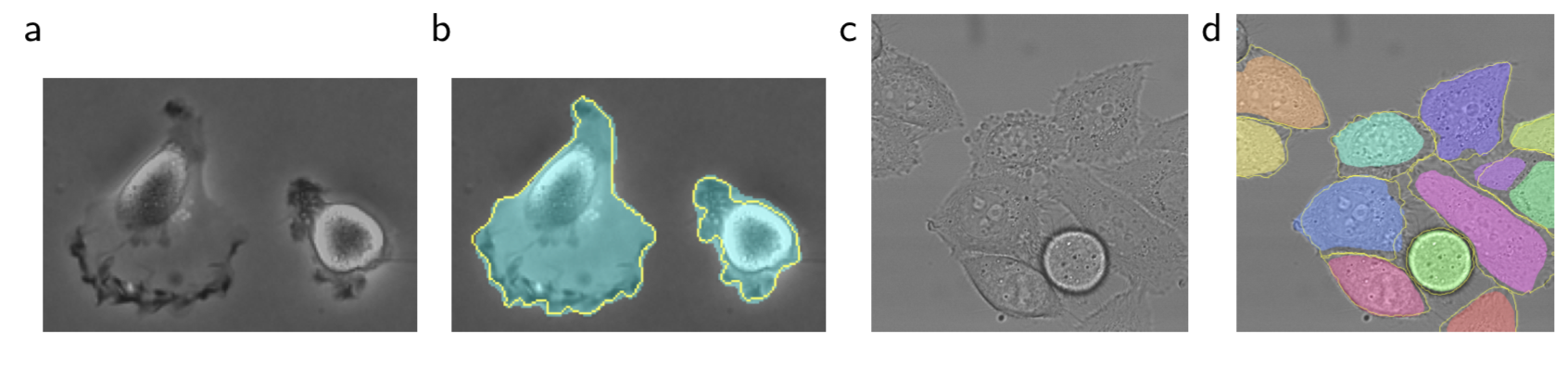}}
\caption{\textit{Results of proposed U-Net cell segmentation \cite{b112} on ISBI cell detection challenge (\textcopyright IEEE2015)}}
\label{fig:28-2}
\end{figure}

\begin{figure}
\centerline{\includegraphics[width=9cm, height=5.5cm]{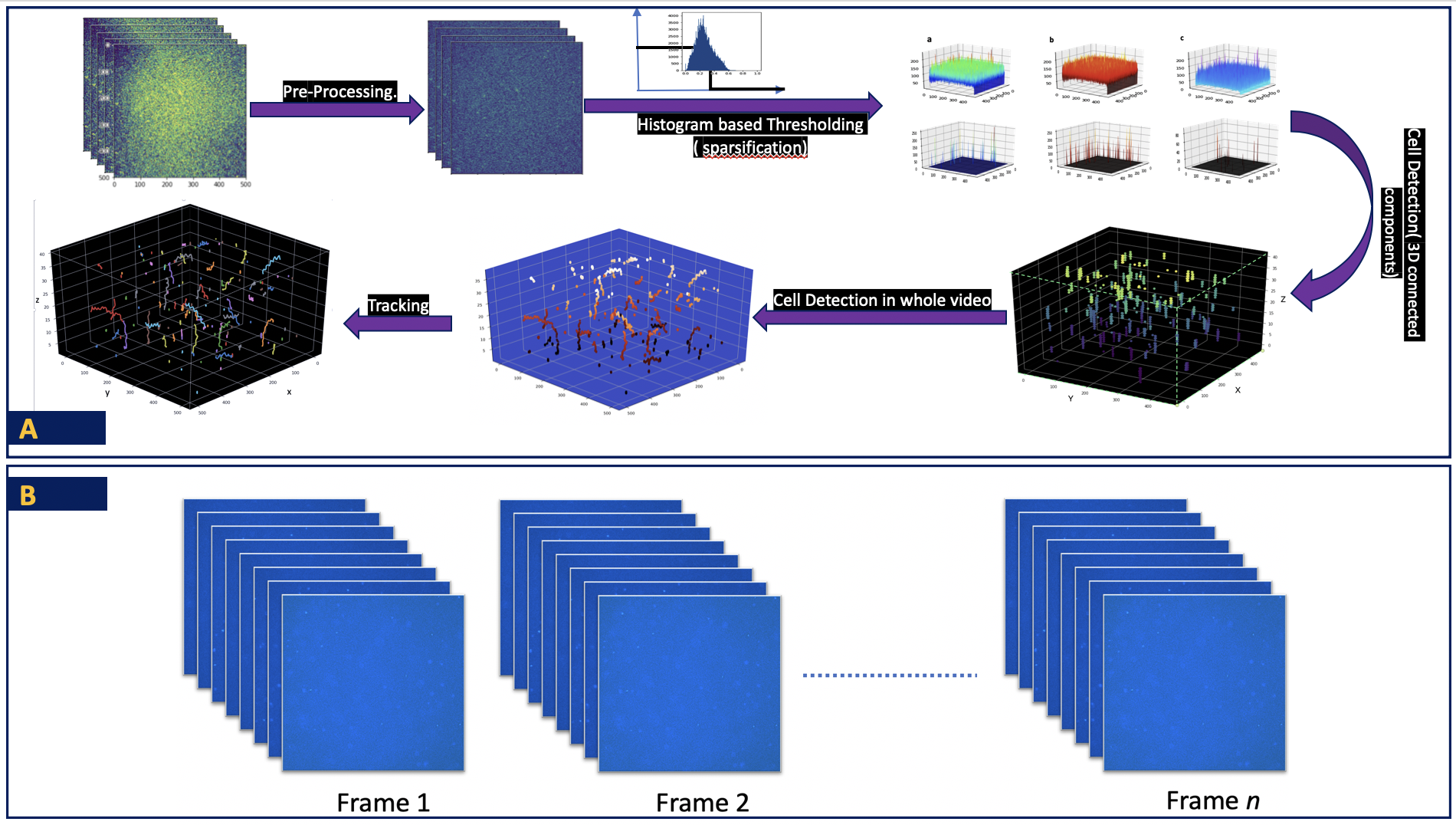}}
\caption{\textit{\textbf{A}: Proposed pipeline for tracking 3D video microscopy of \textit{T. gondii} \textbf{B}: slices formation in a 3D video microscopy of \textit{T. gondii}}}
\label{fig:29}
\end{figure}

\begin{figure*}[h!]
\centering
\includegraphics[width=1.0\textwidth, , height=6cm]{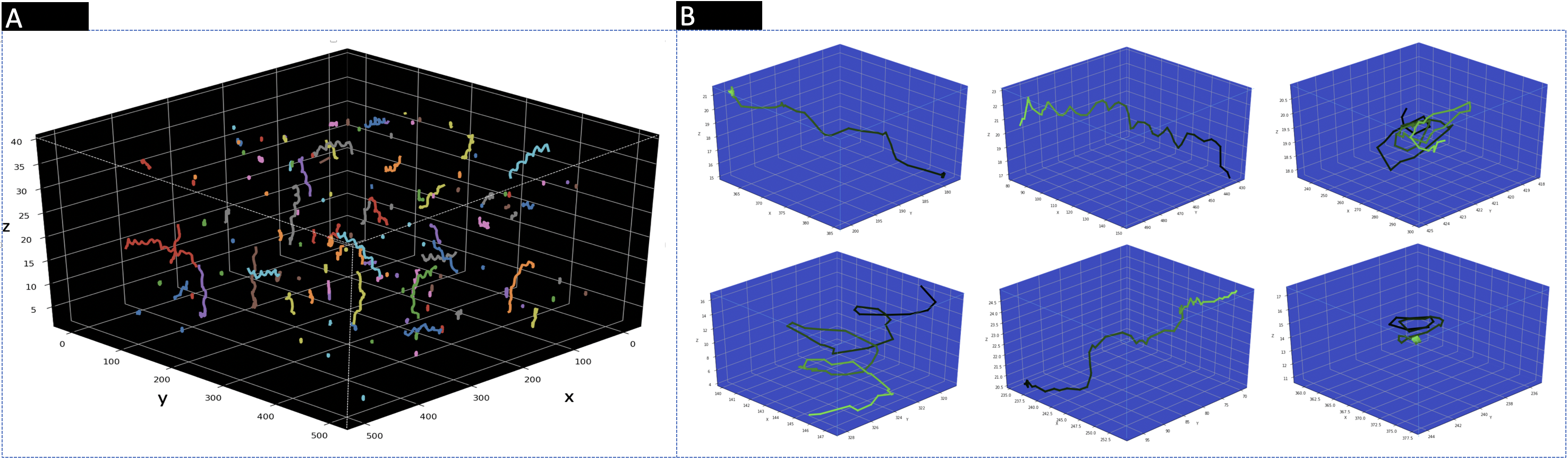}
\caption{\label{fig:30} \textit{:  A) 3D cell trajectories of a sample 3D video of \textit{T. gondii}. B) 4D illustration of some sample tracked cells. X, Y, and Z demonstrate the spatial locations, and the trajectory colors indicate the temporal dimension. The darkest point indicates the first frame, and the lightest point denotes the last frame.\cite{b114}}}
\end{figure*}

\begin{figure*}[h!]
\centering
\includegraphics[width=0.8\textwidth]{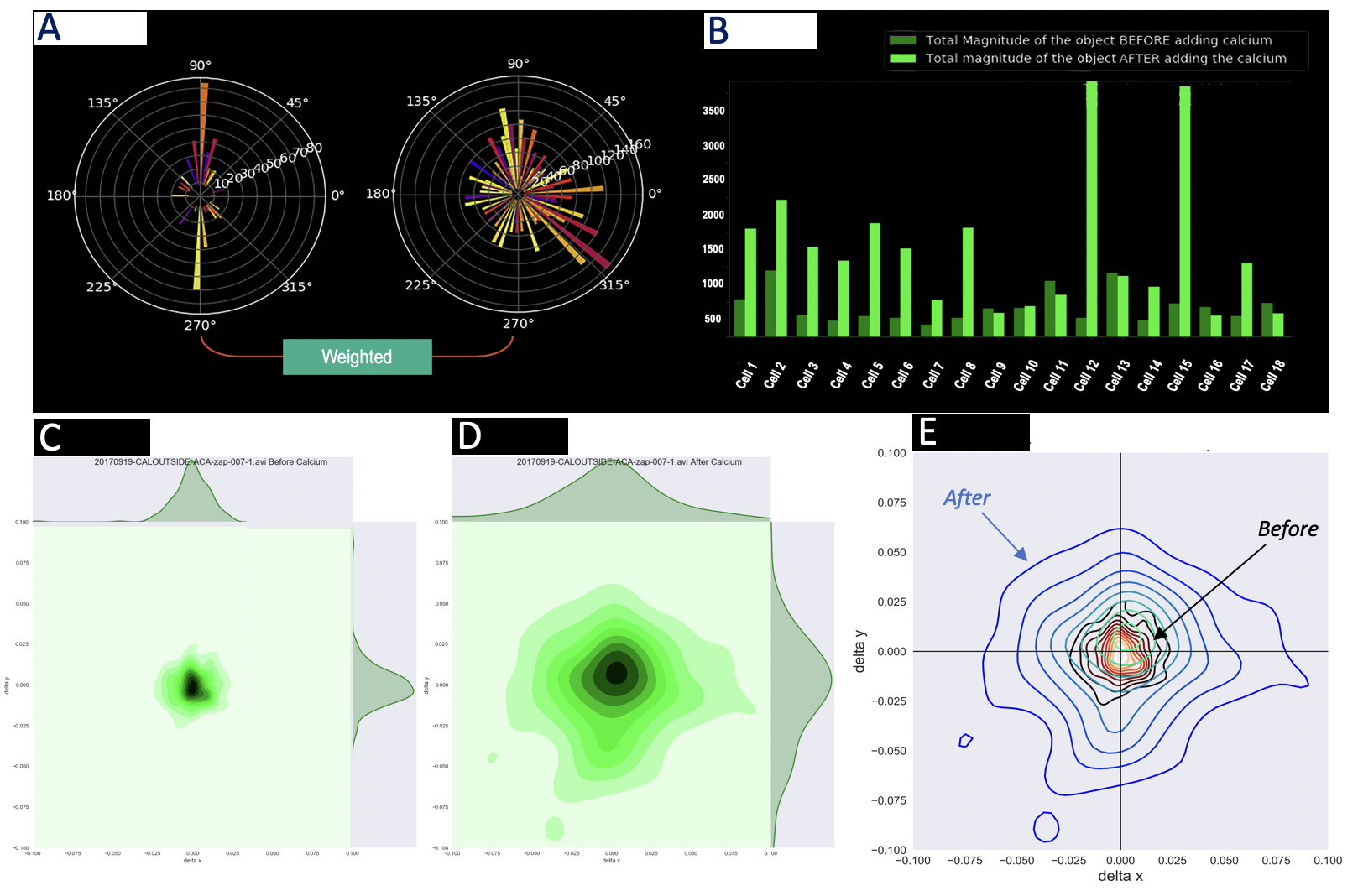}
\caption{\label{fig:33} \textit{Sample statistics extracted from cell tracking and motion capturing of \textit{T. gondii} in 2D microscopy videos. \textbf{A}: Histogram of angles showing the directional movement of cells across the video. \textbf{B}: Total magnitude of cell motions before and after stimulation with extracellular $Ca^{2+}$ in a sample video with 18 cells. \textbf{C}: 2D histogram of movement before calcium addition. \textbf{D}: 2D histogram of movement after calcium addition. \textbf{E}: Contour plot combining both 2D histograms from \textbf{C} and \textbf{D}.}}
\end{figure*}

The aforementioned studies predominantly rely on conventional, feature-based, and statistical methods. Compared to general object tracking applications, there has been relatively limited application of deep convolutional methods in cell tracking. In a recent study \cite{b94}, a multi-feature fusion re-tracking algorithm based on the tracking-by-detection method was proposed for automated multiple cell tracking. The method begins with a region proposal approach based on faster R-CNN to generate cell candidate proposals. These proposals are then used in a cell tracking method that fuses the bounding box and feature vector of cell candidates based on previous results. Finally, a re-tracking algorithm integrates historical matching frame information, as depicted in Figure \ref{fig:26}.

Another deep learning-based method for cell tracking was proposed by Tao He et al. \cite{b107}, utilizing convolutional neural networks (CNNs) and multi-task learning (MTL) techniques. The CNNs learn cell features, while MTL improves tracking performance. The method involves a particle filter motion model, a multi-task learning observation model, and an optimized model update strategy. The training process is divided into an online tracking phase and a classification phase using the MTL technique. The observation model is trained by a CNN to learn robust cell features, with the overall model depicted in Figure \ref{fig:27-1}. However, the method's main limitation is its design for single-cell tracking, limiting its application in multiple-cell scenarios.

Further advancements in cell tracking include the application of Recurrent Neural Networks (RNNs), Autoencoders, and other deep learning methods for cell motility discrimination and prediction \cite{b109}. A method for data association in particle tracking is explained in \cite{b110}. In another study, \cite{b111} utilized a vanilla U-Net model for cell counting, detection, and morphometry. This work, also presented as a plugin for ImageJ, allows researchers without deep learning expertise to benefit from U-Net in biological image discovery. The U-Net model's application to various cell types yielded promising results, as shown in Figure \ref{fig:28-2}. Arabelle and colleagues \cite{b134} combined U-Net with Long Short-Term Memory (LSTM) networks, harnessing LSTM's ability to capture temporal dimensions, enabling accurate segmentation of cells in scenarios where temporal cues are crucial.

These advancements highlight the ongoing evolution of 2D cell tracking techniques, as researchers continue to push the boundaries of what is possible in this critical area of biomedical research.

\begin{table*}[htbp]
\caption{Comprehensive and Systematic Classification of Vision Models and Referenced Papers Across Categories}
\label{comprehensive-table}
\centering
\scriptsize 
\begin{tabularx}{\textwidth}{@{}p{3cm}*{5}{>{\centering\arraybackslash}X}@{}}
\toprule
\textbf{Category} & \textbf{Extensive} & \textbf{Robust} & \textbf{Trainable} & \textbf{Multi-Domain} & \textbf{Scalable} \\ 
\midrule

\textbf{Conventional and Classic Methods} 
& \cite{b3, b4, b6, b8, b10, b19, b21, b22, b51, b52, b53, b123, b174, b216, b217, b219, b220, b244, b245, b270, b277} 
& \cite{b3, b4, b6, b8, b10, b19, b21, b22, b51, b53, b54, b57, b123, b174, b216, b217, b219, b220, b244, b245} 
& 
& \cite{b8, b10, b15, b19, b218, b219, b220, b244} 
& \cite{b5, b54, b174} \\ 
\midrule

\textbf{Feature-Based Tracking Models} 
& \cite{b19, b20, b23, b24, b60, b213, b214, b223, b243, b248, b255, b256, b257, b259, b260, b274, b275} 
& \cite{b19, b20, b23, b24, b50, b57, b60, b213, b214, b223, b243, b248, b256, b259, b274, b275} 
& 
& \cite{b19, b20, b23, b24, b60, b218, b220, b223, b243, b248, b256, b257, b259, b274} 
& \cite{b23, b24} \\ 
\midrule

\textbf{Probabilistic and Statistical Methods} 
& \cite{b8, b10, b15, b54, b55, b56, b123, b124, b211, b218, b219, b220, b221, b244, b270} 
& \cite{b8, b10, b15, b54, b55, b56, b57, b123, b126, b213, b218, b219, b220, b221, b244, b270} 
& \cite{b8, b10, b15, b55, b56, b57, b126, b128, b220, b244} 
& \cite{b15, b55, b56, b128, b220, b221, b244} 
& \\ 
\midrule

\textbf{Machine Learning and Deep Learning-Based Methods} 
& \cite{b25, b26, b28, b29, b30, b31, b33, b34, b35, b36, b42, b43, b45, b75, b77, b78, b79, b101, b105, b111, b114, b115, b134, b146, b147, b149, b150, b152, b153, b157, b160, b180, b181, b185, b188, b189, b190, b192, b196, b223, b224, b225, b229, b233, b234, b235, b236, b239, b240, b241, b243, b245, b246, b247, b248, b250, b258, b259, b260, b266, b267, b268, b269, b276, b278, b279, b280, b283} 
& \cite{b27, b30, b31, b42, b43, b45, b77, b101, b105, b134, b157, b162, b163, b180, b189, b190, b197, b222, b224, b227, b239, b240, b241, b248, b249, b258, b259, b260, b266, b267, b268} 
& \cite{b28, b30, b31, b34, b42, b43, b78, b79, b101, b105, b114, b115, b134, b153, b157, b162, b186, b188, b198, b225, b226, b229, b233, b235, b236, b243, b245, b246, b247, b248, b250, b258, b259, b260, b276, b278, b280} 
& \cite{b29, b32, b33, b35, b42, b75, b101, b108, b114, b134, b146, b149, b150, b152, b160, b180, b181, b185, b188, b189, b223, b224, b225, b226, b233, b235, b236, b243, b248, b258, b259, b266, b269, b276, b279} 
& \cite{b28, b33, b35, b42, b77, b79, b115, b146, b147, b149, b157, b190, b192, b196, b224, b226, b229, b233, b236, b238, b240, b243, b248, b258, b259, b283} \\ 
\midrule

\textbf{Cell Tracking and Biomedical Applications} 
& \cite{b81, b82, b84, b88, b101, b107, b120, b121, b134, b149, b157, b162, b163, b203, b205, b206, b229, b230, b231, b232, b242, b251, b252, b253, b254} 
& \cite{b81, b82, b84, b89, b90, b104, b106, b107, b120, b122, b134, b149, b157, b162, b163, b203, b205, b206, b229, b230, b231, b232, b242, b251, b252, b253, b254} 
& \cite{b82, b84, b86, b89, b90, b101, b107, b122, b134, b149, b162, b163, b205, b206, b229, b231, b233, b242, b251, b253} 
& \cite{b81, b82, b85, b88, b89, b104, b107, b114, b120, b121, b134, b149, b157, b162, b163, b203, b205, b206, b229, b230, b231, b232, b242, b251, b252, b254} 
& \cite{b84, b86, b89, b115, b134, b149, b203, b205, b206, b229, b230, b231, b232, b242, b251, b253} \\ 
\bottomrule
\end{tabularx}
\end{table*}

\subsection{3D Cell Tracking: Challenges and Related Works}

The advent of advanced imaging technologies has revolutionized the field of bio-imaging, presenting researchers with the opportunity to explore cellular behaviors in unprecedented detail. However, this technological leap comes with its own set of challenges, especially when dealing with 3D imaging data. Unlike 2D videos, which often suffer from occlusions and missed detections, 3D cell tracking offers a more comprehensive view of cellular motility, capturing intricate spatial relationships that are otherwise hidden in 2D projections. Despite these advantages, 3D datasets introduce significant computational complexity due to the sheer volume of data and the need to process information across multiple dimensions—typically three spatial dimensions plus a temporal dimension, resulting in a minimum of four dimensions in 3D videos.

This increase in dimensionality not only demands more robust computational resources but also introduces greater degrees of freedom in object tracking, which further complicates the analysis. As a result, the computational burden is substantially higher compared to 2D video processing, necessitating advanced techniques in distributed computing to handle the massive data volumes efficiently. In response to these challenges, we previously developed a distributed pipeline using Apache Spark for large-scale fMRI image analysis \cite{b170, b171}. For 3D microscopy videos of \textit{T. gondii}, we leveraged the power of distributed and parallel programming with Dask, significantly improving the processing efficiency.

In 3D cell tracking, the data is typically organized into slices, where each slice represents a cross-section of the depth (z-slice). Each frame in a 3D video is composed of multiple such slices, forming a volumetric dataset. Figure \ref{fig:29}-B illustrates the slice arrangement in a sample 3D video of \textit{T. gondii}, while Figure \ref{fig:29}-A shows the model we used in \cite{b114} for tracking cells in a 3D environment. The comprehensive details of data acquisition and imaging processes are provided in \cite{b114}.

In our work \cite{b114}, we employed a multi-step approach to track cells in 3D videos. Initially, we preprocessed the cell data and used a detection mechanism to identify individual cell particles. We then combined these discrete particles into 3D masses by applying the 3D Connected Component Labeling method. Following this, we computed the centers of mass for each cell and filtered out noise from the system. The final tracking of the cells was accomplished using the Hungarian algorithm, a method known for its efficiency in solving assignment problems. The results of this process are depicted in Figure \ref{fig:30}. 

To address the challenges posed by large data volumes, we analyzed our pipeline to identify bottlenecks and subsequently parallelized these components using Joblib backends and Python’s multiprocessing library \cite{b114}. In our later work \cite{b115}, we extended this approach by developing a distributed pipeline using Dask, deploying it on clusters within the Google Cloud Platform (GCP) to handle large-scale 3D microscopy data more effectively.

The application of 3D convolutional networks (3D CNNs) represents another significant advancement in this field. These networks operate similarly to their 2D counterparts but are designed to process 3D input patches, making them particularly well-suited for tasks such as mitosis detection in 3D biomedical images. For instance, in \cite{b108}, the authors introduced an end-to-end framework called F3D-CNN, specifically designed for mitosis detection. This model is trained on domain-specific data, with video frames sequentially fed into a Fully Convolutional Network (FCN) to generate score maps that indicate the location of candidate events in each frame. These candidates are then categorized by integrating temporal information from neighboring frames, with a Conditional Random Field (CRF)-based model used for temporal localization.

Further advancements include the V-net model for 3D biomedical image segmentation, proposed by F. Milletari et al. \cite{b113}, and its application in 3D brain image analysis. While numerous deep learning-based methods have been developed for cell segmentation, there is a noticeable gap in research focused specifically on cell tracking and motion discovery in 3D datasets.

Several other studies have explored 3D cell tracking from different perspectives and applications \cite{b116, b117, b118, b119, b120}. Notably, C.P. Kappe et al. \cite{b121} proposed a method based on dense optical flow, where local flow information is reconstructed from 3D + time videos. By using interpolation, they reconstructed the global vector field and computed the center of mass for each cell. To manage the large-scale data, they employed various interpolation methods, further enhancing the efficiency of their approach.

In our research on 3D cell tracking, we also explored the clustering of motion phenotypes. Figure \ref{fig:33-1} illustrates the clustering results of 3D + time \textit{T. gondii} trajectories, where specific motion patterns were identified and grouped. We employed a sophisticated pipeline that involves AR (Autoregressive) parameterization, dimensionality reduction, and geodesic metric computation to analyze and cluster these trajectories. The detailed model is depicted in Figure \ref{fig:34}-A. By qualitatively examining the AR matrices within each cluster (Figure \ref{fig:34}-B), we observed a high degree of similarity within clusters and clear distinctions between different clusters.

The ongoing advancements in 3D cell tracking techniques continue to push the boundaries of biomedical research, enabling more accurate and detailed analysis of cellular behaviors in complex environments. As imaging technologies evolve, so too will the methods we use to track and analyze cells, ultimately leading to a deeper understanding of the fundamental processes that govern life at the cellular level.

\section{Conclusion}

In this paper, we have undertaken a comprehensive review of both traditional and state-of-the-art object tracking methods, analyzing them through various lenses to provide a thorough understanding of the evolution and current trends in this critical field. We began by exploring foundational techniques and tools, setting the stage for a deeper dive into the classifications and methodologies that have shaped the landscape of object tracking. Through our proposed taxonomy, we systematically categorized the vast array of research into four distinct groups: \textit{Traditional Methods, Statistical Methods, Feature-Based Methods}, and \textit{Machine Learning and Deep Learning-Based Methods}. This structure allowed us to not only examine each method's core principles and applications but also to highlight the progression from conventional approaches to the cutting-edge techniques that are now pushing the boundaries of what's possible in object tracking.

A particular focus of this review has been on the biomedical applications of object tracking, where the fusion of computer vision and deep learning with cell biology has unlocked new potentials for disease diagnosis, drug discovery, and the study of complex biological processes. The ability to accurately segment and track cells and sub-cellular components in biomedical images is not merely an academic pursuit; it has profound implications for understanding the mechanobiology of diseases, the development of new therapies, and the advancement of personalized medicine. As we discussed, the recent advancements in computer vision—particularly those driven by deep learning—have revolutionized how researchers analyze video data in cellular biology, offering unprecedented precision and insight.

Our review of the latest technologies in computer vision and cell tracking reveals a clear trajectory toward more sophisticated and integrated systems. The future of spatiotemporal analysis in biological systems lies in the development of end-to-end, unsupervised models that are robust, multi-domain, and multi-scale, capable of tracking multiple objects simultaneously while providing rapid and reliable motion analysis. Such systems will need to navigate the challenges of high-dimensional data, diverse biological environments, and the demand for real-time analysis, all while maintaining accuracy and adaptability across various applications.

Table \ref{comprehensive-table} systematically classifies the referenced papers based on the key characteristics essential for next-generation object tracking systems: \textit{Extensiveness, Robustness, Trainability, Multi-Domain Compatibility,} and \textit{Scalability}. Each category reflects a critical capability required to meet the complex demands of modern biomedical research. This structured classification not only underscores the depth of existing methodologies but also guides researchers in identifying gaps and opportunities for innovation.

We believe that this extensive survey will serve as a valuable resource for both seasoned researchers and newcomers to the field, offering a solid foundation upon which future innovations can be built. The intersection of computational sciences and biology is a fertile ground for breakthroughs that can have a profound impact on healthcare and life sciences. By understanding the strengths and limitations of current object tracking methods, researchers are better equipped to develop the next generation of tools that will further bridge the gap between technology and biology.

As we look to the future, it is evident that the integration of bio-imaging with advanced computer vision and deep learning techniques will continue to transform the landscape of biomedical research. The need for robust, scalable, and adaptable tracking systems is more pressing than ever, and the innovations in this field will undoubtedly contribute to significant advancements in our understanding of biological systems and their applications in medicine. We anticipate that this review will inspire new research directions and foster collaborations that will drive the field forward, ultimately leading to more effective and insightful analyses of complex biological phenomena.

\begin{IEEEbiography}[{\includegraphics[width=1.25in,height=1.75in,clip,keepaspectratio]{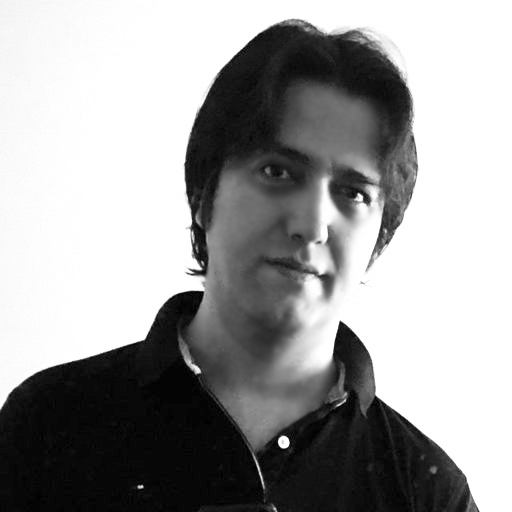}}]{Mojtaba S. Fazli } 

is a Postdoctoral Research Scholar and Lead AI Scientist at \textit{Stanford University}. Before joining Stanford, Dr. Fazli held the role of Senior Research Data Scientist and Open Innovation Scholar at the \textit{Novartis Institute for Biomedical Research}. In addition to his work at Novartis, he served as a Fellow with the \textit{Gates Foundation} under the MalDA Consortium, contributing to impactful global health research. Dr. Fazli also pursued postdoctoral research at \textit{Harvard University}, where he led pioneering advancements in biomedical AI. He holds a Ph.D. in Computer Science with a minor in Mathematics from the University of Georgia, USA. He also earned a Doctorate in Business Administration and an M.Sc. in Economics and Management from the University of Montesquieu Bordeaux IV, France. Additionally, he received an M.Sc. in Artificial Intelligence and Robotics and a B.Sc. in Computer Engineering from the University of Tehran, Iran.
\end{IEEEbiography}

\begin{IEEEbiography}[{\includegraphics[width=1in,height=1.25in,clip,keepaspectratio]{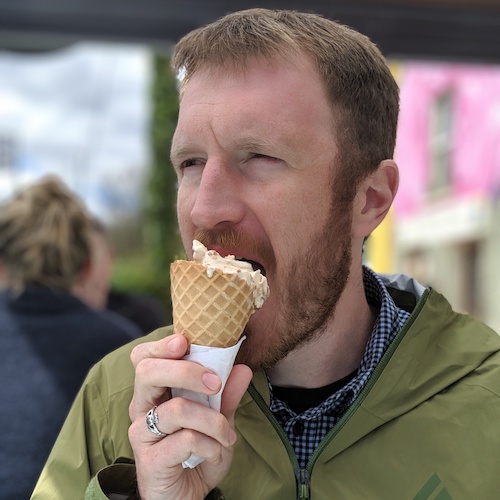}}]{Shannon Quinn} is an Associate Professor jointly appointed in the Computer Science and Cellular Biology Departments at the University of Georgia.

He received his Ph.D. in Computational Biology from the Joint Carnegie Mellon-University of Pittsburgh Ph.D. Program in Computational Biology, working with his advisor Dr. Chakra Chennubhotla. He received his M.S. in Computational Biology from Carnegie Mellon University under the direction of his thesis advisor Dr. Robert Murphy. He completed his B.S. in Computer Science at the Georgia Institute of Technology.

\end{IEEEbiography}

\EOD

\end{document}